\documentclass[10pt,twocolumn,letterpaper]{article}
\usepackage{cvpr}              

\usepackage[percent]{overpic}
\usepackage{contour}
\contourlength{0.05em}
\usepackage{graphicx}
\usepackage{tikz}

\newcommand{\TODO}[1]{\textbf{\color{red}[TODO: #1]}}

\renewcommand{\TODO}[1]{}

\usepackage{microtype}

\renewcommand{\paragraph}[1]{\vspace{.5em}\noindent\textbf{#1.}}

\usepackage{graphicx}
\usepackage{float}
\usepackage{tikz}
\usepackage{adjustbox}
\usepackage{makecell}
\usepackage{cuted}
\usepackage{array}
\graphicspath{{assets/}}

\makeatletter
\@ifundefined{ifolbedopagebackref}{%
  \newif\ifolbedopagebackref
  \olbedopagebackreftrue
}{}
\makeatother
\definecolor{cvprblue}{rgb}{0.21,0.49,0.74}
\ifolbedopagebackref
  \usepackage[pagebackref,breaklinks,colorlinks,allcolors=cvprblue]{hyperref}
\else
  \usepackage[breaklinks,colorlinks,allcolors=cvprblue]{hyperref}
\fi

\def\paperID{} 
\def\confName{CVPR}
\def\confYear{2026}

\title{\textit{Olbedo}: An Albedo and Shading Aerial Dataset for Large-Scale Outdoor Environments}

\author{
Shuang Song\textsuperscript{a\textdagger }\quad Debao Huang\textsuperscript{a\textdagger }\quad Deyan Deng\textsuperscript{a}\quad Haolin Xiong\textsuperscript{b,c}\quad Yang Tang\textsuperscript{a} \\
 Yajie Zhao\textsuperscript{b,c}\quad Rongjun Qin\textsuperscript{a*}
\vspace{1em}
\\
a The Ohio State University \\
b University of Southern California \\
c Institute for Creative Technologies \\
\textdagger Contributed equally \quad * Corresponding author \\
{\tt\small \url{https://gdaosu.github.io/olbedo/}}
}

\begin{document}

\maketitle

\begin{strip}\centering
    \vspace{-50px}
    \captionsetup{type=figure}

    \begin{subfigure}{0.455\textwidth} 
\centering 
\begin{overpic}[width=\linewidth]{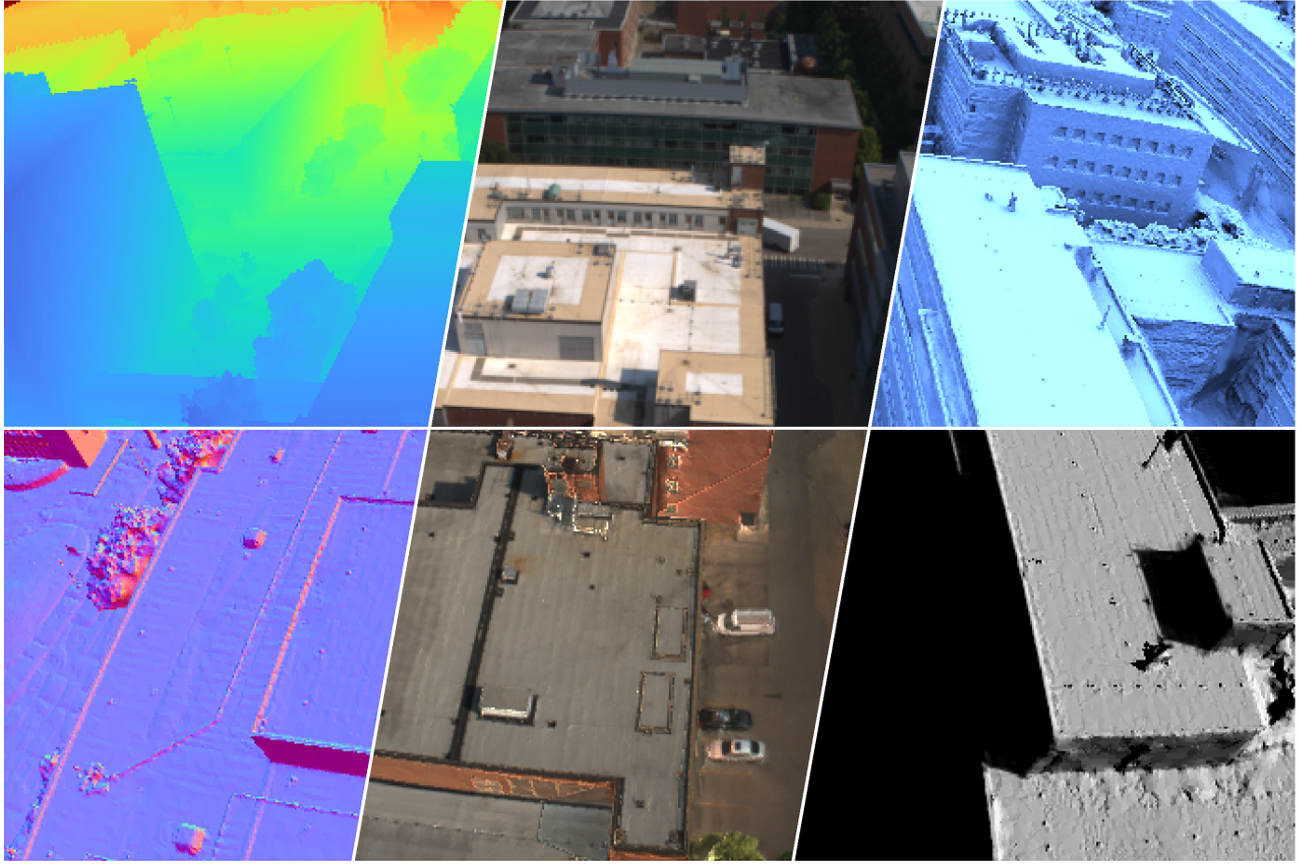}
    \put(5,60){\contour{black}{\color{white}\bfseries Depth}}
    \put(45,60){\contour{black}{\color{white}\bfseries RGB}}
    \put(75,60){\contour{black}{\color{white}\bfseries Sky Shading}}
    \put(5,27){\contour{black}{\color{white}\bfseries Normal}}
    \put(42,27){\contour{black}{\color{white}\bfseries Albedo}}
    \put(74,27){\contour{black}{\color{white}\bfseries Sun Shading}}
\end{overpic}
\vspace{-15px}
\caption{}
\label{fig:overview1} 
\end{subfigure} 
\begin{subfigure}{0.535\textwidth} 
\centering 
\includegraphics[width=\linewidth]{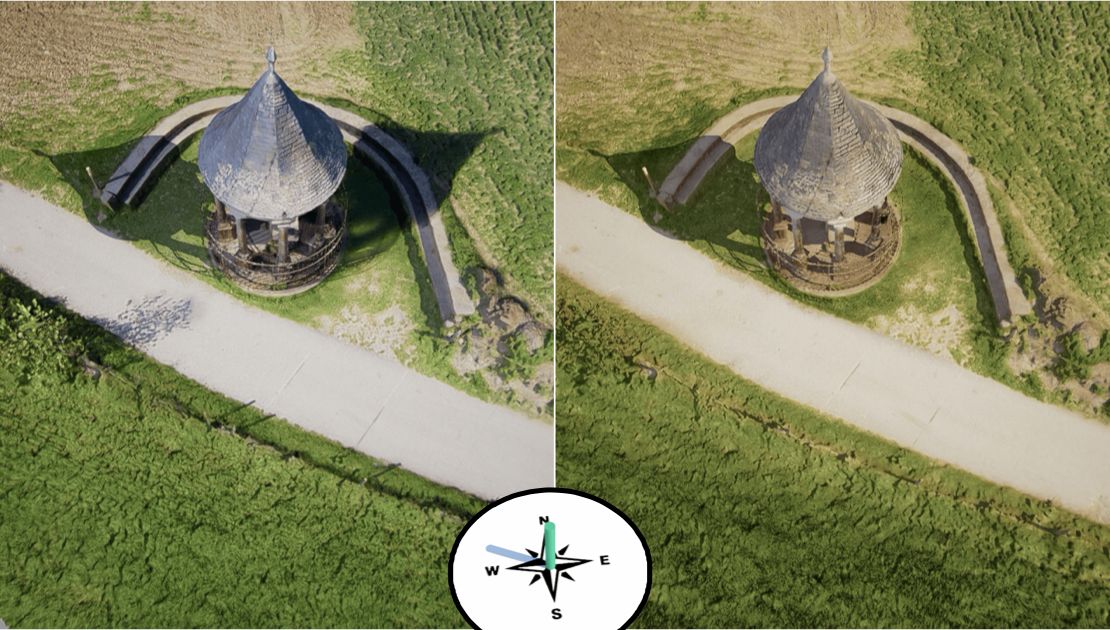}
\vspace{-15px}
\caption{}
\label{fig:overview2} 
\end{subfigure} 
\vspace{-10px}
\caption{(a) Overview of the \textit{Olbedo} dataset. For each RGB image, \textit{Olbedo} provides pseudo-ground-truth layers for albedo, metric depth, surface normal, sun shading, and sky shading. The RGB and albedo layers are stored as HDR images. (b) Relighting 3D models under novel lighting conditions (direction specified by the compass). RGB textures (left) retain shading, causing shadow duplication artifacts, while albedo textures (right) eliminate them for plausible results. } 
\label{fig:overview} 
\vspace{-10px}
\end{strip}
\begin{abstract}
Intrinsic image decomposition (IID) of outdoor scenes is crucial for relighting, editing, and understanding large-scale environments, but progress has been limited by the lack of real-world datasets with reliable albedo and shading supervision. We introduce \textit{Olbedo}, a large-scale aerial dataset for outdoor albedo--shading decomposition in the wild. \textit{Olbedo} contains 5,664 UAV images captured across four landscape types, multiple years, and diverse illumination conditions. Each view is accompanied by multi-view consistent albedo and shading maps, metric depth, surface normals, sun and sky shading components, camera poses, and, for recent flights, measured HDR sky domes. These annotations are derived from an inverse-rendering refinement pipeline over multi-view stereo reconstructions and calibrated sky illumination, together with per-pixel confidence masks. We demonstrate that \textit{Olbedo} enables state-of-the-art IID models, including both CNN-based and diffusion-based architectures originally trained on synthetic indoor data, to generalize to real outdoor imagery: fine-tuning on \textit{Olbedo} significantly improves single-view outdoor albedo prediction on the MatrixCity benchmark. We further illustrate applications of \textit{Olbedo}-trained models to multi-view consistent relighting of 3D assets, material editing, and scene change analysis for urban digital twins. We release the dataset, baseline models, and an evaluation protocol to support future research in outdoor intrinsic decomposition and illumination-aware aerial vision.
\vspace{-15px}
\end{abstract}

\section{Introduction}
\label{sec:intro}

Intrinsic image decomposition (IID)~\cite{garcesSurveyIntrinsicImages2022, Fan_2018_CVPR}, the process of separating an image into its reflectance (albedo) and shading components, is a fundamental, long-standing problem in computer vision and graphics. Recovering intrinsic properties is essential for robust material understanding, physically based rendering, and high-fidelity 3D reconstruction. While deep learning has significantly advanced IID~\cite{litmanMaterialFusionEnhancingInverse2025,zengRGB-XImageDecomposition2024,yuInverseRenderNetLearningSingle2019,chenIntrinsicAnythingLearningDiffusion2024, luoNIIDNetAdaptingSurface2020,careaga2024colorful}, much of this progress remains confined to indoor or synthetic environments.

The primary barrier to advancing IID for real-world scenarios is the profound difficulty of acquiring dense, high-fidelity ground-truth albedo in large-scale outdoor environments. Unlike indoor scenes where lighting can be controlled~\cite{zhuLearningbasedInverseRendering2022,Roberts_2021_ICCV}, outdoor scenes are subject to dynamic, complex illumination from the sun and sky. Capturing ground-truth albedo in the wild, at scale, has been considered extremely challenging. Consequently, there is no large-scale, real-world dataset to train IID models for complex aerial scenarios, and models trained on synthetic or indoor data often fail to generalize to real aerial imagery.

To bridge this gap, we present \textit{Olbedo}, the first large-scale, real-world dataset specifically designed for albedo recovery in aerial multi-view stereo (MVS). \textit{Olbedo} comprises high-resolution UAV imagery, dense pseudo-ground-truth albedo and shading maps, and auxiliary geometry and illumination data generated via a physics-based inverse-rendering pipeline~\cite{songNOVELINTRINSICIMAGE2022,songGeneralAlbedoRecovery2024}.

This dataset addresses a bottleneck in 3D content creation: weather‑dependent capture bakes illumination into textures, which undermines physically based rendering (PBR) workflows and imposes significant logistical burdens. \textit{Olbedo} provides the supervision needed to learn lighting‑invariant reflectance with multi‑view consistency, enabling flexible, weather‑agnostic capture pipelines.

Using our dataset, we adapt state-of-the-art IID models to single-view outdoor albedo prediction: diffusion-based~\cite{Kocsis_2024_CVPR,keMarigoldAffordableAdaptation2025,zengRGB-XImageDecomposition2024} via Low-Rank Adaptation (LoRA)~\cite{hu2022lora} fine-tuning and CNN-based~\cite{luoNIIDNetAdaptingSurface2020} via standard fine-tuning, yielding improved generalization to complex outdoor scenes. Moreover, \textit{Olbedo} complements synthetic datasets in IID: combining synthetic pretraining with \textit{Olbedo} fine-tuning further boosts outdoor albedo inference across both architectures, as shown in our experiments. Subsequent applications demonstrate benefits for relighting, 3D modeling, and scene understanding.

\section{Related Work}
\label{sec:related}

Our work is positioned at the intersection of IID and methods for creating ground-truth albedo datasets. We address a critical gap by providing the first large-scale, real-world dataset for aerial albedo recovery, which has been a major bottleneck for the field.

\paragraph{Intrinsic Image Decomposition} IID is a foundational, yet highly ill-posed problem in computer vision~\cite{garcesSurveyIntrinsicImages2022, Fan_2018_CVPR}. It aims to factor an input image $I$ into illumination-invariant reflectance (albedo) $R$ and illumination-variant shading $S$, such that $I = R \cdot S$~\cite{landLightnessRetinexTheory1971}.
Early approaches relied on hand-crafted priors and optimization frameworks, assuming that shading is spatially smooth while reflectance is piecewise constant~\cite{barronShapeAlbedoIllumination2012,barronColorConstancyIntrinsic2012,bousseauUserAssistedIntrinsicImages2009,shengIntrinsicImageDecomposition2020}. These priors often fail on complex, non-Lambertian surfaces in real scenes~\cite{barronShapeAlbedoIllumination2012}.

Learning-based approaches have significantly advanced IID by leveraging large datasets to learn complex priors~\cite{Fan_2018_CVPR,baslamisliShadingNetImageIntrinsics2021}. Convolutional neural networks (CNNs) have been trained to learn these priors directly from data~\cite{innamoratiDecomposingSingleImages2017}. These methods have evolved across several paradigms. Supervised methods learn a direct mapping from an input image to its decomposed layers but are entirely dependent on large-scale, pixel-aligned ground-truth~\cite{baslamisliJointLearningIntrinsic2018,zengRGB-XImageDecomposition2024}. Self-supervised and unsupervised methods leverage the image formation model itself as a loss, often using reconstruction error to guide training on unlabeled data~\cite{yuInverseRenderNetLearningSingle2019,yuOutdoorInverseRendering2021}. More recently, diffusion-based models~\cite{keMarigoldAffordableAdaptation2025,Kocsis_2024_CVPR,zengRGB-XImageDecomposition2024,chenIntrinsicAnythingLearningDiffusion2024} have achieved state-of-the-art performance by leveraging the rich foundational priors in large diffusion models~\cite{rombach2021highresolution,ramesh2022hierarchical}.

\paragraph{The Ground-Truth Dataset Bottleneck} The primary limitation in IID research is the profound difficulty of acquiring dense, pixel-perfect ground-truth for albedo and shading. This challenge has shaped the field, with existing datasets falling into three main categories based on their creation methodology.

\begin{itemize}
\item Synthetic datasets: To obtain perfect, dense ground-truth, many datasets rely on PBR. Key examples include MPI Sintel~\cite{butlerNaturalisticOpenSource2012}, CGIntrinsics~\cite{liCGIntrinsicsBetterIntrinsic2018}, InteriorVerse~\cite{zhuLearningbasedInverseRendering2022}, and Hypersim~\cite{Roberts_2021_ICCV}. While providing high-quality labels, they are often object-centric or indoor-focused and suffer from a significant "sim-to-real" gap. Models trained on them struggle to generalize to the complexity of real-world photography.

\item Real-world controlled datasets: To capture real-world statistics, some datasets use controlled lab environments. The MIT Intrinsics dataset~\cite{grosseGroundTruthDataset2009}, MID~\cite{murmannMultiilluminationDatasetIndoor2019, careagaIntrinsicImageDecomposition2023}, and DIR~\cite{Choi_2025_ICCV} captured objects or indoor scenes under multiple, known lighting conditions. This multi-illumination data allows for the computation of pseudo-ground-truth albedo. However, this methodology is not scalable and is fundamentally restricted to small, indoor environments.

\item Real-world "in-the-wild" datasets: The Intrinsic Images in the Wild (IIW) dataset~\cite{bellIntrinsicImagesWild2014} provided a leap in scale by collecting real-world images. However, its ground-truth consists of sparse, relative human judgments (e.g., "point A is lighter than point B"). It provides no dense albedo maps.
\end{itemize}

Existing IID methods are limited by datasets that are either (a) synthetic, (b) indoor-only, or (c) lack of dense ground-truth. There is a major gap: no large-scale, real-world dataset for intrinsic decomposition in complex, outdoor, aerial environments.

Our work, \textit{Olbedo}, directly addresses this gap by introducing a novel pipeline to generate the first dense, large-scale, real-world pseudo-ground-truth supervision for aerial albedo recovery.


\section{\textit{Olbedo} Dataset}
\label{sec:dataset}

\subsection{Overview}
The \textit{Olbedo} dataset provides a large-scale benchmark for albedo-shading decomposition of outdoor aerial imagery in the wild. Our dataset includes decomposed albedo and shading maps as pseudo-ground-truth for each image. We also provide intermediate geometric features, including metric depth, normals, sun shading, sky shading, and (for recent flights) sky radiance images. While accurate geometry data are already available in existing datasets, our experiments intentionally isolate the contribution of albedo supervision; depth and normals are released as by-products of the reconstruction pipeline and for future multi-task use.


\subsection{Data Collection}

\begin{figure}[!htbp]
    \centering
    \vspace{-10px}
    \begin{subfigure}[b]{0.49\linewidth}
        \centering
        \includegraphics[width=\linewidth]{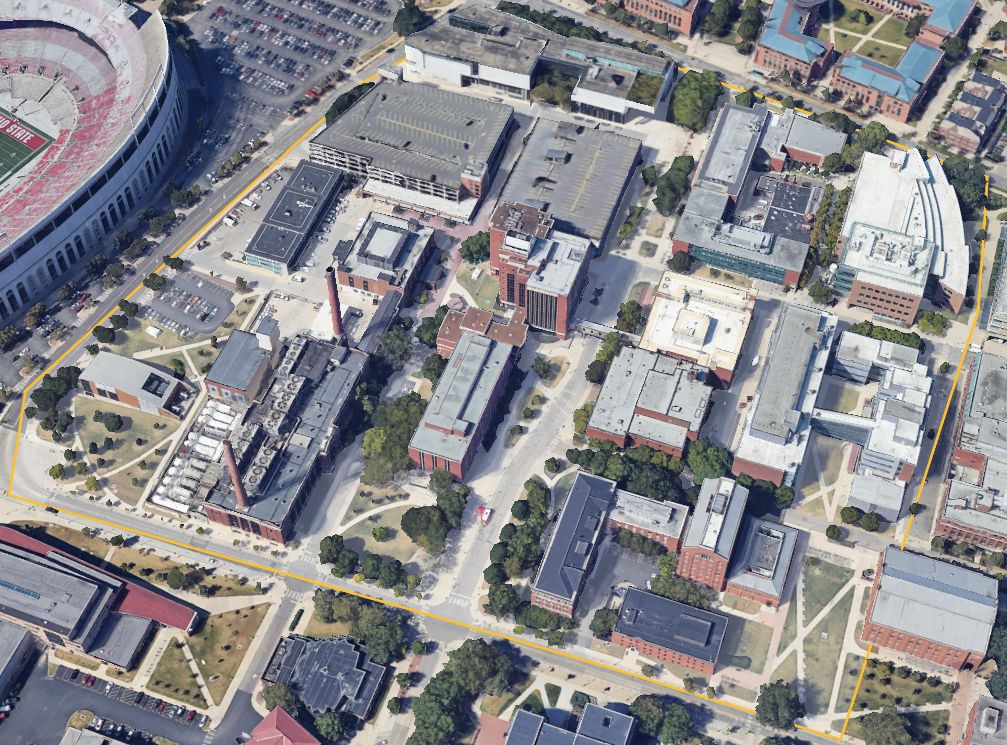}
        \caption{Office}
        \label{fig:subfig1}
    \end{subfigure}
    \hfill
    \begin{subfigure}[b]{0.49\linewidth}
        \centering
        \includegraphics[width=\linewidth]{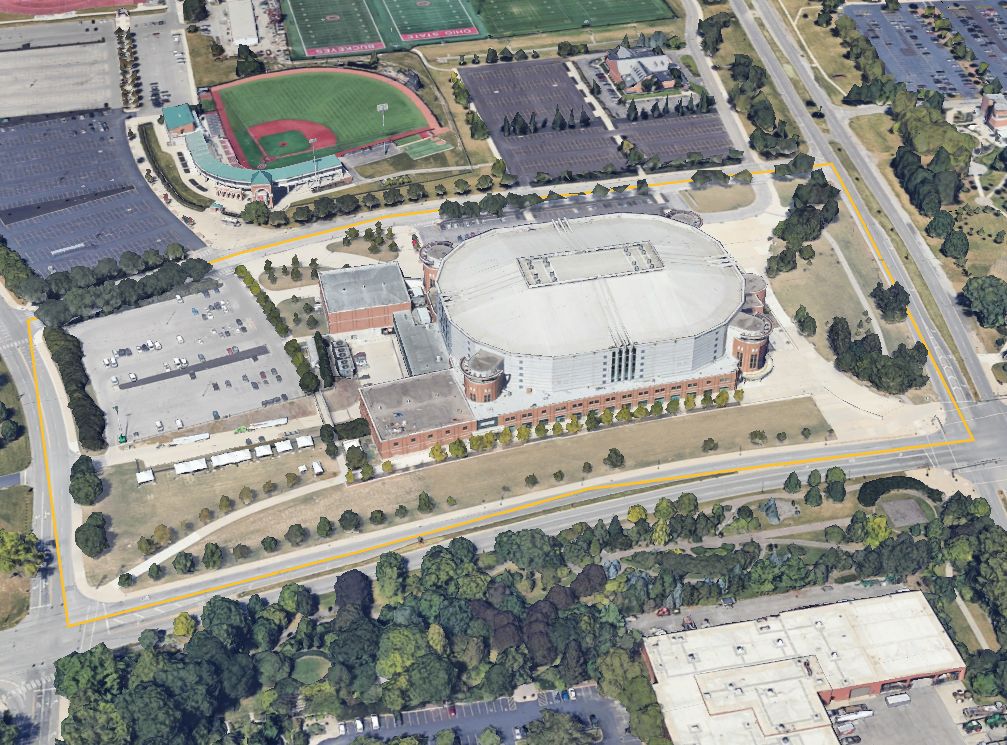}
        \caption{Arena}
        \label{fig:subfig2}
    \end{subfigure}
    \vskip\baselineskip
    \vspace{-10px}
    \begin{subfigure}[b]{0.49\linewidth}
        \centering
        \includegraphics[width=\linewidth]{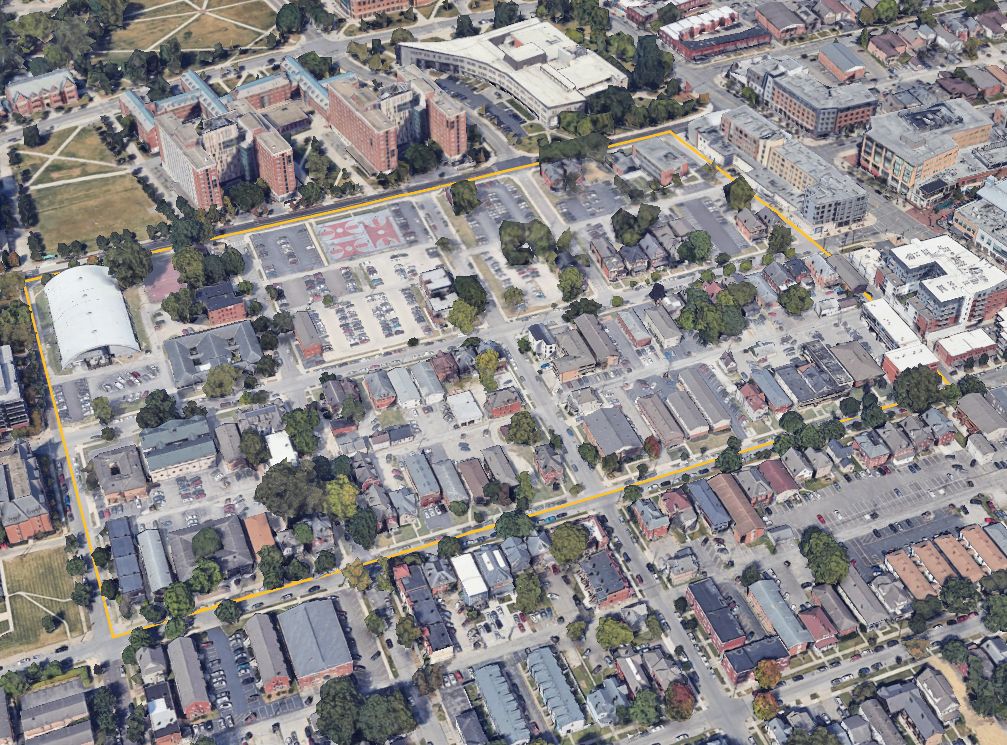}
        \caption{Residential}
        \label{fig:subfig3}
    \end{subfigure}
    \hfill
    \begin{subfigure}[b]{0.49\linewidth}
        \centering
        \includegraphics[width=\linewidth]{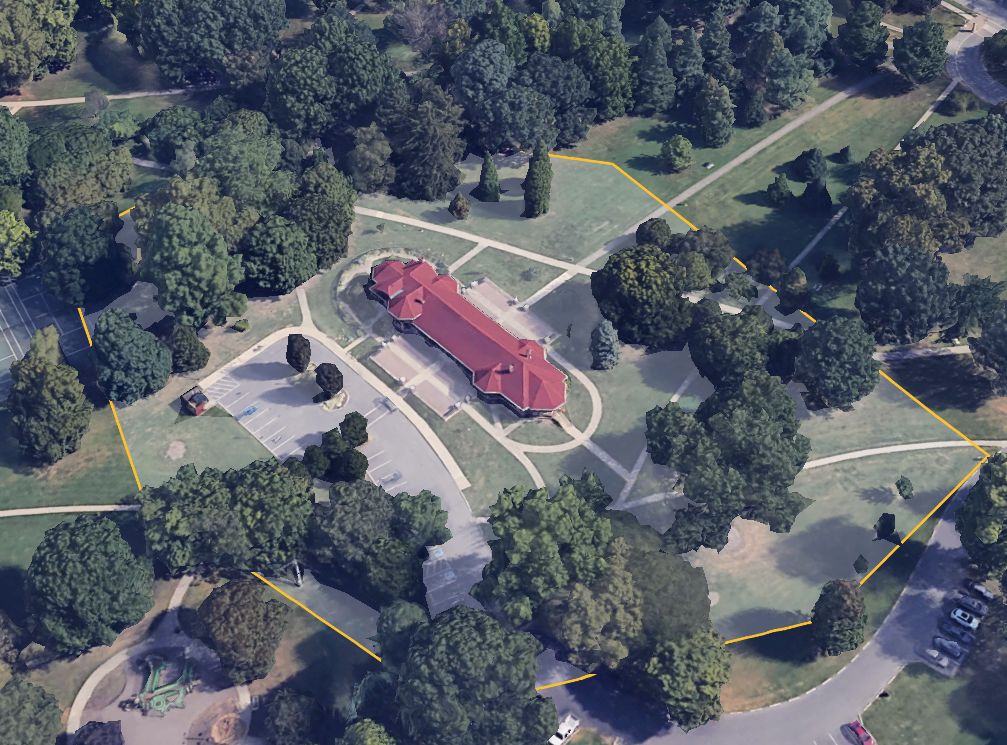}
        \caption{Park}
        \label{fig:subfig4}
    \end{subfigure}
    \vspace{-10px}
    \caption{Landscapes of the \textit{Olbedo} dataset. (Credit: Google)}
    \vspace{-15px}
    \label{fig:types_of_landscape}
\end{figure}

The \textit{Olbedo} dataset is derived from aerial datasets captured in RAW format, along with a sky radiance image captured simultaneously during drone flights. The sun position is determined from astronomical ephemeris using UTC time and site location; no special instruments are required. Precise 3D geometry is essential for albedo-shading decomposition. Accordingly, we use a photogrammetry pipeline to solve accurate camera poses and reconstruct highly precise surface meshes from our in-house collected, multi-temporal image sets. The image sets include repeated data-collection flights over up to four years. The dataset covers four landscape types, as shown in \cref{fig:types_of_landscape}.

\paragraph{Temporal Image Collection}
We collected aerial images over multiple flights for each scene to cover diverse lighting conditions. Each flight lasts approximately 10 to 25 minutes, depending on site area. For each scene, we conducted 2 to 4 flights at different times of day across multiple years to capture the scene under varying illumination. Each flight captures around 50 to 230 images with 80\% forward and side overlap to ensure robust reconstruction. The images are captured in a nadir view with slight oblique angles to cover vertical surfaces.

\begin{figure}[!htpb]
    \centering
    
    \begin{subfigure}[b]{0.48\linewidth}
        \centering
        \includegraphics[width=\linewidth]{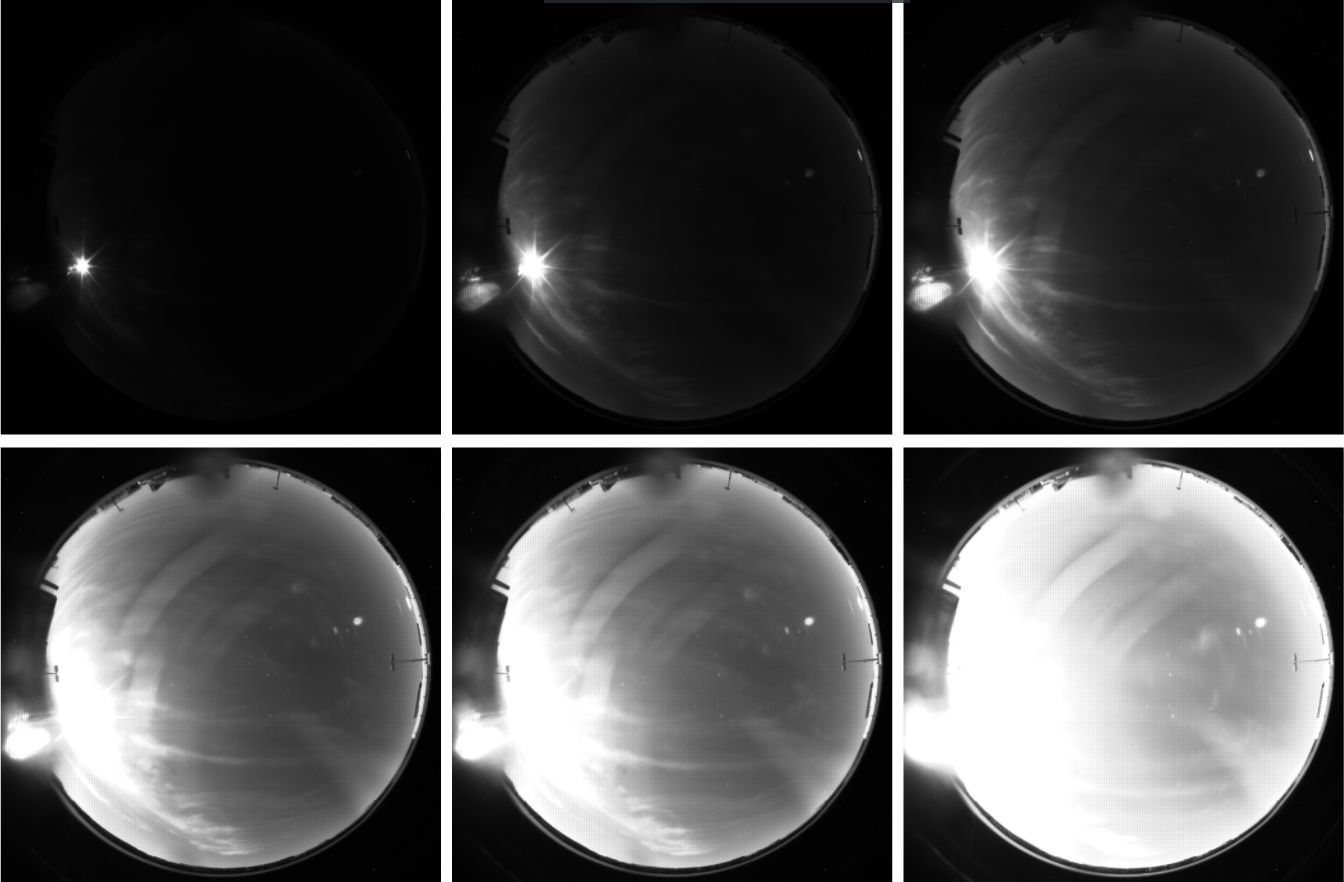}
        \caption{Exposure-bracketed raw sky frames (0.5--32 s).}
        \label{fig:sky_radiance_sequence}
    \end{subfigure}
    \hfill
    \begin{subfigure}[b]{0.48\linewidth}
        \centering
        \includegraphics[width=\linewidth]{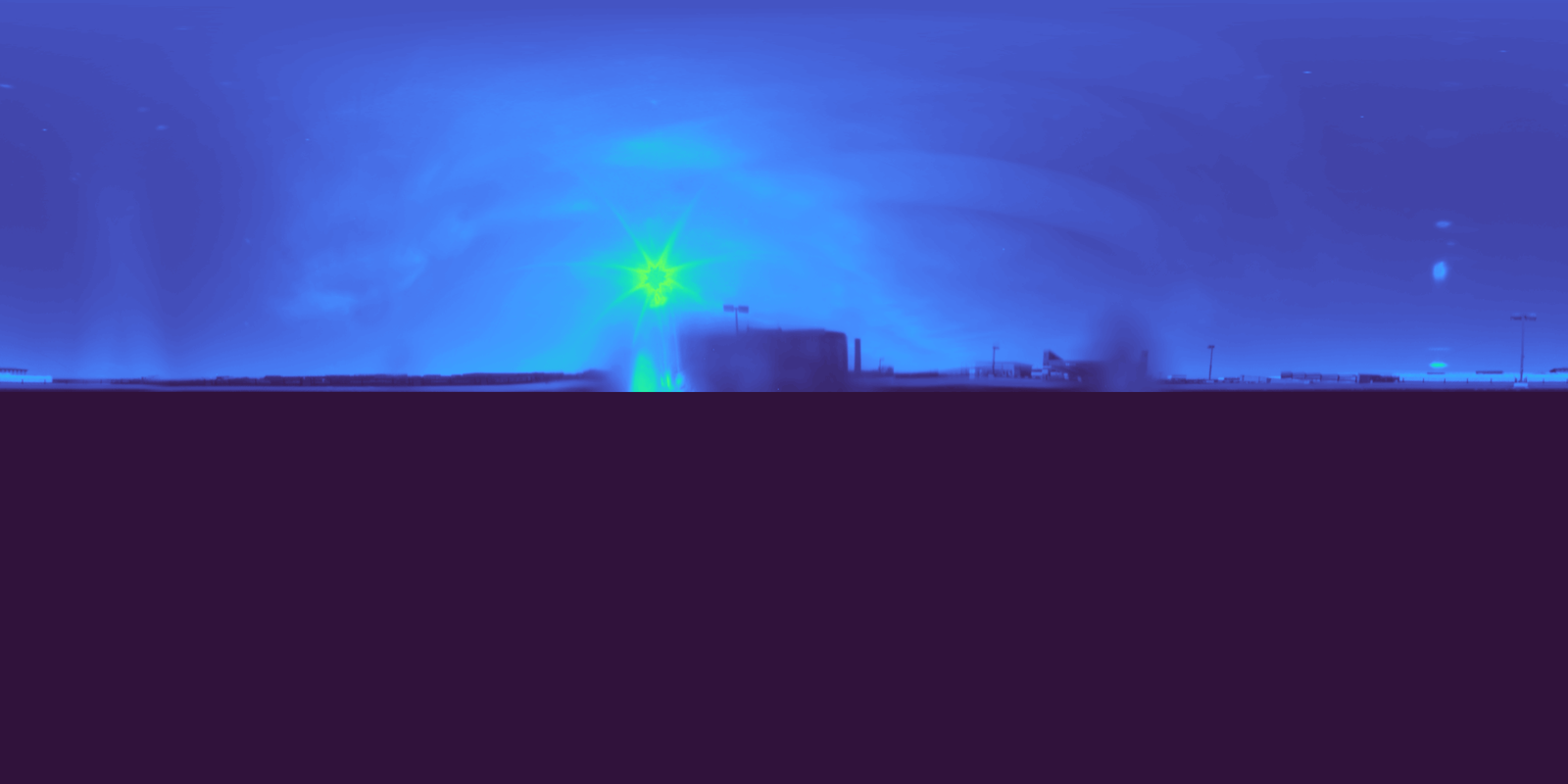}
        \caption{Monochromatic HDR sky radiance image in equirectangular projection.}
        \label{fig:sky_radiance_equirectangular}
    \end{subfigure}
    \vspace{-10px}
    \caption{Sky radiance acquisition and representation. Left: exposure-bracketed raw sky images. Right: merged HDR sky dome in equirectangular format.}
    \label{fig:sky_radiance}
    \vspace{-8px}
\end{figure}

\paragraph{Sky Radiance Capture}
For flights in 2025, we capture an HDR all-sky radiance dome with a 180$^\circ$ fisheye rig (Sony ExView II CCD, IR and ND filters). Each sequence brackets 11 exposures (0.5--32 s) recorded as 16-bit FITS and merged into a 32-bit EXR radiance map following~\cite{debevecRecoveringHighDynamic1997}. The sky camera is aligned to the world frame via surveyed control points (GNSS RTK/total station), and the dome is provided in equirectangular format, as shown in \cref{fig:sky_radiance}. We use luminance-only sky domes to model $S_{sky}$ and ephemeris-based sun direction for $S_{sun}$. Sky radiance maps and metadata are released per flight; when sky radiance is unavailable, an analytic sky model can be applied.

\begin{figure}[!htpb]
    \centering
    \vspace{-5px}
    \begin{subfigure}[b]{0.32\linewidth}
        \centering
        \includegraphics[width=\linewidth]{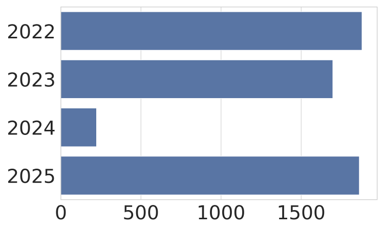}
        \caption{}
        \label{fig:histogram_year}
    \end{subfigure}
    \begin{subfigure}[b]{0.32\linewidth}
        \centering
        \includegraphics[width=\linewidth]{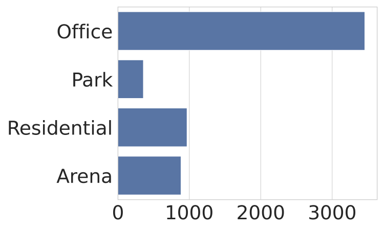}
        \caption{}
        \label{fig:histogram_site}
    \end{subfigure}
    \begin{subfigure}[b]{0.32\linewidth}
        \centering
        \includegraphics[width=\linewidth]{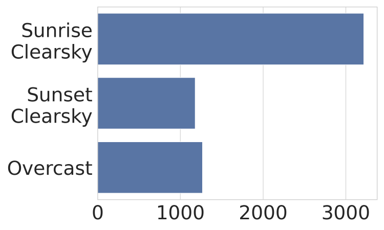}
        \caption{}
        \label{fig:histogram_lighting}
    \end{subfigure}
    \vspace{-10px}
    \caption{Statistics of our multi-temporal albedo dataset. (a) Distribution of images per year. (b) Distribution of images per site. (c) Distribution of images per lighting condition.}
    \vspace{-10px}
    \label{fig:dataset_stats}
\end{figure}

\paragraph{Statistics and Coverage}
\cref{fig:dataset_stats} summarizes dataset statistics, including the distribution of images by year, scene type, and weather condition. Our dataset has 5,664 images from 2022 to 2025, covering four landscape types: office, arena, residential, and natural park. Weather conditions include clear-sky sunrise/sunset and overcast days. Compared with synthetic datasets, the scene diversity is naturally limited by real data collection, and the site distribution is correspondingly imbalanced. We therefore use \textit{Olbedo} primarily for fine-tuning rather than training from scratch, with evaluation conducted on the external MatrixCity benchmark to test whether the learned outdoor albedo prior transfers beyond the captured sites.

\subsection{Pseudo-ground-truth Albedo Curation}
Decomposing albedo from images is a central challenge in developing our dataset. Our dataset builds on a multi-view albedo-shading decomposition method that estimates outdoor shading from a single temporal sample. It requires linear radiance images, accurate geometry, precise image poses, and sun position as inputs to separate albedo from shading.

\paragraph{Aerial Image Preprocessing}
We use a DJI Phantom 4 Pro V2.0 with a 1-inch CMOS camera and capture RAW (DNG) only. In-camera JPEGs are \textit{not} used because they apply a tone curve, gamma, and other vendor processing that destroy the linear relationship between the radiance and the pixel values. This is required by the albedo-shading decomposition method\cite{songGeneralAlbedoRecovery2024}. We read DNGs with OpenImageIO configured for RAW output in linear camera space (e.g., \textit{oiio:RawColor=1}, \textit{raw:ColorSpace=Linear}). No JPEG path, tone mapping, or gamma/CRF is applied. All released RGB, albedo, and shading images remain in linear color space. sRGB is used only for figures in the paper and preview thumbnails. The camera poses, intrinsics and high-quality 3D geometry are solved by the commercial photogrammetry software (Bentley iTwin Capture\footnote{\url{https://www.bentley.com/software/itwin-capture/}}), whose calibrated camera geometry yields approximately 3\,cm reconstruction accuracy, corresponding to only a few pixels at our processing resolution. We provide per-image metadata linking sources to converted EXRs including camera intrinsics/poses. The geo-registered images are downsampled 8 times to 683$\times$455 resolution for image decomposition processing. For more details, please refer to Appendix~\ref{supsec:preprocess} of the supplementary material.

\begin{figure}[!htpb]
\centering

\begin{subfigure}[b]{\linewidth}
    \centering
    \includegraphics[width=0.9\linewidth]{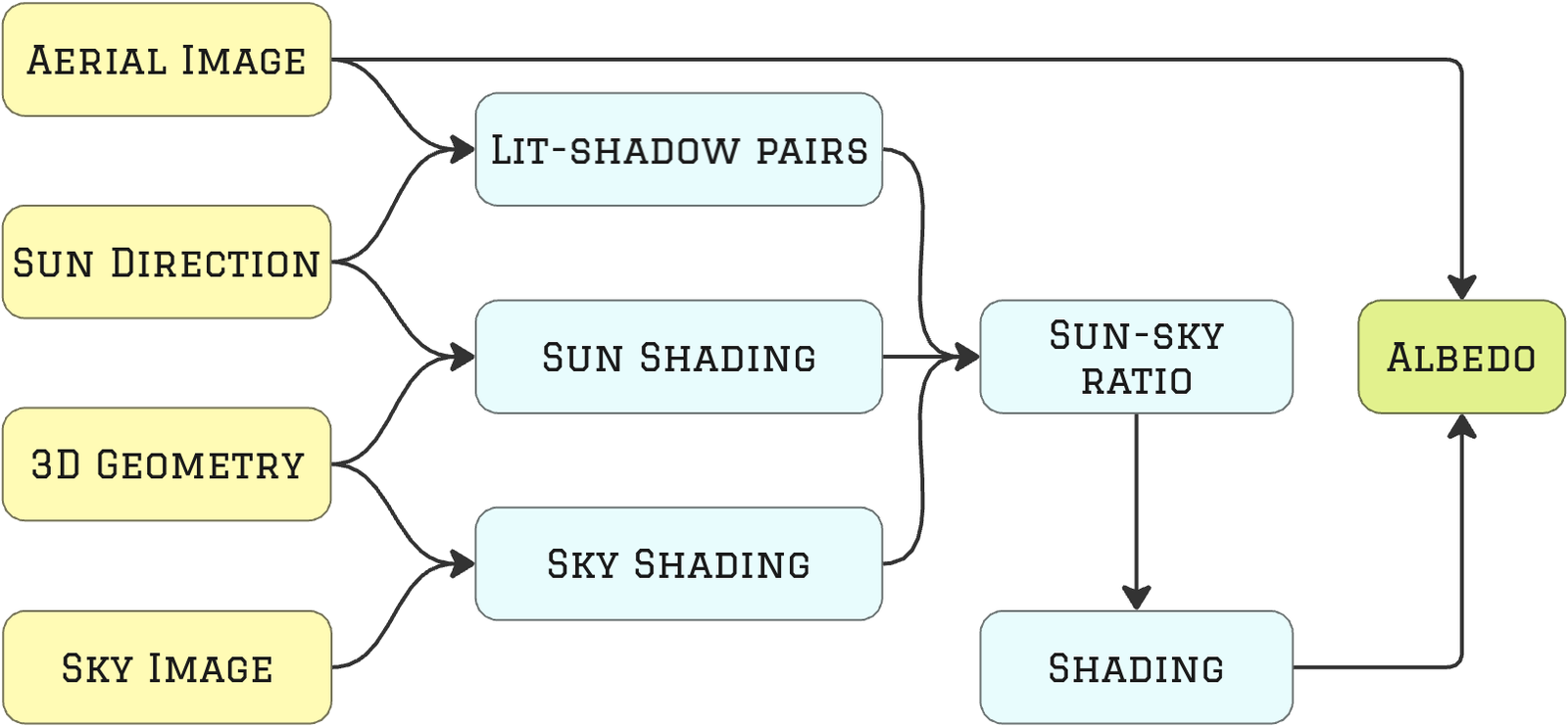}
    \caption{ Yellow boxes represent data sources, blue boxes indicate intermediate processing, and the green box is the final product.}
    \label{fig:method_diagram}
\end{subfigure}
\vskip\baselineskip
\vspace{-10px}
\begin{subfigure}[b]{\linewidth}
    \centering
    \includegraphics[width=0.9\linewidth]{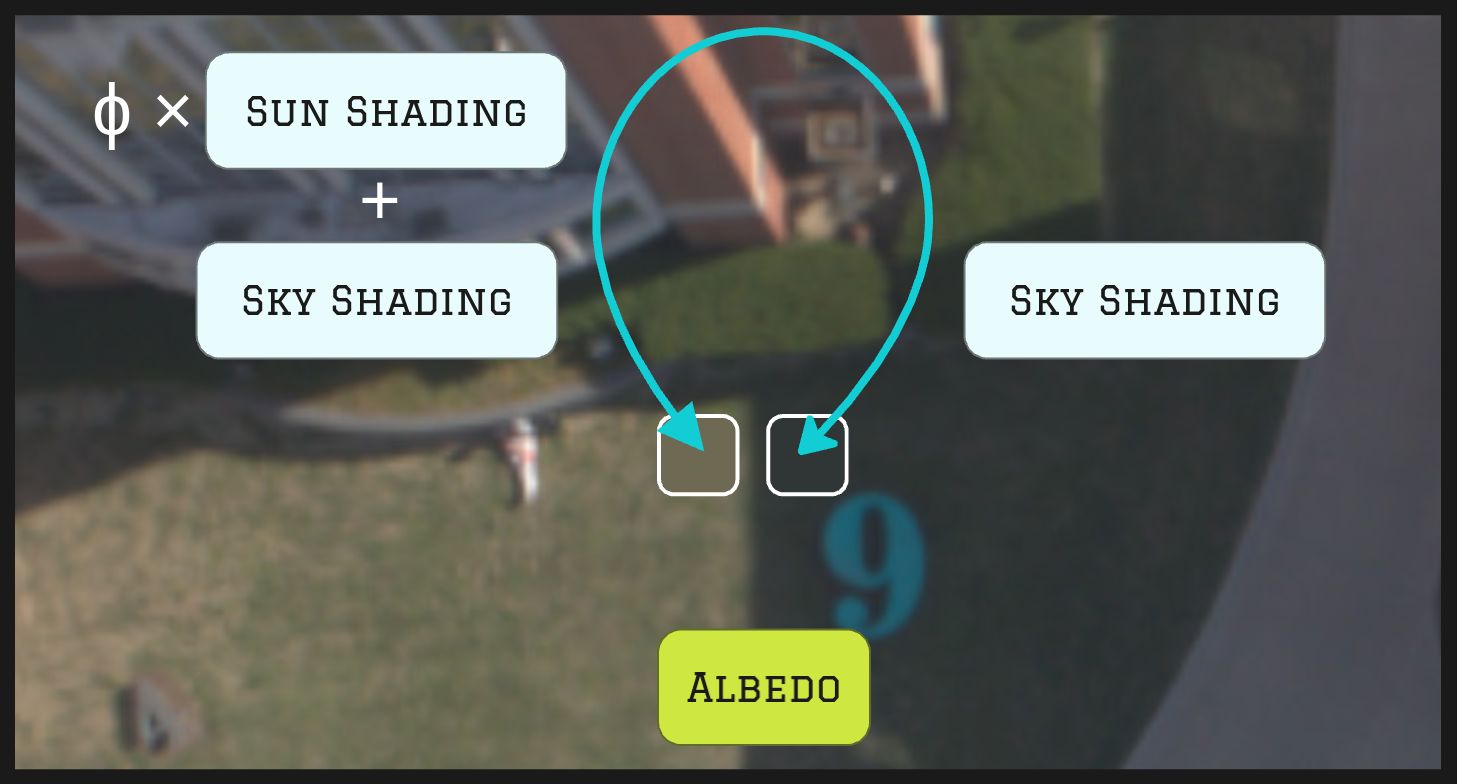}
    \caption{Lit-shadow pair in image, $\phi$ is a ratio between sun and sky shading.}
    \label{fig:lit_shadow_pair}
\end{subfigure}
\vspace{-20px}
\caption{Our image decomposition framework.}
\vspace{-15px}
\label{fig:method_framework}
\end{figure}

\paragraph{Image Decomposition Framework}
Our method follows~\cite{songGeneralAlbedoRecovery2024, songNOVELINTRINSICIMAGE2022}, a physics-based outdoor intrinsic image decomposition approach. The key idea is to model and estimate shading by solving for a sun–sky illumination model. The method relies on pairs of pixels near shadow boundaries that share approximately the same albedo. The image formation model is defined in \cref{eq:image_modeling}:

\begin{equation}
I = R \odot (\phi \odot S_{sun} + S_{sky}),
\label{eq:image_modeling}
\end{equation}
where $I$ is a linear radiance image, $R$ is the albedo (reflectance), and $S_{sun}$ and $S_{sky}$ are the normalized sun and sky shading components, and $\odot$ denotes channel-wise multiplication. These shading terms are derived from geometric features such as surface normals, sun visibility, and sky visibility~\cite{songGeneralAlbedoRecovery2024}. The sun-visibility term, i.e., the cast-shadow map, is also used to find lit–shadow pixel pairs. These pairs are automatically detected and filtered by normal difference, depth difference, and a brightness threshold. As illustrated in \cref{fig:method_framework}, we assume that the albedo is approximately uniform within a local region; thus a lit–shadow pixel pair shares the same albedo. The derivation is shown in \cref{eq:sunlit_pair}. Finally, we estimate the optimal $\phi$, the ratio between sun and sky illumination, from all sunlit pairs in an image using a Gaussian Mixture Model (GMM). The GMM has two components that model the actual distribution of $\phi$ and the noise distribution, both assumed to be Gaussian but with very different variances. For detailed definitions of these two components, please refer to Appendix~\ref{supsec:method} of the supplementary material.

\begin{equation}
\begin{aligned}
\frac{I_{lit} - I_{shadow}}{I_{shadow}} &= \frac{(\phi \odot S_{sun} + S_{sky} - S_{sky}) \odot R}{S_{sky} \odot R}\\
\implies \quad \phi &= \frac{S_{sky} \odot (I_{lit} - I_{shadow})}{S_{sun} \odot I_{shadow}},
\end{aligned}
\label{eq:sunlit_pair}
\end{equation}
where $I_{lit}$ and $I_{shadow}$ are the pixel values of the lit and shadowed regions, respectively. Once $\phi$ is estimated, we compute the albedo with \cref{eq:albedo_from shading}:
\begin{equation}
R = \frac{I}{\phi \odot S_{sun} + S_{sky}}.
\label{eq:albedo_from shading}
\end{equation}

These labels are pseudo-ground-truth produced under the Lambertian-surface and sun--sky-illumination assumptions of the solver; regions with complex reflectance or imperfect geometry, such as specular roofs or glass, can violate this model. Since $\phi$ captures only the sun-to-sky illumination ratio, the recovered albedo is defined up to a global scale, which we fix by normalizing each albedo map by its 98th-percentile intensity while preserving chromaticity. In strongly overcast scenes, valid lit--shadow pairs become insufficient and $\phi$ approaches 0, so we fall back to a sky-only illumination model to avoid numerical instability. Despite these limitations, the resulting labels remain effective for training learning-based albedo--shading models, as shown in \cref{sec:experiments}; representative failure cases are discussed in Appendix~\ref{supsec:failure_cases} of the supplementary material. We also find that the trained model does not strictly require RAW inputs, and the downstream applications in \cref{sec:applications} use linearized sRGB images.

By default (\cref{fig:method_diagram}), we compute sky shading $S_{sky}$ with a \emph{uniform sky lighting} model\cite{songGeneralAlbedoRecovery2024}. When an HDR sky camera is available (our 2025 flights), we replace the uniform dome with the measured sky radiance map to capture subtle anisotropy (e.g., horizon brightening, partial clouds). Because the measured sky radiance maps are not aligned with the estimated shading terms, we first solve the uniform sky-lighting model for each flight to establish a global scale, then align the measured sky radiance to this model via a least-squares gain before evaluating $S_{sky}'$. This preserves the decomposition scale while injecting real sky detail where available.

\paragraph{Confidence Masks}
To facilitate network training, we curate per-pixel confidence masks for the decomposed albedo from dilated geometry boundaries and cast-shadow boundaries, where the decomposition is most unstable. During fine-tuning, supervision is restricted to unmasked regions, preventing unreliable pseudo-ground-truth near geometry holes and shadow edges from overriding the pretrained models' existing priors. We additionally retain a manually screened subset of 2.49k relatively clean images with fewer decomposition artifacts. The confidence masks are released alongside the albedo and shading maps in our dataset; examples of the masks and representative failure cases are shown in Appendices~\ref{supsec:confidence_mask} and \ref{supsec:failure_cases} of the supplementary material.

\paragraph{Validation}
Validating ground-truth is essential for building a reliable dataset. However, measuring ground-truth albedo, especially outdoors and from UAV viewpoints, is very challenging. Therefore, we validate our decomposed albedo from two aspects: (1) we fine-tune recent albedo-shading decomposition networks on our dataset and evaluate them on synthetic datasets with ground-truth albedo to demonstrate training effectiveness (see \cref{sec:experiments}); and (2) we present downstream applications using the decomposed albedo (e.g., relighting, texture editing, segmentation, and change detection) in \cref{sec:applications}, with additional results in the supplementary material, to demonstrate the capability of models trained on our dataset.

\section{Experiments}\label{sec:experiments}
In this section, we evaluate the effectiveness of the \textit{Olbedo} dataset for single-view outdoor albedo estimation. To this end, we fine-tune state-of-the-art IID methods on the \textit{Olbedo} dataset and compare their performance with that of the corresponding pretrained models.

\begin{figure*}[!htbp]
    \centering
    \includegraphics[width=0.85\linewidth]{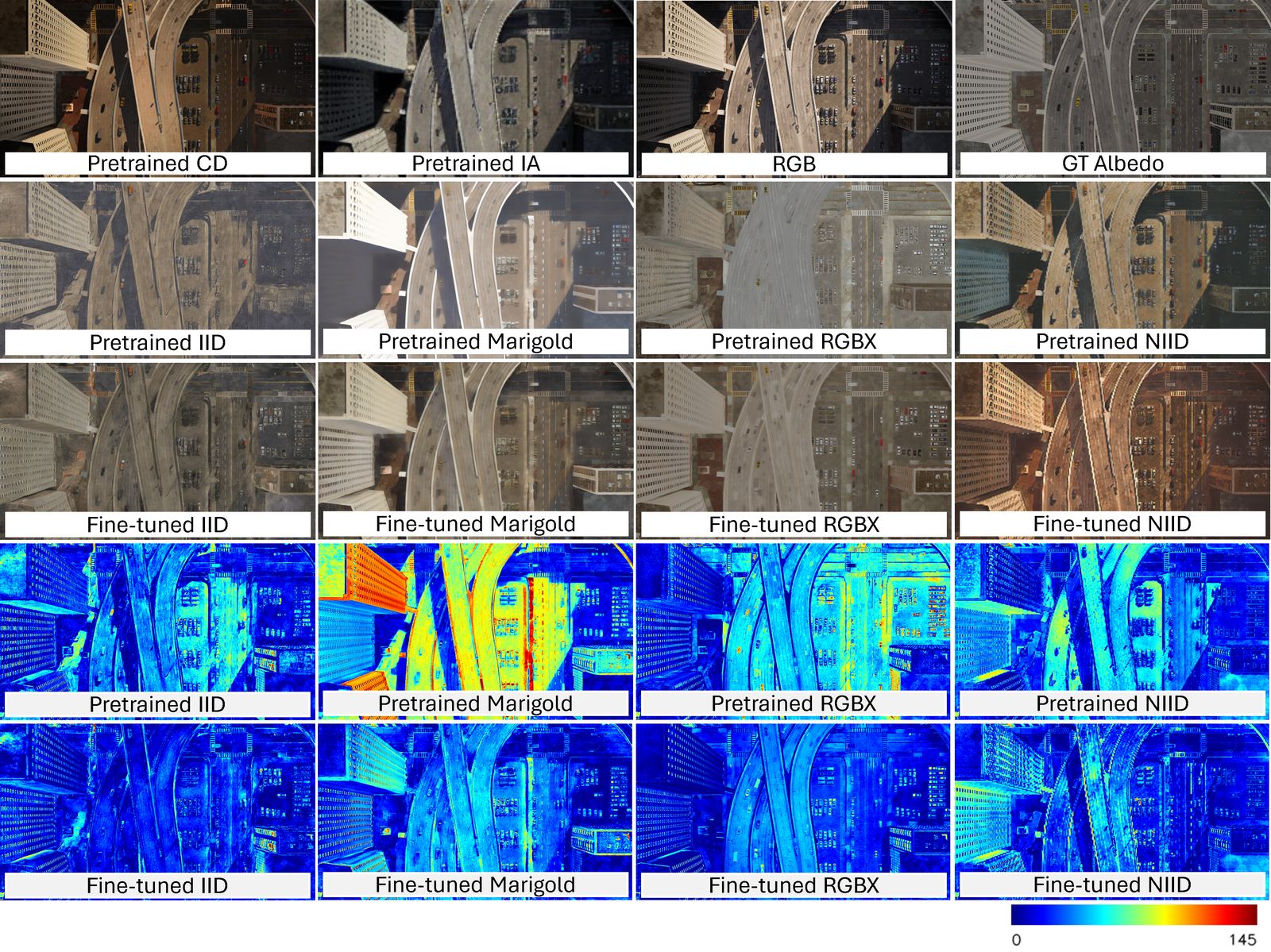}
    \vspace{-10px}
    \caption{Qualitative comparison of fine-tuned and pretrained models for outdoor albedo prediction on the MatrixCity dataset. The first three rows show original image and albedo, and the last two rows show per-pixel absolute error maps for the fine-tunable networks only, where warmer colors indicate larger error and cooler colors indicate smaller error. CD = Colorful Diffuse~\cite{careaga2024colorful}, IA = IntrinsicAnything~\cite{chenIntrinsicAnythingLearningDiffusion2024}, IID = Intrinsic Image Diffusion~\cite{Kocsis_2024_CVPR}, Marigold = Marigold-IID-Appearance~\cite{keMarigoldAffordableAdaptation2025}, RGBX = RGB$\leftrightarrow$X~\cite{zengRGB-XImageDecomposition2024}, and NIID = NIID-Net~\cite{luoNIIDNetAdaptingSurface2020}.}
    \label{fig:matrixcity}
    \vspace{-10px}
\end{figure*}

\subsection{Baselines} 
Recent IID methods span both diffusion-based foundation models and feed-forward intrinsic decomposition networks. To evaluate \textit{Olbedo} across representative methods with different architectures and training domains, we consider six baselines: 
Colorful Diffuse~\cite{careaga2024colorful}, IntrinsicAnything~\cite{chenIntrinsicAnythingLearningDiffusion2024}, 
Intrinsic Image Diffusion~\cite{Kocsis_2024_CVPR}, Marigold-IID-Appearance~\cite{keMarigoldAffordableAdaptation2025}, RGB$\leftrightarrow$X~\cite{zengRGB-XImageDecomposition2024}, and NIID-Net~\cite{luoNIIDNetAdaptingSurface2020}. Intrinsic Image Diffusion and Marigold-IID-Appearance are trained on the synthetic InteriorVerse dataset~\cite{zhuLearningbasedInverseRendering2022}, whereas RGB$\leftrightarrow$X is further trained on additional synthetic data from Hypersim~\cite{Roberts_2021_ICCV} and self-constructed Evermotion datasets, which contain heterogeneous intrinsic properties; it also uses text conditioning to select the target modality. IntrinsicAnything learns diffusion priors for inverse rendering under unknown illumination, NIID-Net is a CNN-based intrinsic decomposition model that leverages surface-normal cues, and Colorful Diffuse targets intrinsic decomposition in the wild. For methods with released training code, namely Intrinsic Image Diffusion, RGB$\leftrightarrow$X, Marigold-IID-Appearance, and NIID-Net, we fine-tune them on \textit{Olbedo} and compare them against their corresponding pretrained versions. For IntrinsicAnything and Colorful Diffuse, whose training code is unavailable, we evaluate the official pretrained checkpoints.


\subsection{Fine-tuning using \textit{Olbedo}}
Among the selected baselines, we fine-tune Intrinsic Image Diffusion, RGB$\leftrightarrow$X, Marigold-IID-Appearance, and NIID-Net for outdoor albedo estimation using the entire \textit{Olbedo} dataset, while IntrinsicAnything and Colorful Diffuse are evaluated only in their pretrained form. We do not reserve an in-domain testing split because \textit{Olbedo} is intended as a training resource rather than an in-domain benchmark, and in-domain evaluation would overstate the precision of pseudo-ground-truth labels. For the diffusion-based models, the original networks are trained on synthetic indoor datasets containing multiple modalities, so we specifically fine-tune the channel corresponding to albedo prediction. For Intrinsic Image Diffusion and Marigold-IID-Appearance, we retain only the albedo channel during fine-tuning. In the case of RGB$\leftrightarrow$X, we fix the text prompt to “albedo” to guide the model toward the desired output.

Given the relatively small size of the \textit{Olbedo} dataset compared to the synthetic indoor datasets used by the diffusion-based baselines, we adopt Low-Rank Adaptation (LoRA)~\cite{hu2022lora} for fine-tuning. This choice matches the role of \textit{Olbedo}: rather than replacing synthetic pretraining, it injects real outdoor lighting diversity through lightweight adaptation.
We employ LoRA instead of full fine-tuning to mitigate the risk of overfitting and to reduce both memory consumption and computational cost. This approach preserves the generalizability of the pretrained model while enabling it to learn the new task of single-view outdoor albedo prediction. 

Specifically, for the diffusion-based models, LoRA is applied to the query, key, value, and output projection matrices within the attention modules of the U-Net. The number of trainable parameters for each diffusion model is approximately 1.66 M, representing only 0.19\% of the total U-Net parameters. For Intrinsic Image Diffusion, we also fine-tune the custom latent encoder, as we find that the pretrained encoder exhibits limited generalization to outdoor images. During LoRA fine-tuning, we reduce the number of epochs to three and set the learning rate to 5e-6 for Intrinsic Image Diffusion and Marigold-IID-Appearance, and 5e-7 for RGB$\leftrightarrow$X. NIID-Net is fine-tuned using its original architecture and training pipeline. All other training settings are adopted from the respective original models.

\subsection{Evaluation}\label{sec:evaluation}
\paragraph{Evaluation Datasets}
We evaluate both the fine-tuned and pretrained models on the MatrixCity benchmark~\cite{liMatrixCityLargescaleCity2023}. This large-scale synthetic outdoor dataset enables the rendering of various scene properties, including depth, normals, and albedo, within city-scale environments. We use MatrixCity to assess whether the fine-tuned models have learned the albedo and shading priors for outdoor scenarios from our \textit{Olbedo} dataset. Using the provided plugin, we capture RGB images under different lighting and shading conditions and generate the corresponding albedo images as ground-truth. The evaluation dataset comprises a total of 520 images at a resolution of $1920 \times 1080$, covering a wide range of scene layouts and lighting variations, thereby providing a representative test set.

\paragraph{Evaluation Protocol}
We use the original inference settings of each method to predict albedo for both the fine-tuned and pretrained models. Due to GPU memory limitations, images are resized so that the longer edge is 1000 pixels for all models, while preserving the aspect ratio. Following the evaluation protocols in~\cite{keMarigoldAffordableAdaptation2025,zengRGB-XImageDecomposition2024}, the predicted albedo is resized back to the original resolution and compared directly with the ground-truth without any alignment. The reported evaluation metrics include Peak Signal-to-Noise Ratio (PSNR), Structural Similarity Index Measure (SSIM)~\cite{wangImageQualityAssessment2004}, and Learned Perceptual Image Patch Similarity (LPIPS)~\cite{zhangUnreasonableEffectivenessDeep2018}.

\begin{table}[ht]
    \centering
    \caption{Quantitative comparison of fine-tuned and pretrained models for outdoor albedo prediction on the MatrixCity dataset. IA = IntrinsicAnything~\cite{chenIntrinsicAnythingLearningDiffusion2024}, CD = Colorful Diffuse~\cite{careaga2024colorful}, NIID = NIID-Net~\cite{luoNIIDNetAdaptingSurface2020}, IID = Intrinsic Image Diffusion~\cite{Kocsis_2024_CVPR}, Marigold = Marigold-IID-Appearance~\cite{keMarigoldAffordableAdaptation2025}, and RGBX = RGB$\leftrightarrow$X~\cite{zengRGB-XImageDecomposition2024}.}
    \vspace{-8px}
    \label{tab:quantitative_evaluation}
    \begin{tabular}{l|c c c}
        \hline
        \textbf{Model} & \textbf{PSNR$\uparrow$} & \textbf{SSIM$\uparrow$} & \textbf{LPIPS$\downarrow$} \\
        \hline
        Pretrained IA & $13.422$ & $0.478$ & $0.910$ \\
        Pretrained CD & $14.905$ & $0.545$ & $0.554$ \\
        \hline
        Pretrained NIID & $12.782$ & $0.549$ & $0.793$ \\
        Fine-tuned NIID & $\textbf{16.152}$ & $\textbf{0.594}$ & $\textbf{0.769}$ \\
        \hline
        Pretrained IID & $15.594$ & $0.493$ & $0.554$ \\
        Fine-tuned IID & $\textbf{17.249}$ & $\textbf{0.531}$ & $\textbf{0.485}$ \\
        \hline
        Pretrained Marigold & $10.153$ & $0.508$ & $0.591$ \\
        Fine-tuned Marigold & $\textbf{17.118}$ & $\textbf{0.570}$ & $\textbf{0.461}$ \\
        \hline
        Pretrained RGBX & $15.050$ & $0.559$ & $0.472$ \\
        Fine-tuned RGBX & $\textbf{17.735}$ & $\textbf{0.611}$ & $\textbf{0.413}$ \\
        \hline
    \end{tabular}
    \vspace{-10px}
\end{table}

\paragraph{Quantitative Evaluation}
As shown in \cref{tab:quantitative_evaluation}, the methods fine-tuned on \textit{Olbedo} achieve consistent improvements across all evaluation metrics. NIID-Net improves from 12.782 to 16.152 in PSNR, from 0.549 to 0.594 in SSIM, and from 0.793 to 0.769 in LPIPS. Intrinsic Image Diffusion achieves PSNR, SSIM, and LPIPS improvements of 10.6\%, 7.7\%, and 12.5\%, respectively. Marigold-IID-Appearance shows the largest gain in PSNR, increasing by 68.6\%, together with a 12.2\% increase in SSIM and a 22.0\% decrease in LPIPS. For RGB$\leftrightarrow$X, PSNR increases by 17.8\%, SSIM by 9.3\%, and LPIPS decreases by 12.5\%, yielding the best overall performance among all compared methods. Although IntrinsicAnything and Colorful Diffuse are evaluated only with their released pretrained checkpoints, they provide additional reference points for cross-method comparison. Overall, these results demonstrate that fine-tuning on \textit{Olbedo}, even with a small dataset and lightweight LoRA adaptation, substantially improves both the pixel-level accuracy and perceptual quality of single-view outdoor albedo predictions.

\paragraph{Qualitative Evaluation}
As shown in \cref{fig:matrixcity}, the pretrained models generally exhibit residual shading artifacts, desaturated colors, or flattened textures when applied to outdoor scenes. Among the pretrained methods, RGB$\leftrightarrow$X and Colorful Diffuse produce relatively cleaner shading separation, but they still miss fine structural details and often introduce indoor-style appearance biases. After fine-tuning on \textit{Olbedo}, NIID-Net, Intrinsic Image Diffusion, Marigold-IID-Appearance, and RGB$\leftrightarrow$X all show clear visual improvements, with reduced shading leakage, colors that better match the ground-truth albedo, and sharper recovery of scene details such as vehicles, lane markings, and building facades. The absolute error maps further support this trend: red and yellow regions denote larger deviations from the ground-truth, whereas blue regions indicate smaller error. Before fine-tuning, the error is concentrated on large planar structures such as the elevated roads, bright building facades, and parking-lot surfaces, indicating that the pretrained models still confuse reflectance with illumination in outdoor scenes. After fine-tuning, these high-error regions are markedly suppressed and the maps become predominantly blue, especially on roads, facades, and fine structures such as parked vehicles and lane markings. Overall, the fine-tuned RGB$\leftrightarrow$X achieves the best qualitative performance with the most uniformly low error, consistent with the quantitative results reported in \cref{tab:quantitative_evaluation}.

\section{Applications} \label{sec:applications}
We present four representative applications enabled by outdoor albedo prediction. Unless noted otherwise, we use the fine-tuned RGB$\leftrightarrow$X model, which performs best in \cref{sec:evaluation}.

\begin{figure}[!ht]
    \centering
    \vspace{-10px}
    \includegraphics[width=\linewidth]{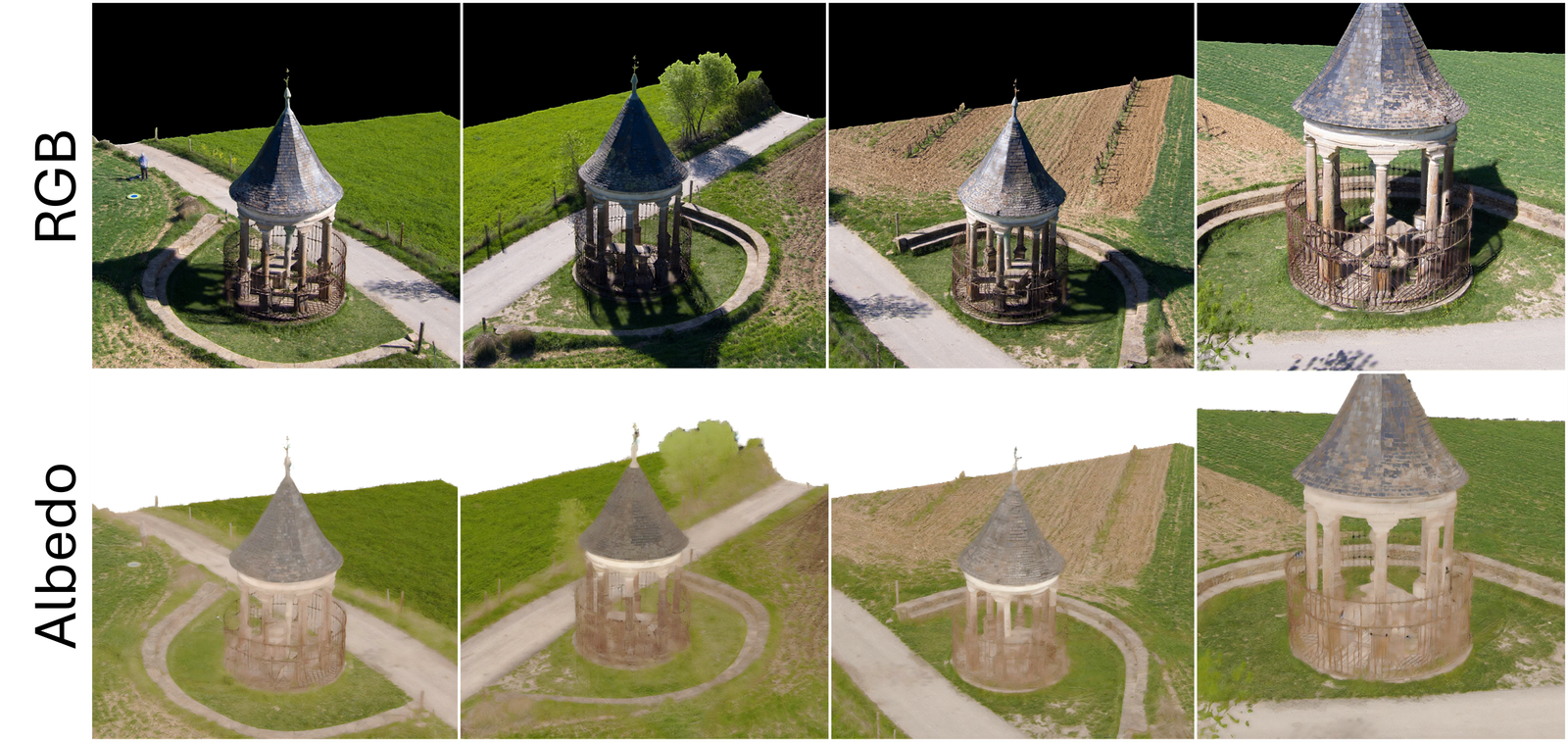}
    \vspace{-20px}
    \caption{Predicted albedo on a BlendedMVS scene from the fine-tuned RGB$\leftrightarrow$X model. Across viewpoints, baked-in cast shadows and shading gradients are largely removed while roofs, walls, and roads retain a consistent appearance, making the results suitable for multi-view retexturing.}
    \label{fig:wexner_blendedmvs}
    \vspace{-10px}
\end{figure}

\paragraph{Retexturing and Relighting}
Outdoor digital-twin capture is often performed under overcast conditions to avoid strong baked-in shadows~\cite{linCapturingReconstructingSimulating2022, luo2020deeply}. Our predicted albedo removes much of this shading while remaining view-consistent on BlendedMVS~\cite{yaoBlendedMVSLargescaleDataset2020}, as shown in \cref{fig:wexner_blendedmvs}. This allows the recovered albedo to be retextured onto reconstructed models and then relit under novel illumination. We render the retextured scenes in a commercial renderer\footnote{\url{https://www.d5render.com/}}; compared with RGB textures, albedo textures yield shadows and shading that better match the new sun direction and weather, as shown in Appendices~\ref{supsec:retexturing} and \ref{supsec:relighting}.


\begin{figure*}[!ht]
\centering
    \vspace{-10px}
    \setlength{\tabcolsep}{0pt}
    \renewcommand{\arraystretch}{1.1}

    \begin{tabular}{
        >{\centering\arraybackslash}p{0.16666\linewidth}
        >{\centering\arraybackslash}p{0.16666\linewidth}
        >{\centering\arraybackslash}p{0.16666\linewidth}
        >{\centering\arraybackslash}p{0.16666\linewidth}
        >{\centering\arraybackslash}p{0.16666\linewidth}
        >{\centering\arraybackslash}p{0.16666\linewidth}
    }
        \multicolumn{6}{c}{
            \adjustbox{trim=0 {0.05\height} 0 {0.65\height}, clip}%
            {\includegraphics[width=\linewidth]{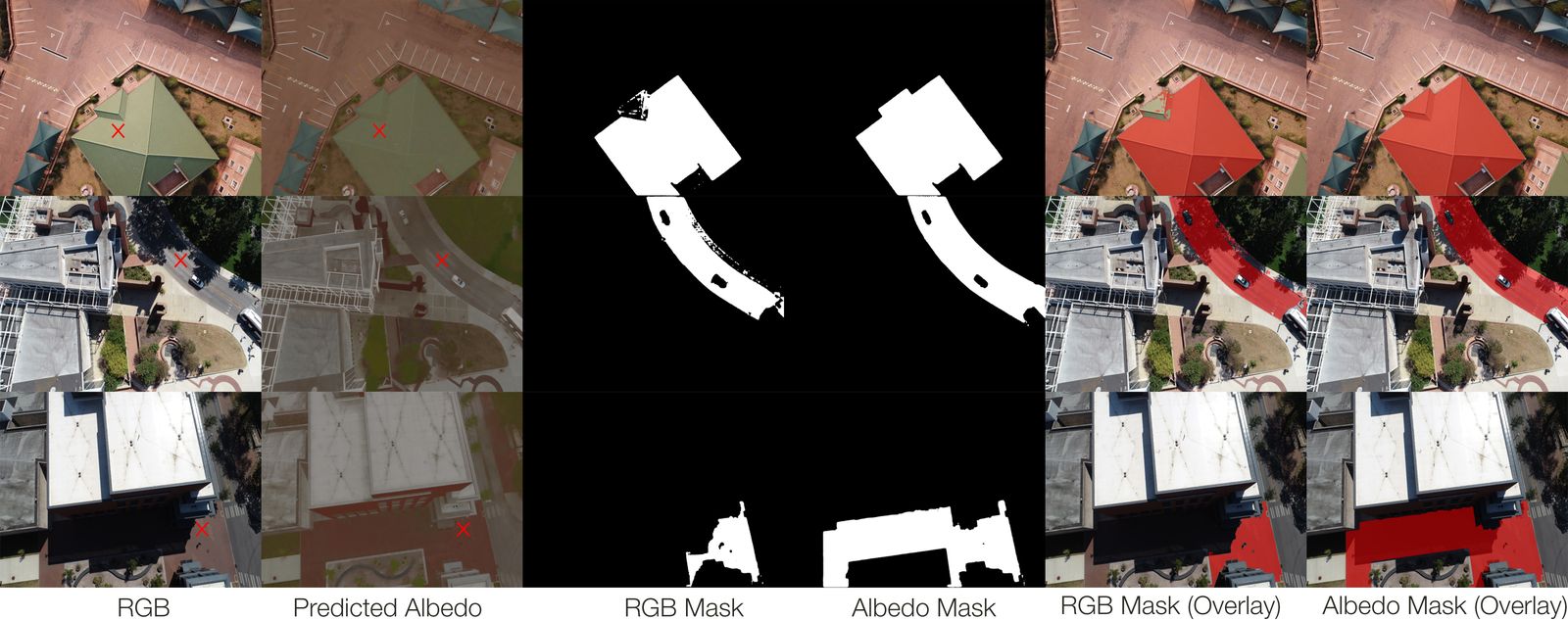}}
        } \\
        Original & Albedo & Original Mask & Albedo Mask & Original Overlay & Albedo Overlay \\
    \end{tabular}
    \vspace{-10px}
    \caption{Segmentation with SAM on RGB and albedo using the same point of interest. From left to right, the columns show the RGB image, predicted albedo, SAM mask from RGB, SAM mask from albedo, and the two overlays. Albedo suppresses shadow-induced edges and yields more coherent masks, especially when cast shadows interrupt object continuity.}
    \label{fig:SAM_comp}
    \vspace{-10px}
\end{figure*}

\begin{figure*}[!ht]
    \centering
    \setlength{\tabcolsep}{0pt}
    \renewcommand{\arraystretch}{1.1}

    \begin{tabular}{
        >{\centering\arraybackslash}p{0.16666\linewidth}
        >{\centering\arraybackslash}p{0.16666\linewidth}
        >{\centering\arraybackslash}p{0.16666\linewidth}
        >{\centering\arraybackslash}p{0.16666\linewidth}
        >{\centering\arraybackslash}p{0.16666\linewidth}
        >{\centering\arraybackslash}p{0.16666\linewidth}
    }
        \multicolumn{6}{c}{
            \adjustbox{trim=0 {0.05\height} 0 {0.35\height}, clip}%
            {\includegraphics[width=\linewidth]{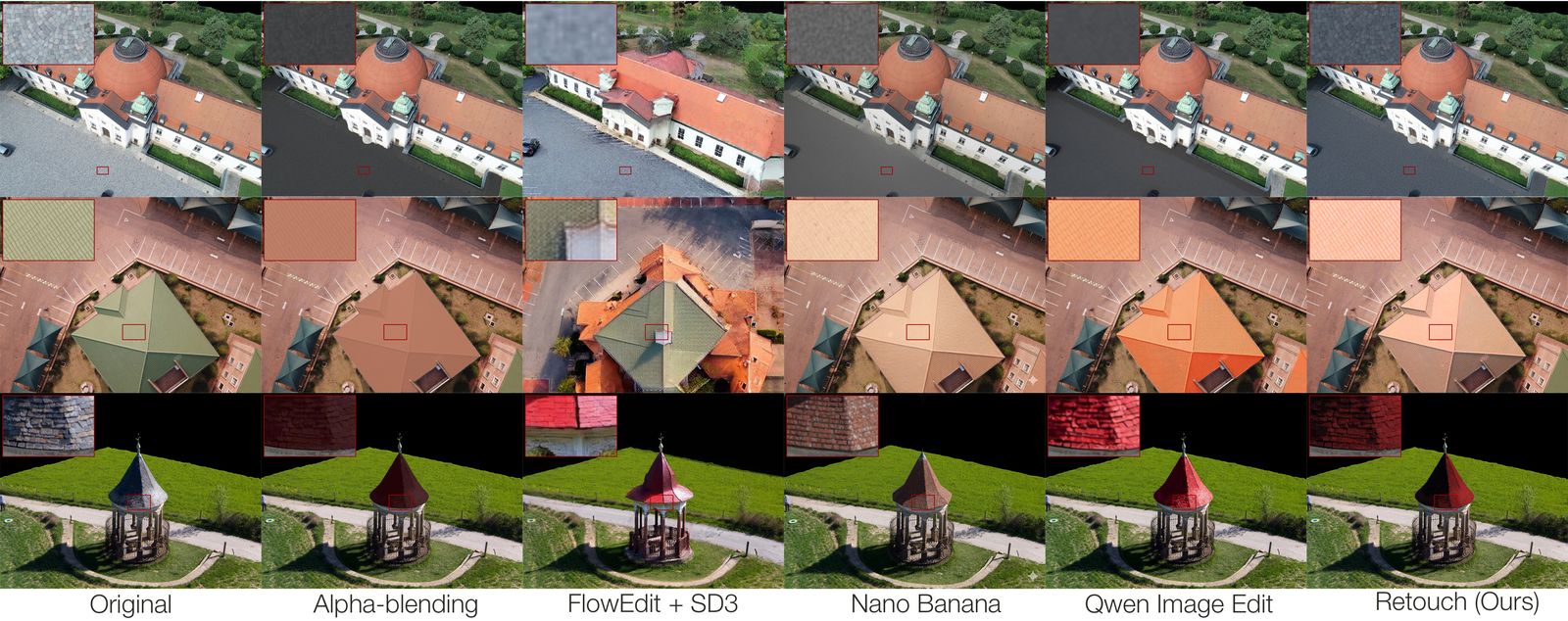}}
        } \\
        Original & Alpha-blending & FlowEdit+SD3 & Nano Banana & Qwen Image Edit & Ours
    \end{tabular}
    \vspace{-10px}
    \caption{Material editing comparison on an aerial scene. The target colors are \textcolor[rgb]{0.90,0.75,0.56}{\rule{1em}{0.6em}} (RGB: 190, 115, 92), \textcolor[rgb]{0.5,0,0}{\rule{1em}{0.6em}} (RGB: 40, 0, 0), respectively. Alpha blending flattens the original shading, and AI editing baselines often alter illumination or wash out roof details, whereas our intrinsic-decomposition workflow preserves both the original lighting and local texture.}
    \label{fig:retouch_comparison}
    \vspace{-10px}
\end{figure*}

\paragraph{Segmentation on Albedo Image}
Shadows in RGB introduce artificial edges that can mislead segmentation. Using SAM~\cite{raviSAM2Segment2024} with the same point of interest on RGB and albedo images, we observe more stable masks and better object continuity on albedo, especially when cast shadows cover a large portion of the target (\cref{fig:SAM_comp}). Additional results are provided in Appendix~\ref{supsec:segmentation}.

\paragraph{Material Editing}
We edit predicted albedo and then recombine it with shading estimated by inverse Retinex, $S = \frac{I}{R}$, to preserve the original illumination. Compared with alpha blending, FlowEdit+SD3~\cite{kulikov2025flowedit,rombach2021highresolution}, Nano Banana~\cite{ye2025echo4o}, and Qwen Image Edit~\cite{wu2025qwenimagetechnicalreport}, our method better preserves both lighting and fine texture details in \cref{fig:retouch_comparison}. 

\begin{figure*}[!htbp]
    \centering
    
    \setlength{\tabcolsep}{0.5pt}
    \renewcommand{\arraystretch}{1.0}
    \begin{tabular}{cccccc}
        \includegraphics[width=0.155\linewidth]{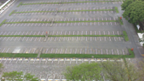} &
        \includegraphics[width=0.155\linewidth]{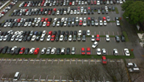} &
        \includegraphics[width=0.155\linewidth]{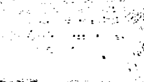} &
        \includegraphics[width=0.155\linewidth]{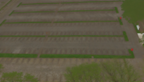} &
        \includegraphics[width=0.155\linewidth]{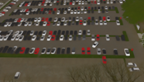} &
        \includegraphics[width=0.155\linewidth]{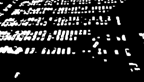} \\
        \includegraphics[width=0.155\linewidth]{assets/change_detection/reference.png} &
        \includegraphics[width=0.155\linewidth]{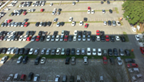} &
        \includegraphics[width=0.155\linewidth]{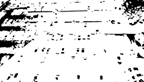} &
        \includegraphics[width=0.155\linewidth]{assets/change_detection/reference_albedo.png} &
        \includegraphics[width=0.155\linewidth]{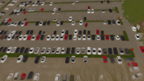} &
        \includegraphics[width=0.155\linewidth]{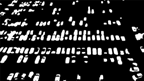} \\
        \includegraphics[width=0.155\linewidth]{assets/change_detection/reference.png} &
        \includegraphics[width=0.155\linewidth]{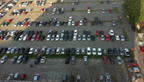} &
        \includegraphics[width=0.155\linewidth]{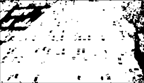} &
        \includegraphics[width=0.155\linewidth]{assets/change_detection/reference_albedo.png} &
        \includegraphics[width=0.155\linewidth]{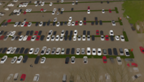} &
        \includegraphics[width=0.155\linewidth]{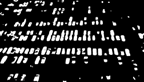} \\
        {RGB reference} & {RGB source} & {RGB diff.} & {Albedo reference} & {Albedo source} & {Albedo diff.}
    \end{tabular}
    \vspace{-10px}
    \caption{Lighting-invariant scene change analysis with RGB differencing and albedo differencing. The left three columns show the RGB reference, RGB source, and naive RGB difference; the right three columns show the corresponding albedo pair and albedo-based difference. Albedo suppresses false positives caused by shadows and illumination changes while preserving true scene changes such as parked vehicles.}
    \label{fig:albedo_change_detection}
    \vspace{-10px}
    
\end{figure*}

\paragraph{Scene Change Analysis with Fixed Cameras}
We further evaluate albedo for lighting-invariant scene change analysis on a parking-lot surveillance dataset captured by a stationary camera~\cite{Hsieh_2017_ICCV}. Starting from a clean reference frame, we compare naive RGB differencing and albedo differencing using the same simple thresholding pipeline. As shown in \cref{fig:albedo_change_detection}, albedo suppresses most false positives caused by cast shadows and time-of-day variation while preserving true scene changes.

\section{Conclusion} \label{sec:conclusion}
We introduce \textit{Olbedo}, a large-scale real-world aerial dataset with dense, multi-view consistent albedo and shading, geometry, camera poses, and HDR sky domes for outdoor intrinsic image decomposition. Fine-tuning recent diffusion IID models with lightweight LoRA on \textit{Olbedo} improves single-view outdoor albedo estimation on MatrixCity (PSNR/SSIM/LPIPS) and yields multi-view consistent predictions that enable relighting, segmentation, material editing, and scene change analysis. Crucially, \textit{Olbedo} complements synthetic datasets: combining \textit{Olbedo} with synthetic training data transfers strong indoor/synthetic priors to real outdoor imagery and significantly boosts generalization, as confirmed by our experiments. We release our dataset, baselines, and evaluation protocol at \url{https://gdaosu.github.io/olbedo/} to support research on outdoor albedo inference and 3D content creation.

{
\small
\section*{Acknowledgements}
This work is supported by the Office of Naval Research (Award No. N000142312670) and Intelligence Advanced Research Projects Activity (IARPA) via Department of Interior/Interior Business Center (DOI/IBC) contract number 140D0423C0075. This work also used High Performance Computing resources provided by the Ohio Supercomputer Center~\cite{Ohio_Supercomputer_Center1987-dl}.

\bibliographystyle{ieeenat_fullname}
\bibliography{main}
}

\clearpage
\appendix
\maketitlesupplementary
\setcounter{section}{0}
\setcounter{figure}{0}
\setcounter{table}{0}
\setcounter{equation}{0}
\renewcommand\thesection{\Alph{section}}
\renewcommand\thesubsection{\thesection.\arabic{subsection}}
\numberwithin{figure}{section}
\numberwithin{table}{section}
\numberwithin{equation}{section}

\section{Image Processing Details} \label{supsec:preprocess}

This section expands the description of our aerial image preprocessing pipeline in the main paper. We detail how RAW images are decoded, converted to linear RGB EXRs, georegistered, and downsampled before albedo--shading decomposition.

\begin{figure}[!htpb]
    \centering
    \begin{subfigure}[b]{0.48\linewidth}
        \centering
        \includegraphics[width=\linewidth]{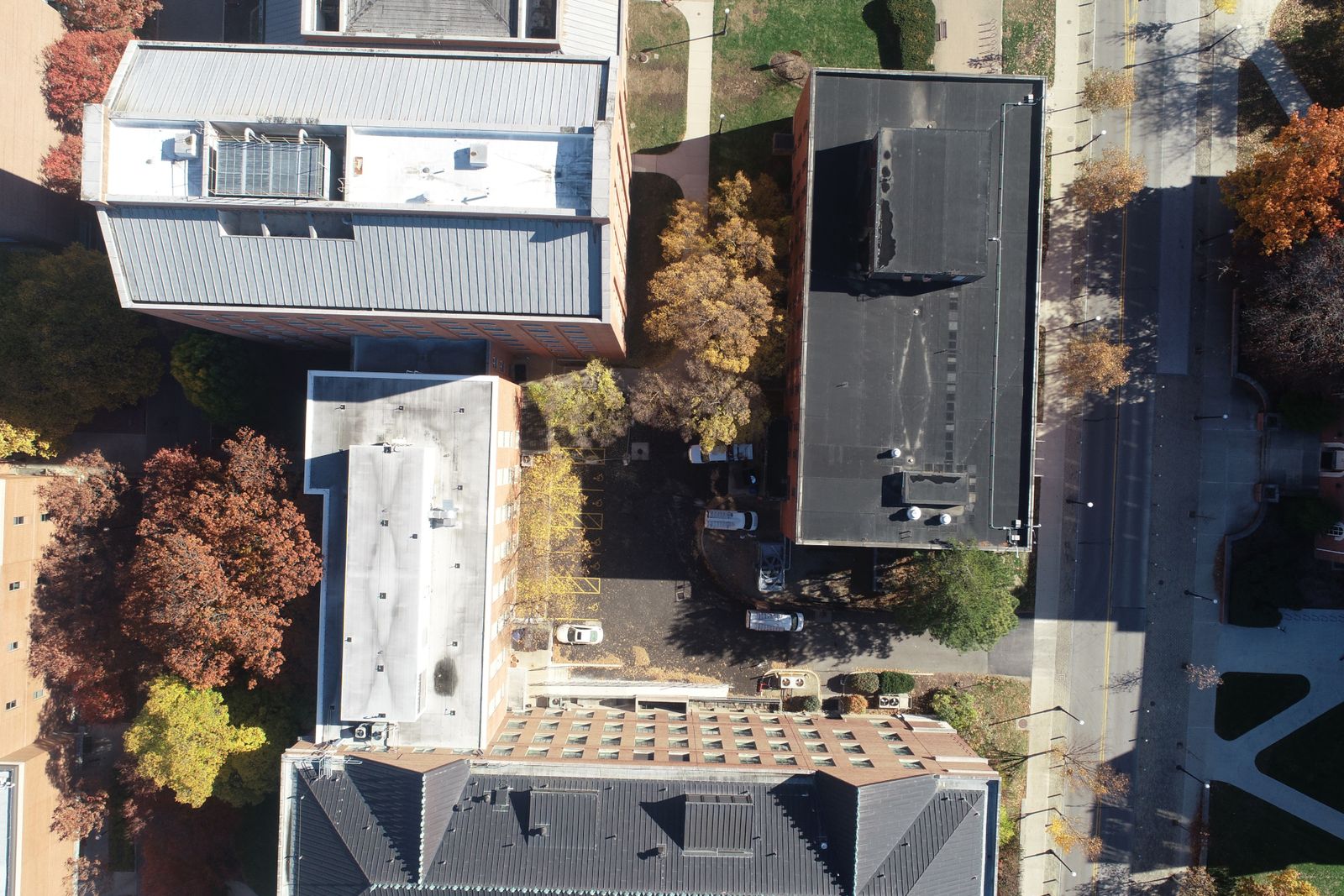}
        \caption{In-camera JPEG}
    \end{subfigure}
    \vspace{-10px}
    \begin{subfigure}[b]{0.48\linewidth}
        \centering
        \includegraphics[width=\linewidth]{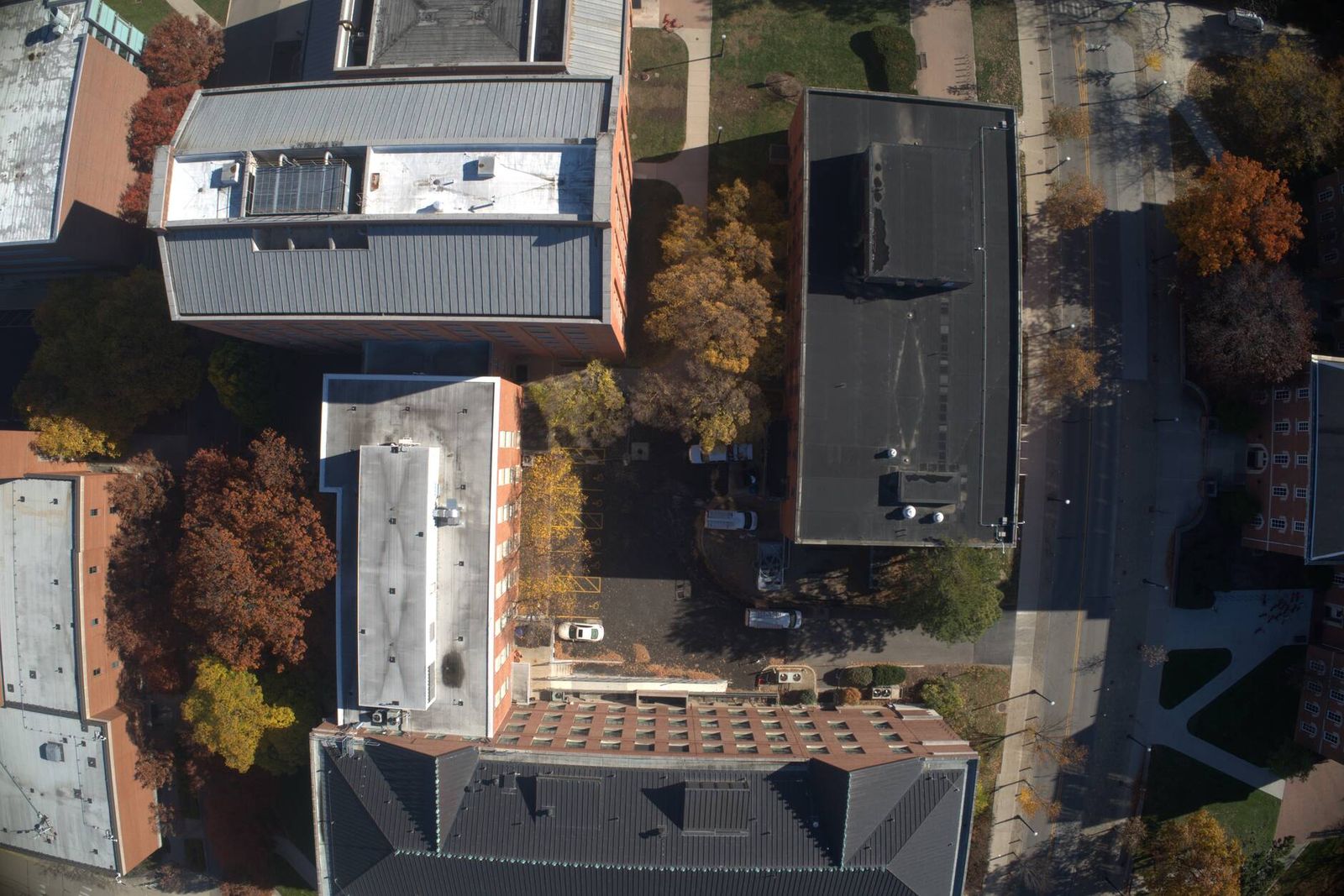}
        \caption{RAW DNG}
    \end{subfigure}

    \caption{Comparison between in-camera JPEG (a) and RAW DNG (b) from the same scene. The JPEG image exhibits visible artifacts such as color clipping in highlights, loss of shadow details, and posterization effects due to aggressive compression and tone mapping. In contrast, the RAW DNG preserves a wider dynamic range and more accurate color representation, which are essential for reliable albedo–shading decomposition.}
    \label{fig:raw_vs_jpeg}
\end{figure}

\paragraph{Acquisition setup}
All aerial images are captured using a DJI Phantom 4 Pro V2.0 with the built-in 1-inch CMOS camera. We record \emph{only} RAW Digital Negative (DNG) files; in-camera JPEGs are not used anywhere in our pipeline. First, JPEGs apply vendor-specific tone curves, camera response functions (CRFs), and other non-linear image signal processing that destroy the linear relationship between scene radiance and pixel values. Second, the JPEG images have been undistorted with the built-in coefficients, which can conflict with the radial–tangential distortion model used in photogrammetry software. Third, RAW DNGs provide a higher bit depth (12 or 14 bits per channel) compared to JPEGs (8 bits per channel), which helps preserve subtle details in shadows and highlights that are crucial for accurate albedo–shading decomposition, as illustrated in \cref{fig:raw_vs_jpeg}.

\paragraph{RAW decoding and linearization}
We decode all DNGs using OpenImageIO (OIIO) configured to return values in the camera’s linear RAW color space:
\begin{itemize}
  \item \texttt{oiio:RawColor = 1} to request RAW sensor-space values without applying DNG color transforms.
  \item \texttt{raw:ColorSpace = Linear} to ensure the output is linear with respect to sensor radiance.
\end{itemize}
No in-camera JPEG pipeline, tone mapping, gamma curve, or camera response function (CRF) is applied at any stage of our preprocessing. We use OIIO’s demosaicing and white-balance handling as provided by its RAW interface, while keeping the output in a linear radiometric scale.

\paragraph{Conversion to EXR}
All RGB, albedo, and shading images that we release are stored as OpenEXR (EXR) files in linear color space:
\begin{itemize}
  \item The RGB EXRs are obtained directly from the RAW decoding described above.
  \item The albedo and shading EXRs are generated by running the intrinsic decomposition method~\cite{songGeneralAlbedoRecovery2024} on the downsampled linear RGB images.
  \item No gamma correction, tone mapping, or sRGB transfer function is baked into these EXRs.
\end{itemize}
sRGB is used only for visualization in the paper (figures) and for low-resolution preview thumbnails. Any user wishing to work in display space should explicitly convert the provided EXRs using a standard sRGB transfer function.

\paragraph{Camera calibration and geometry}
Camera intrinsics, poses, and 3D geometry are estimated using the commercial photogrammetry software Bentley iTwin Capture\footnote{\url{https://www.bentley.com/software/itwin-capture/}}. To handle multiple flights of the same area, we use virtual ground control points to align different captures into a single, georegistered coordinate system. The virtual GCPs are created by manually selecting keypoints on prominent, static features visible across different flights (e.g., building corners, road intersections). iTwin Capture uses these points to optimize camera poses and align the reconstructed models into a common world coordinate system with real-world scale and orientation. A uniform coordinate system allows us to reproject from one image to another using depth maps for cross-view consistency checks and change detection. We provide the calibrated camera intrinsics and extrinsics for each image in the dataset. The reconstructed 3D geometry is exported as a textured mesh in OBJ format.


\begin{figure*}[!ht]
    \centering
    \vspace{-10px}
    \setlength{\tabcolsep}{0.5pt}
    \renewcommand{\arraystretch}{1.0}
    \begin{tabular}{cccccc}
        \includegraphics[width=0.164\linewidth]{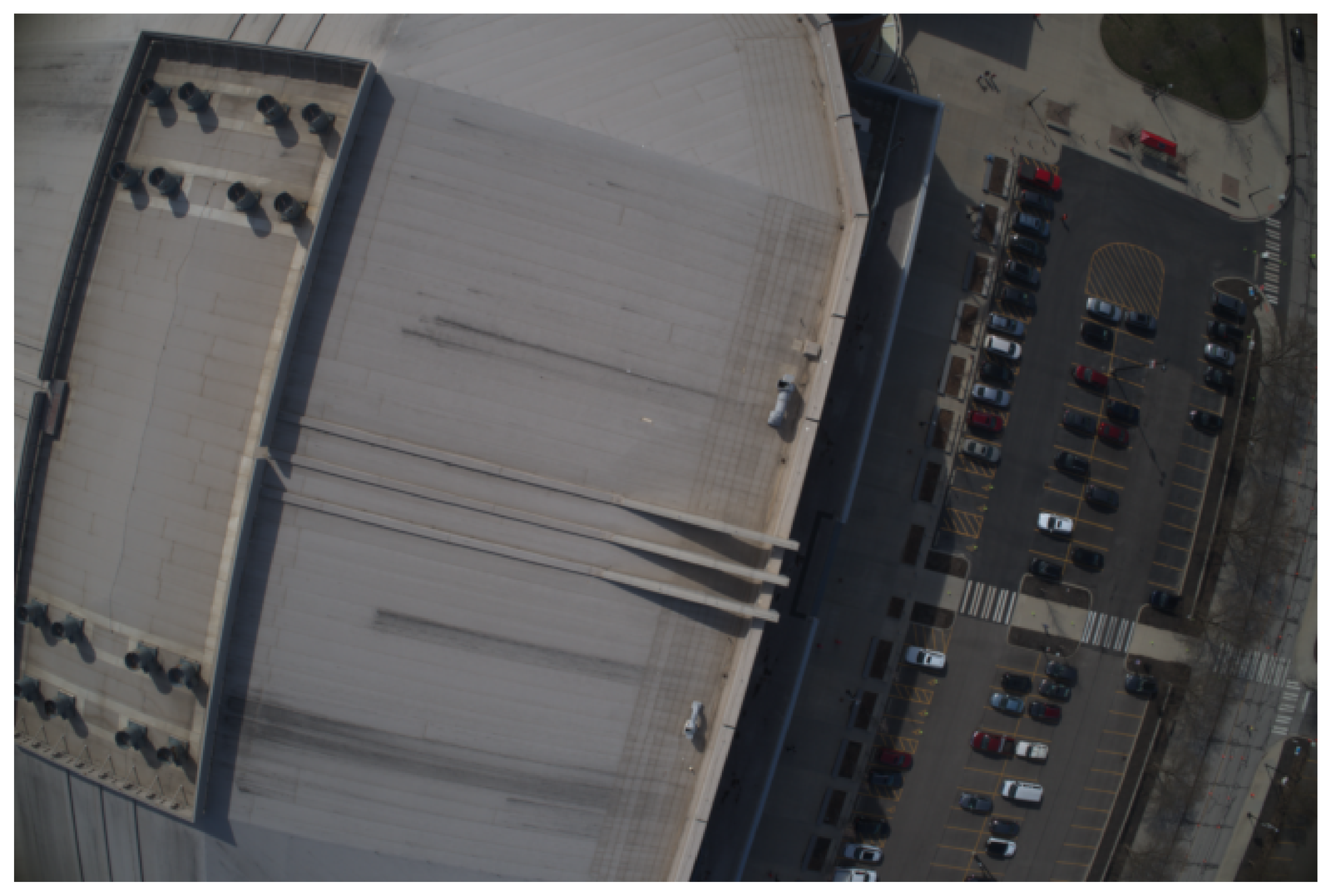} &
        \includegraphics[width=0.164\linewidth]{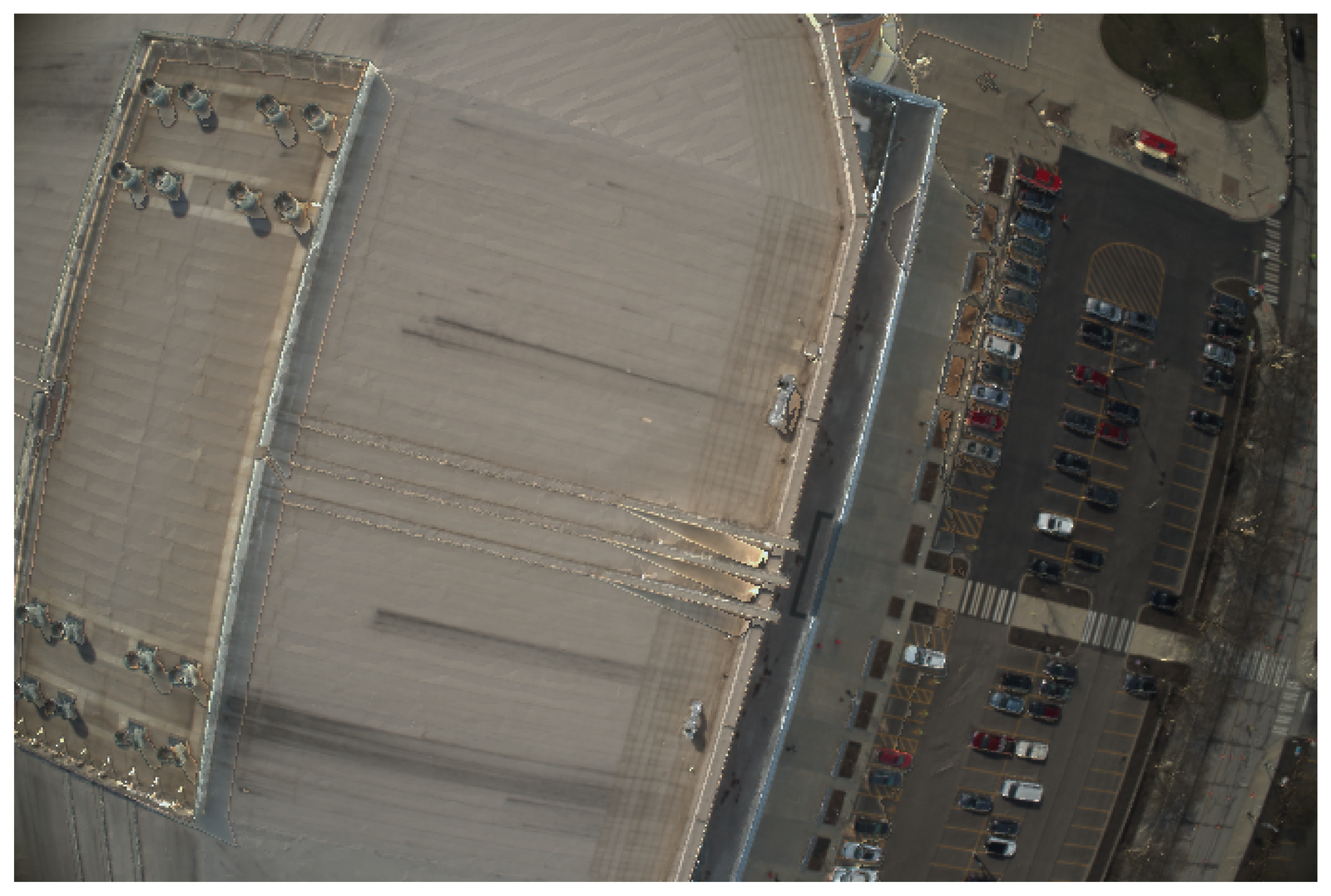} &
        \includegraphics[width=0.164\linewidth]{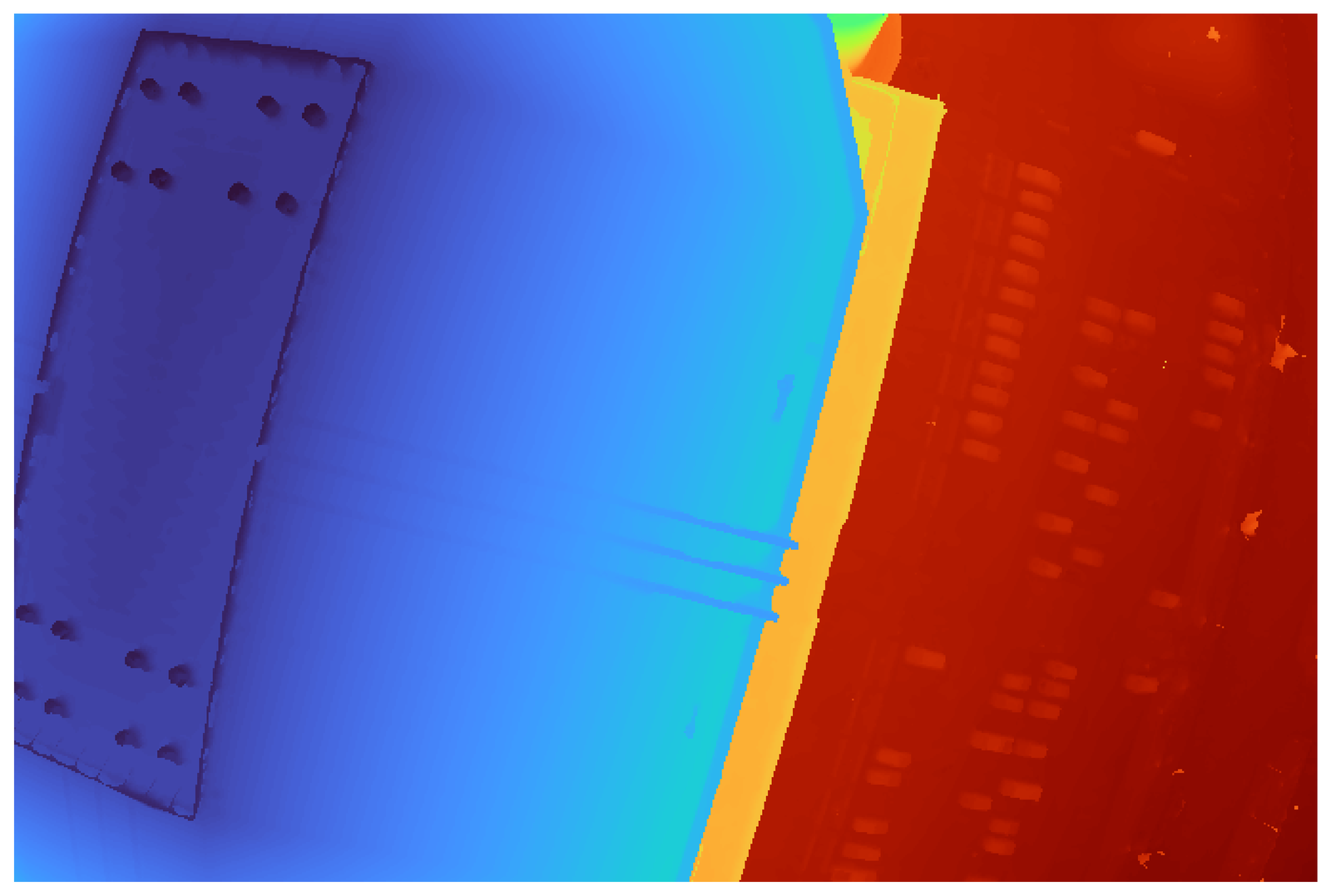} &
        \includegraphics[width=0.164\linewidth]{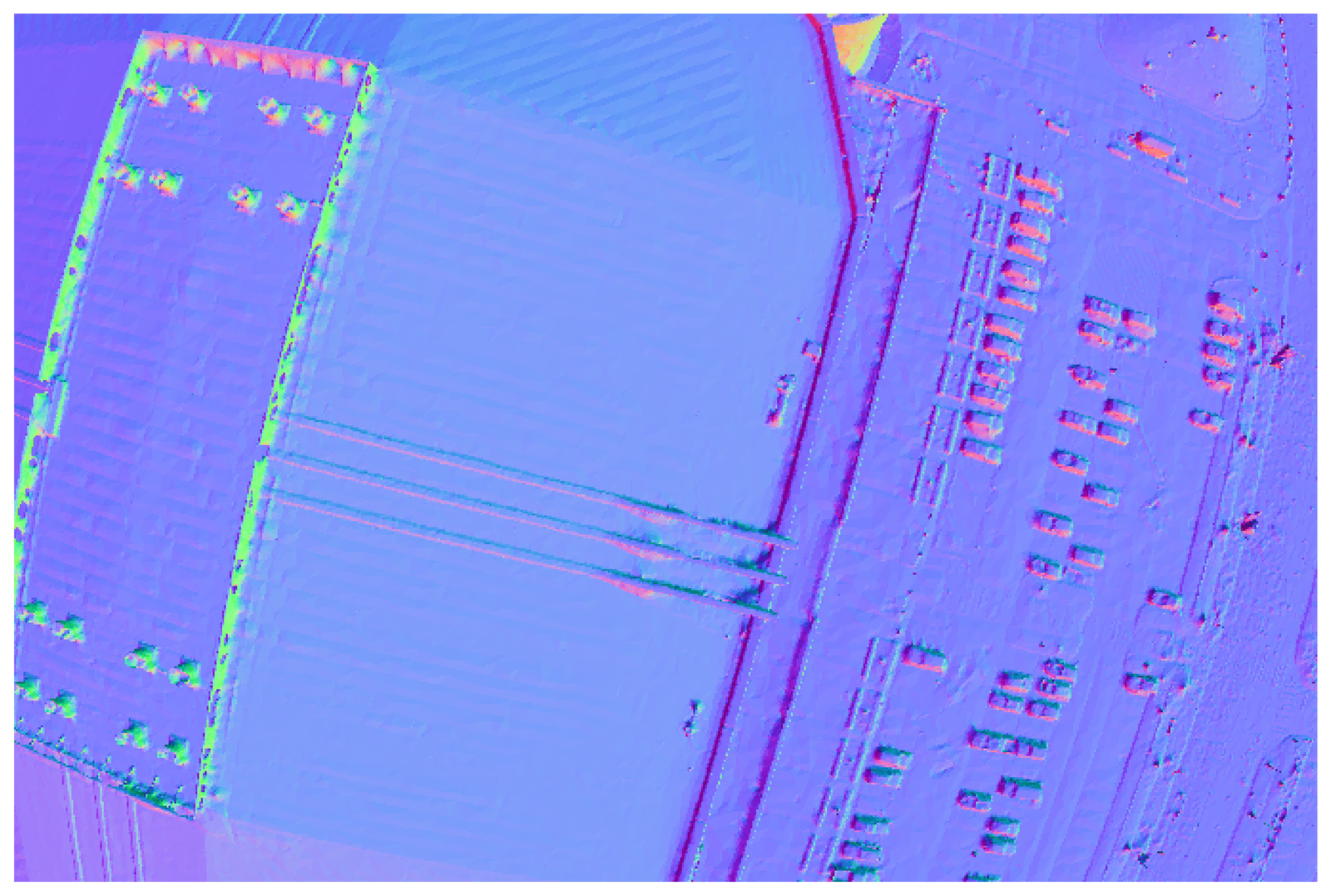} &
        \includegraphics[width=0.164\linewidth]{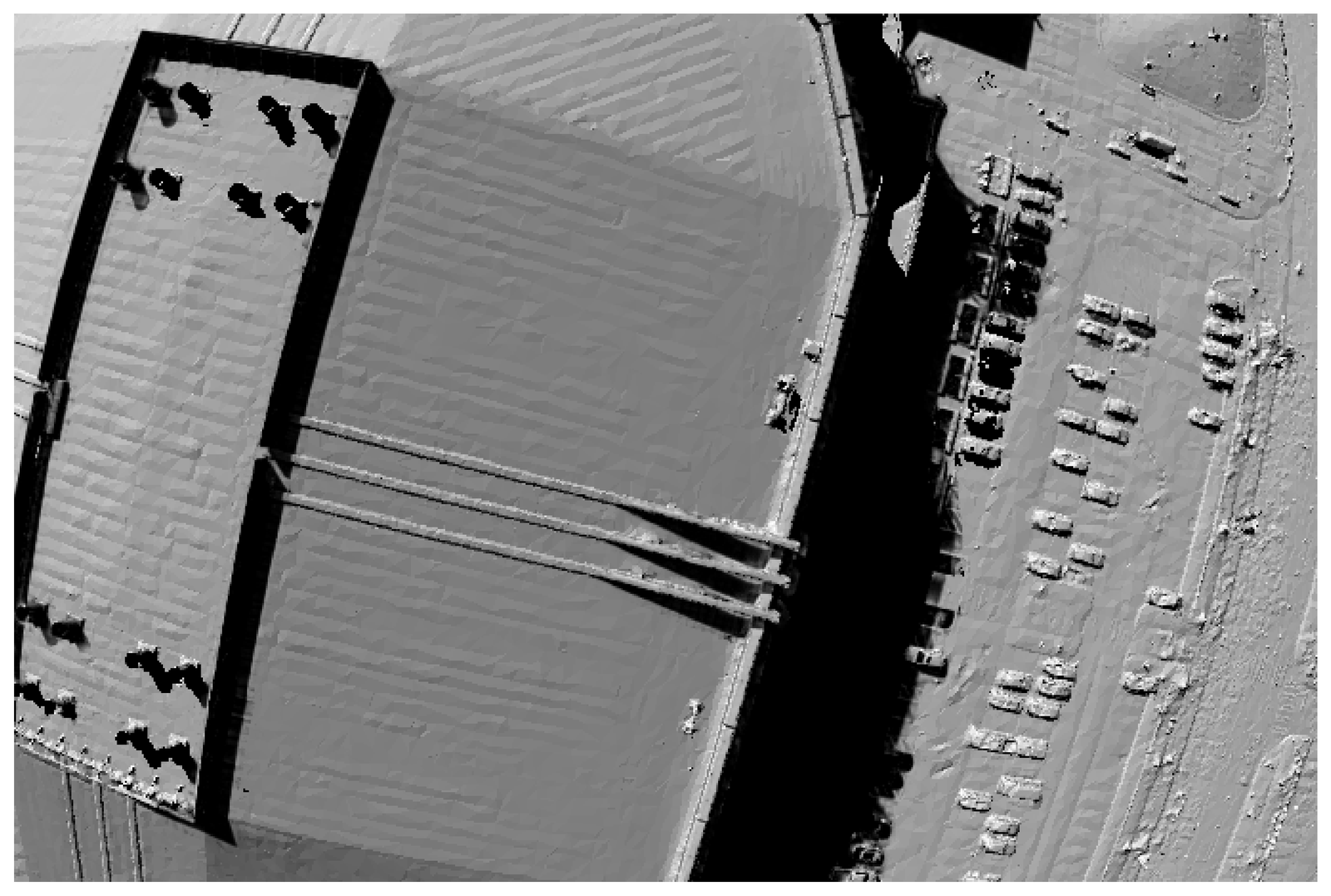} &
        \includegraphics[width=0.164\linewidth]{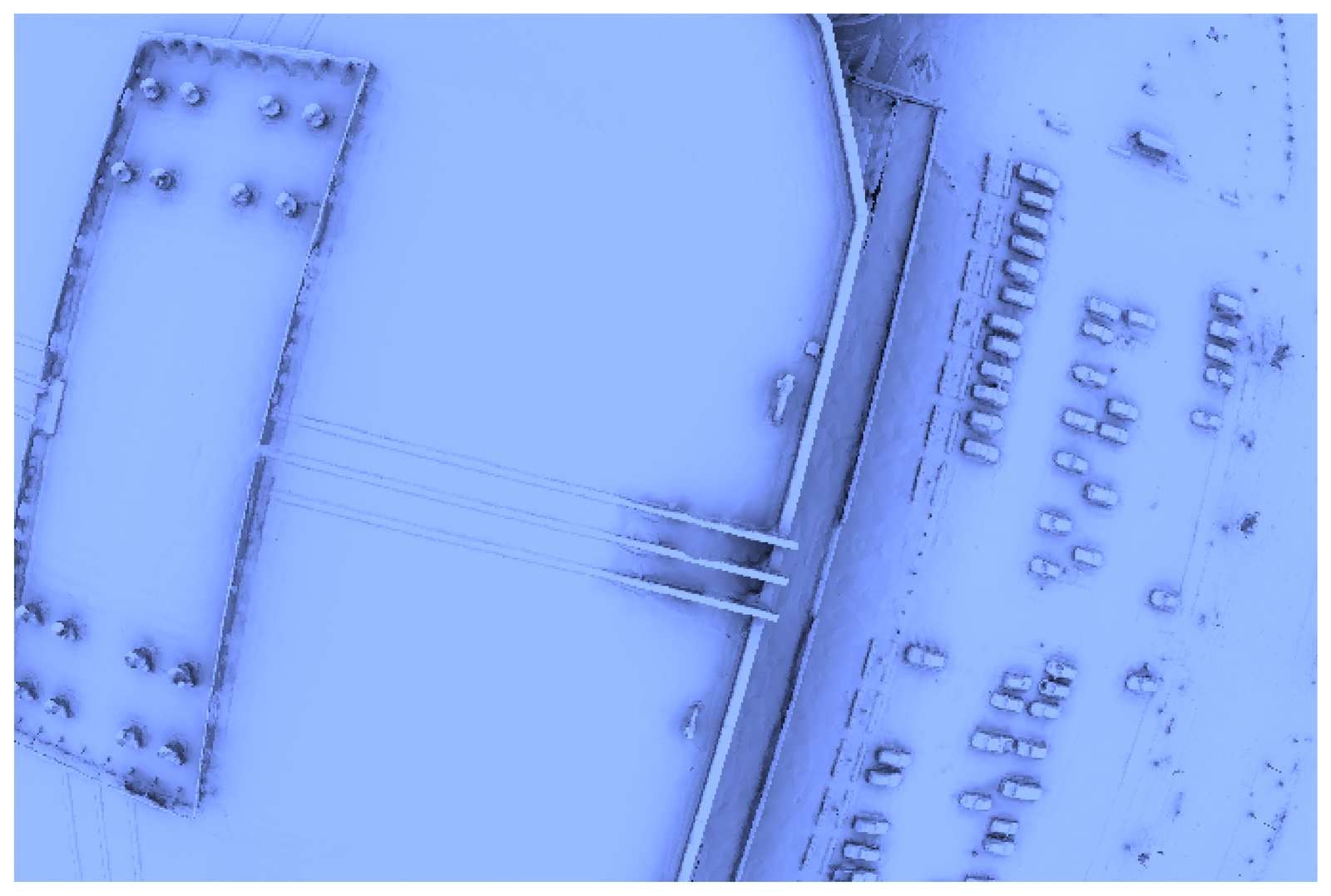} 
        \\
        \includegraphics[width=0.164\linewidth]{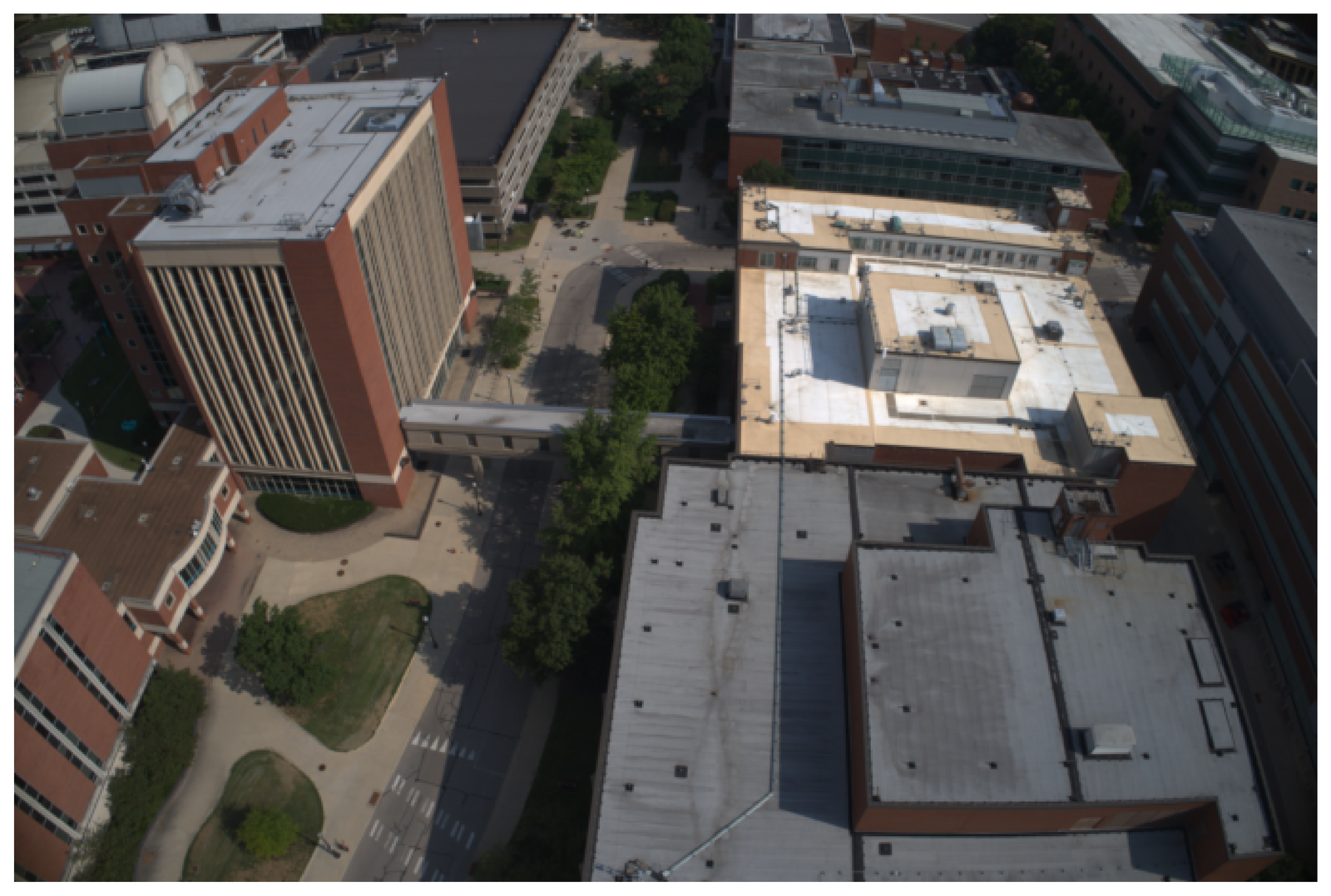} &
        \includegraphics[width=0.164\linewidth]{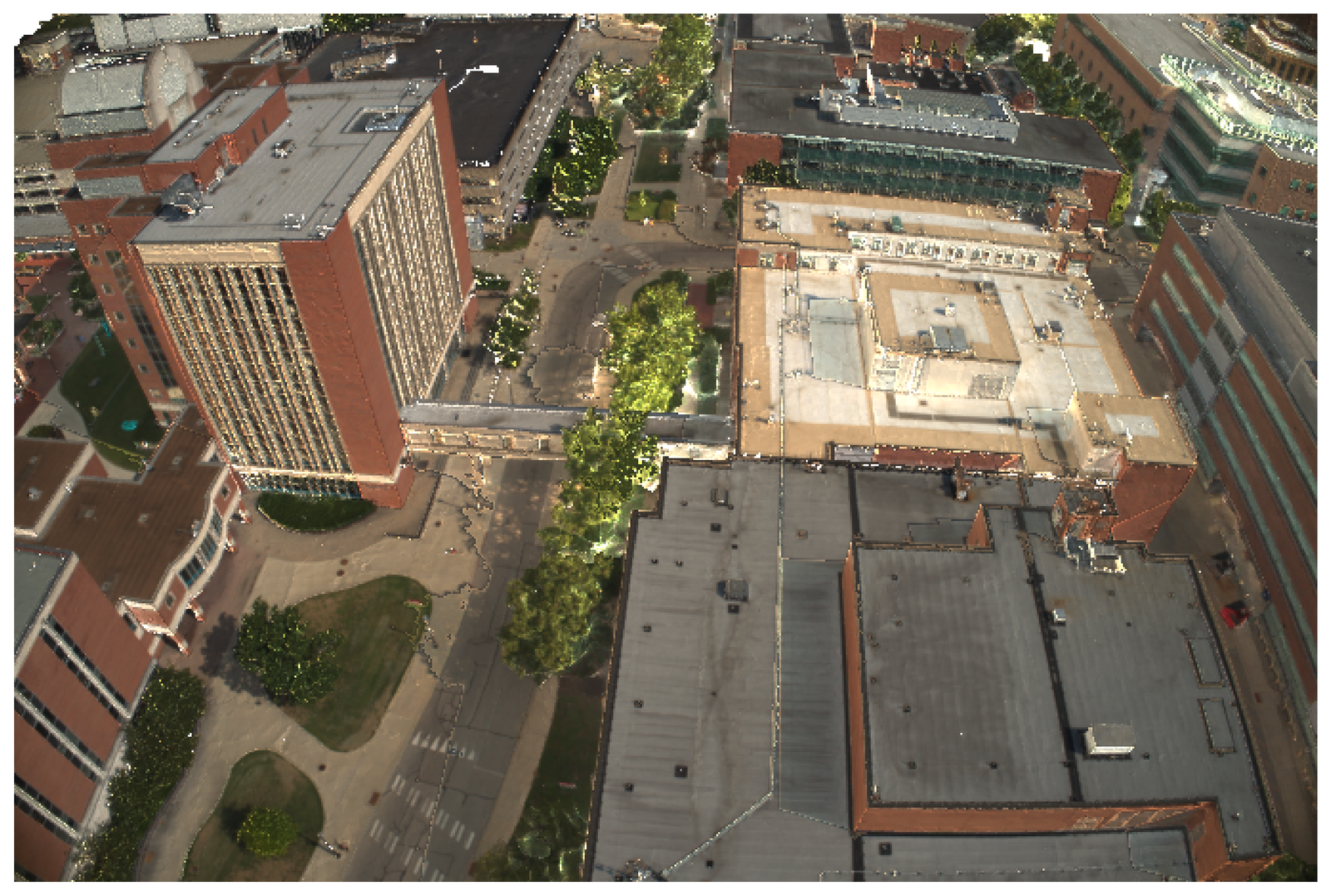} &
        \includegraphics[width=0.164\linewidth]{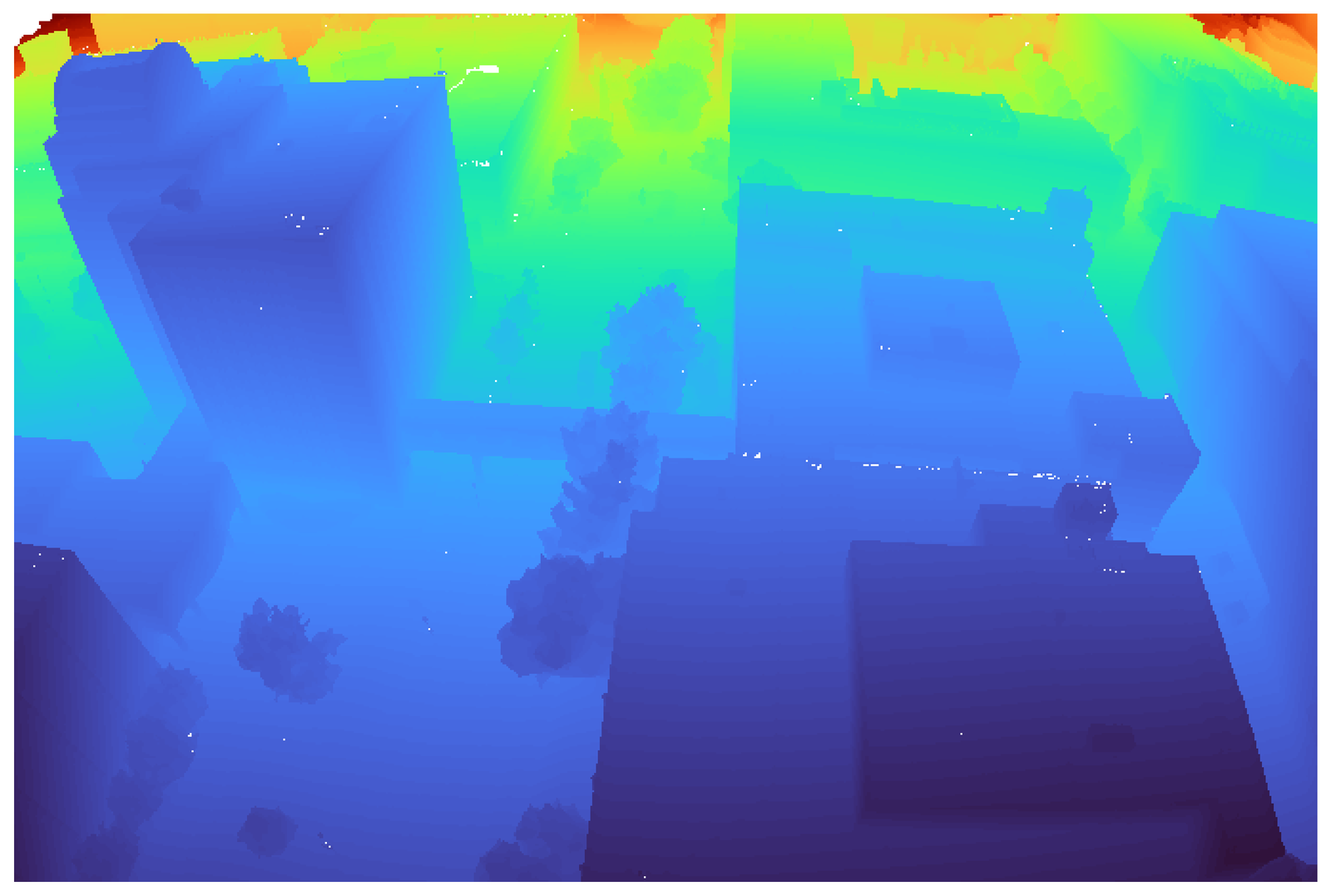} &
        \includegraphics[width=0.164\linewidth]{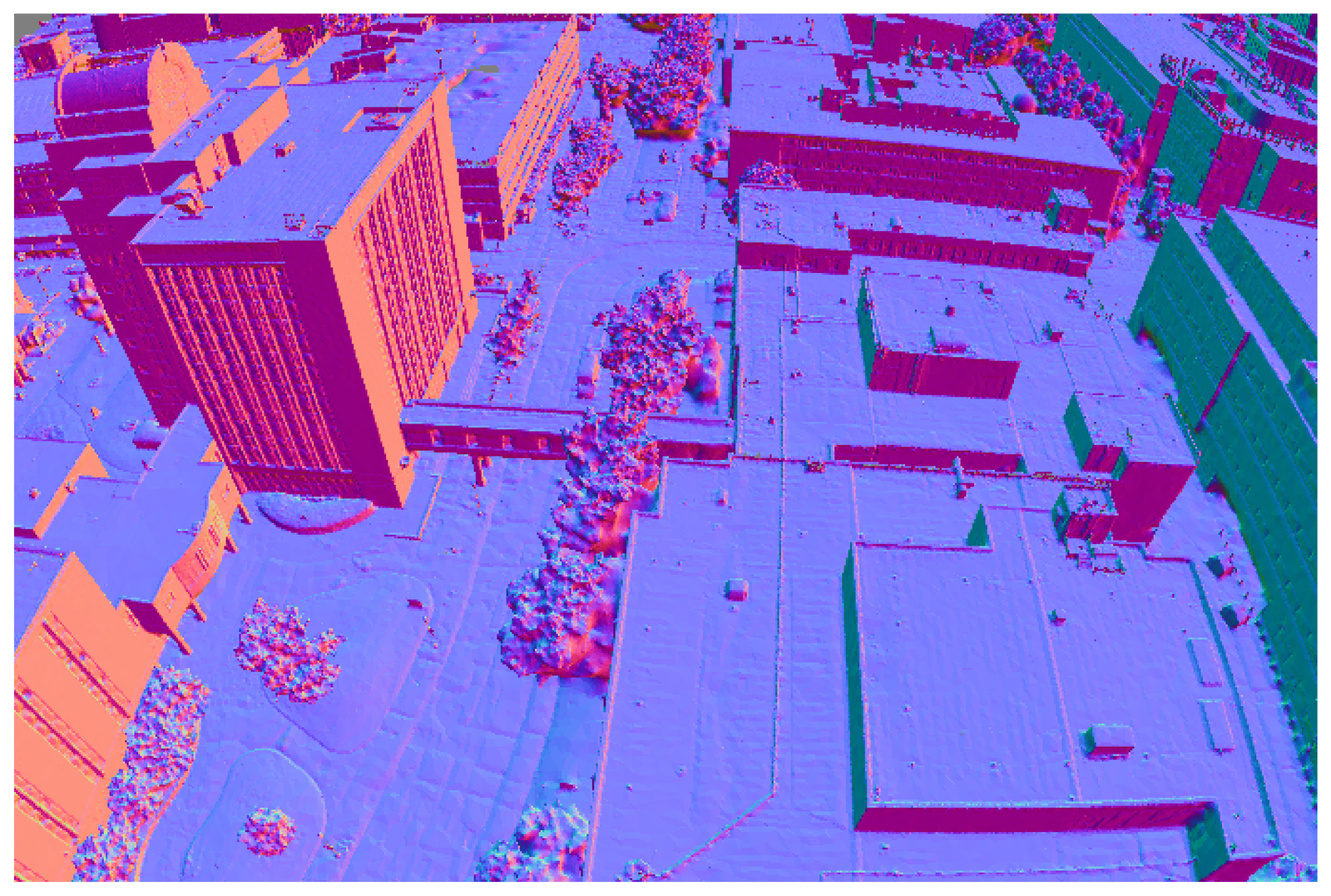} &
        \includegraphics[width=0.164\linewidth]{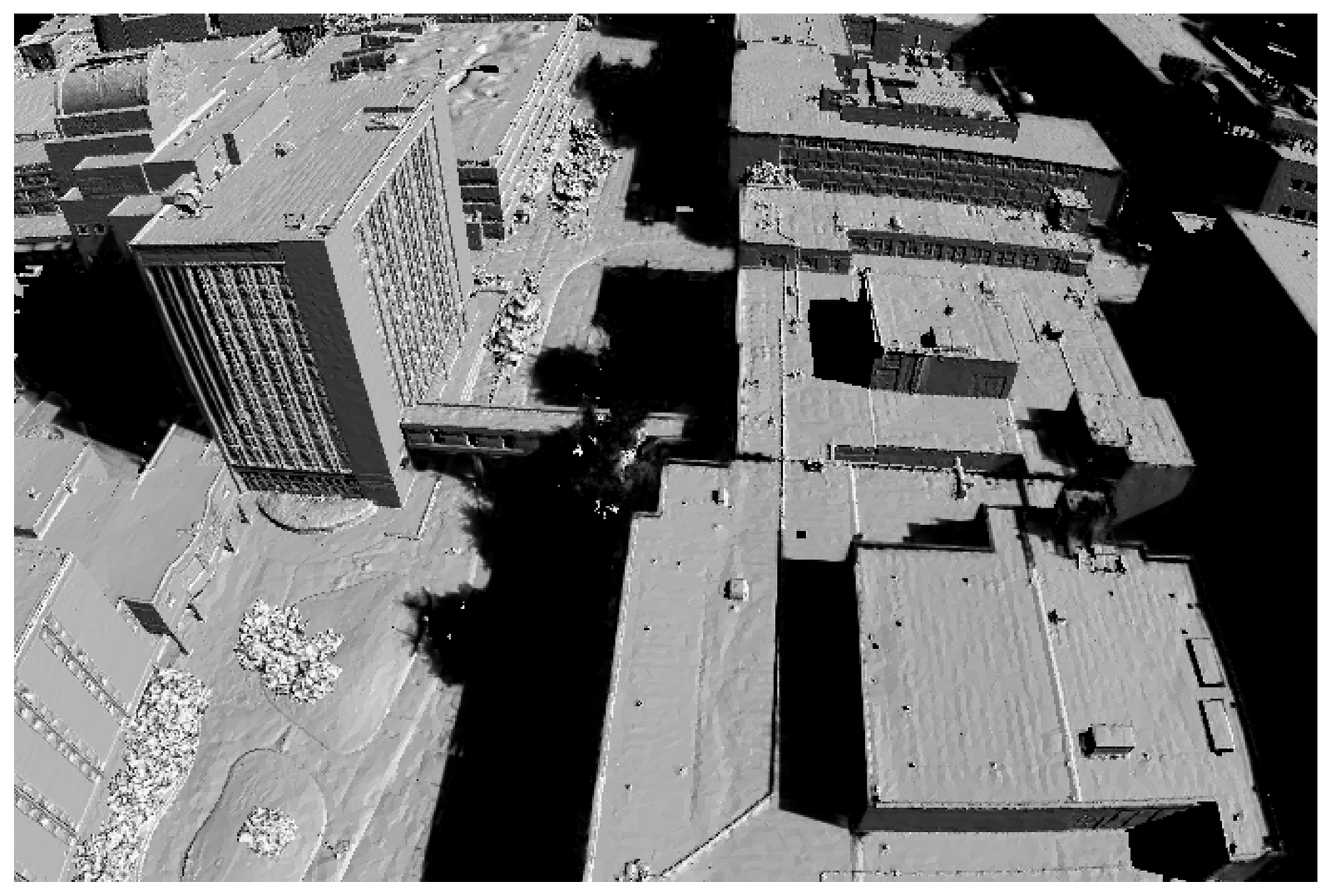} &
        \includegraphics[width=0.164\linewidth]{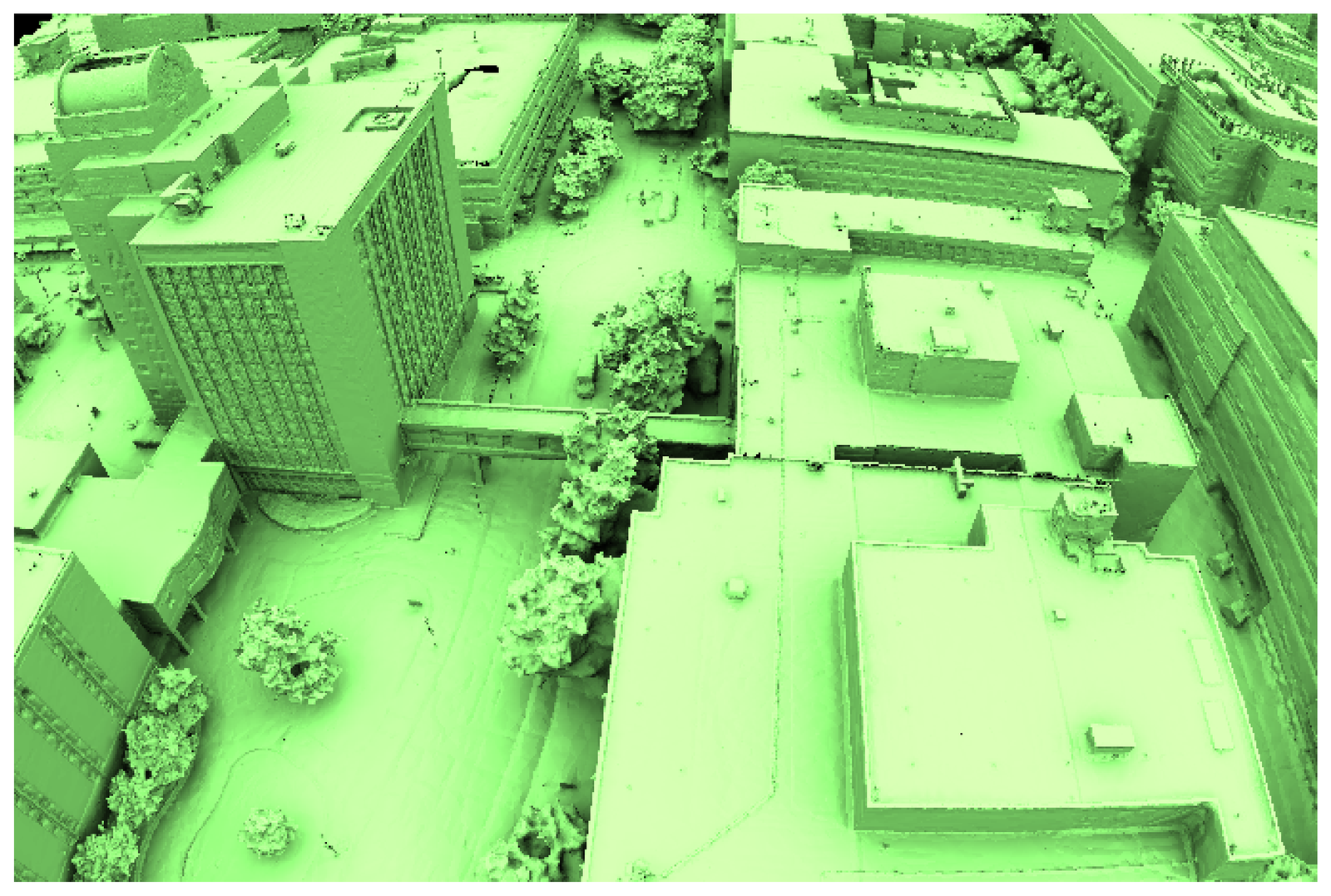} 
        \\
        \includegraphics[width=0.164\linewidth]{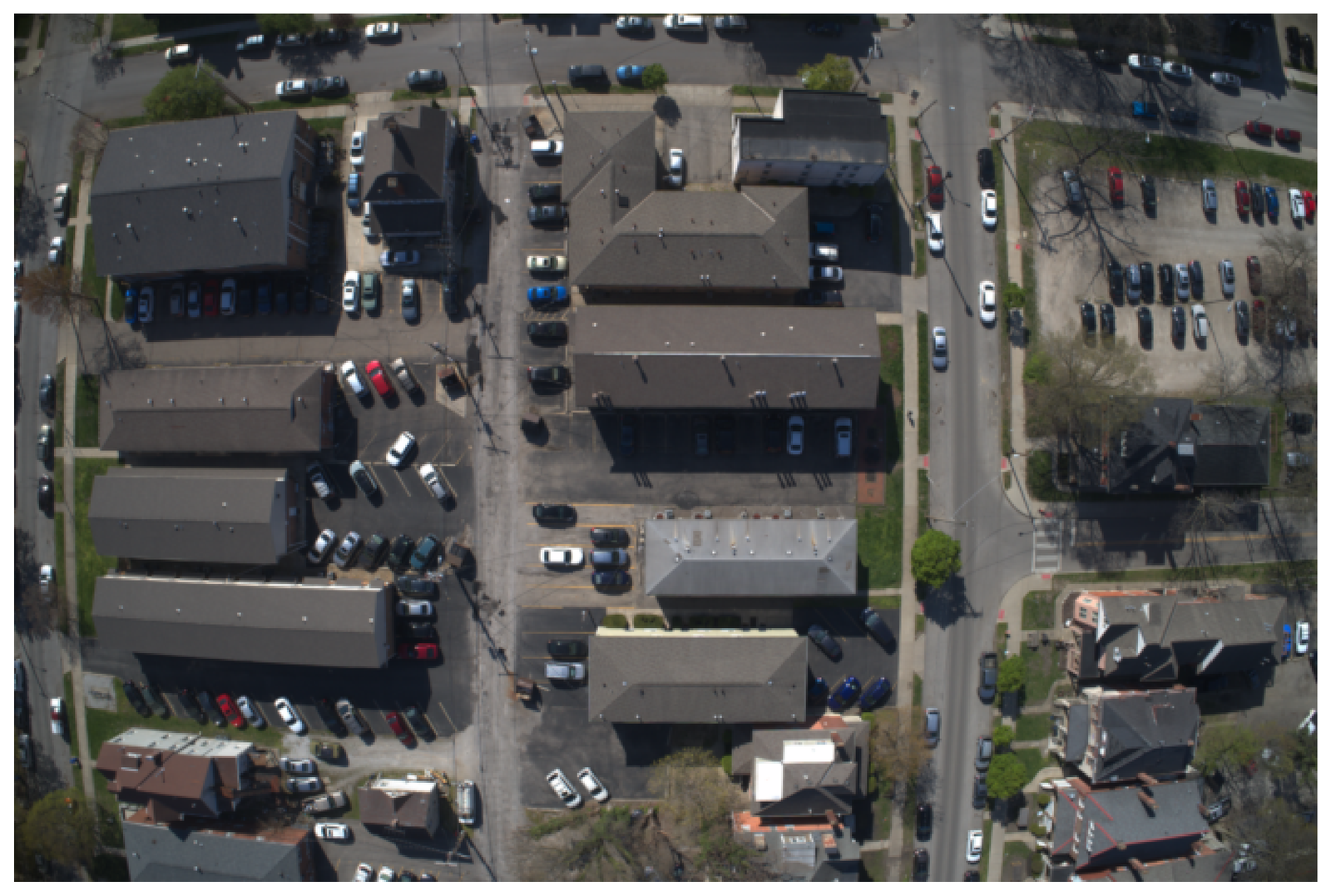} &
        \includegraphics[width=0.164\linewidth]{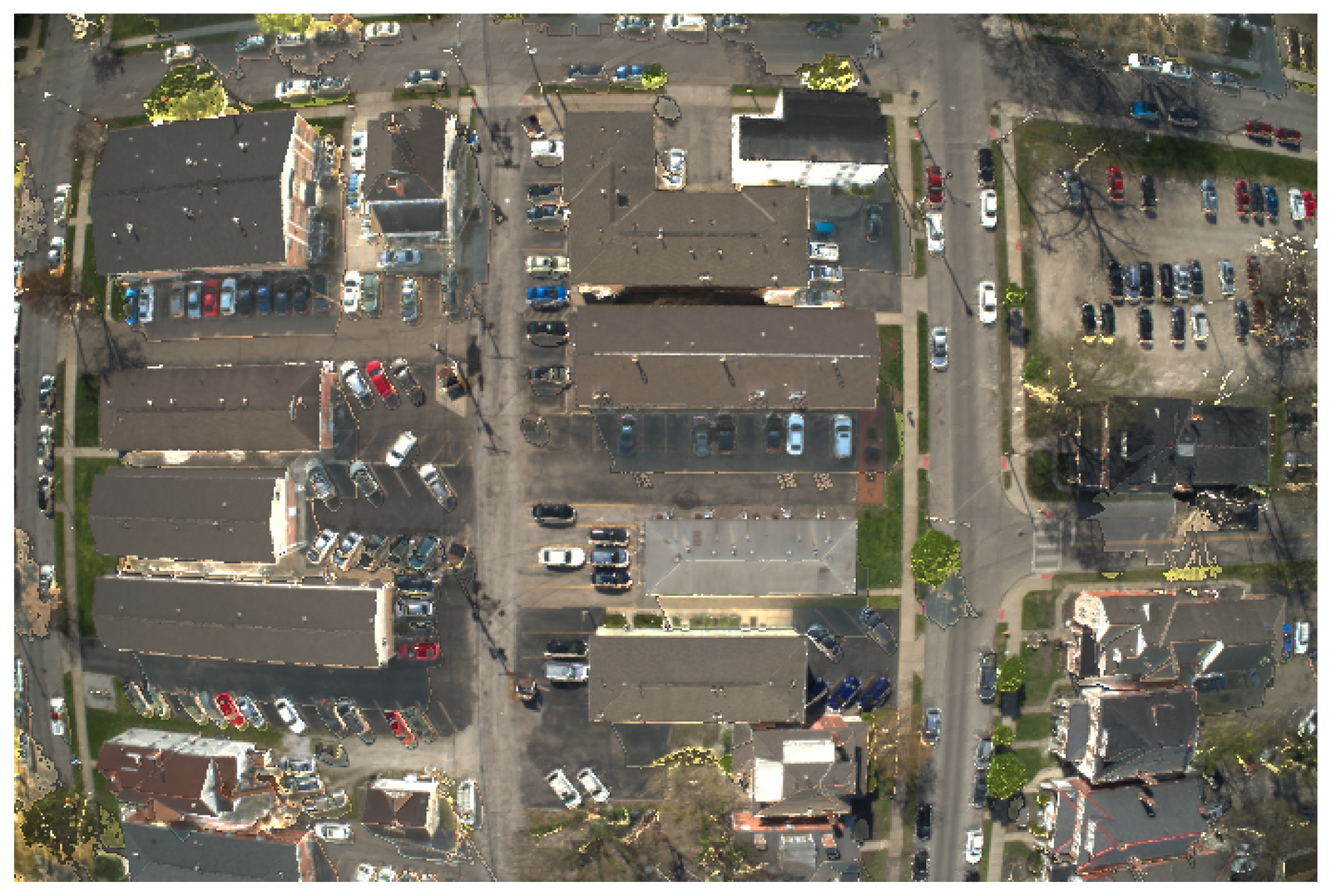} &
        \includegraphics[width=0.164\linewidth]{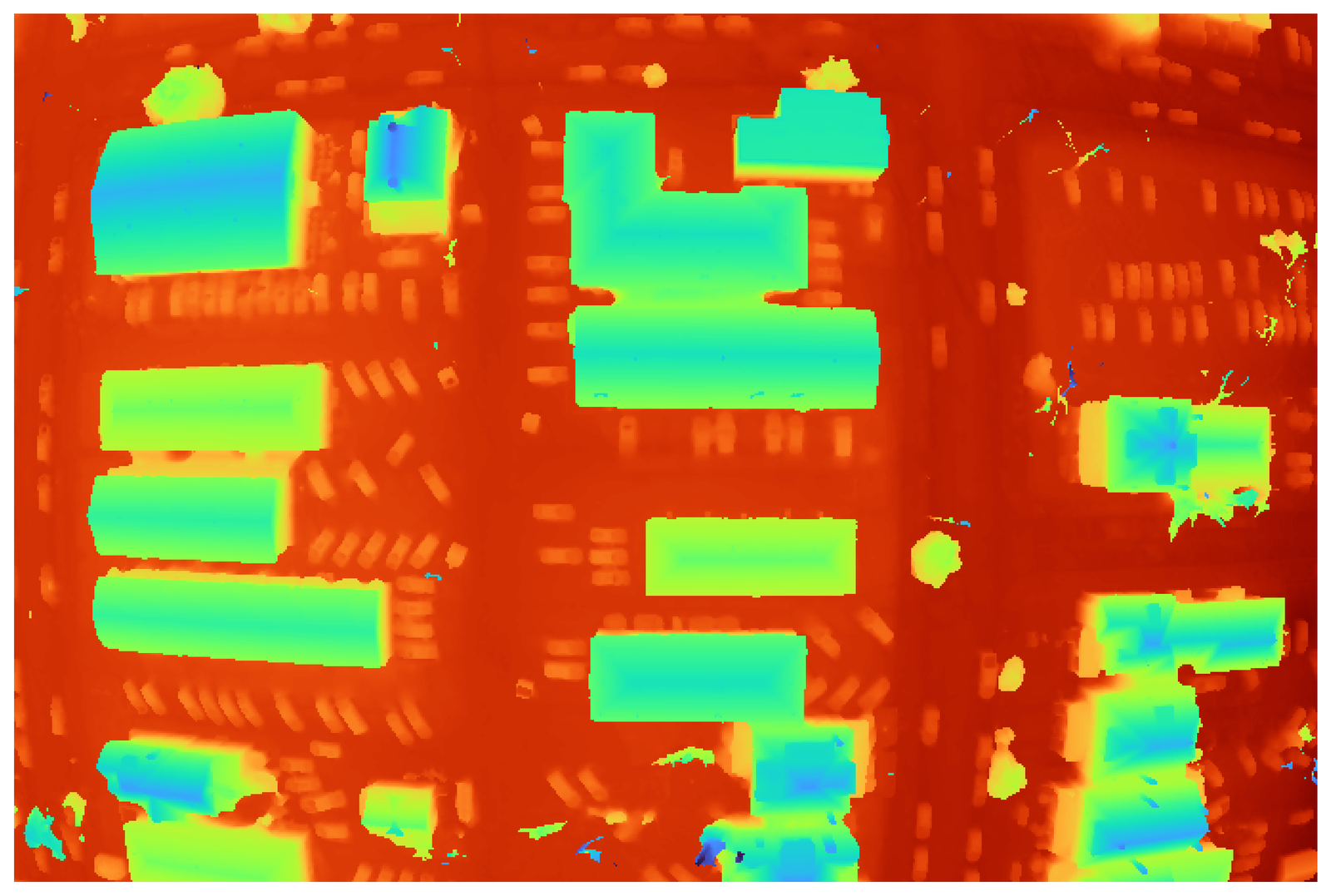} &
        \includegraphics[width=0.164\linewidth]{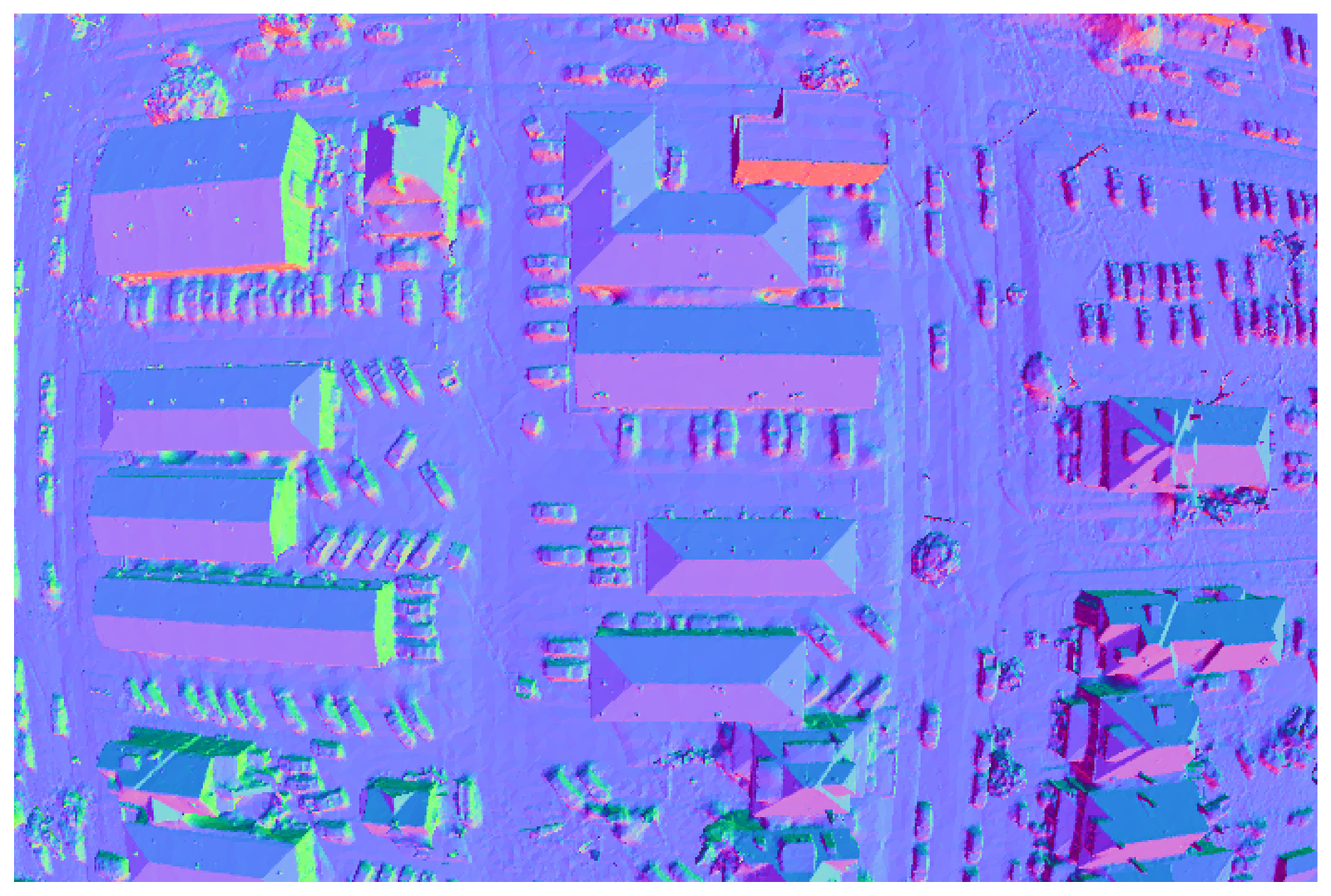} &
        \includegraphics[width=0.164\linewidth]{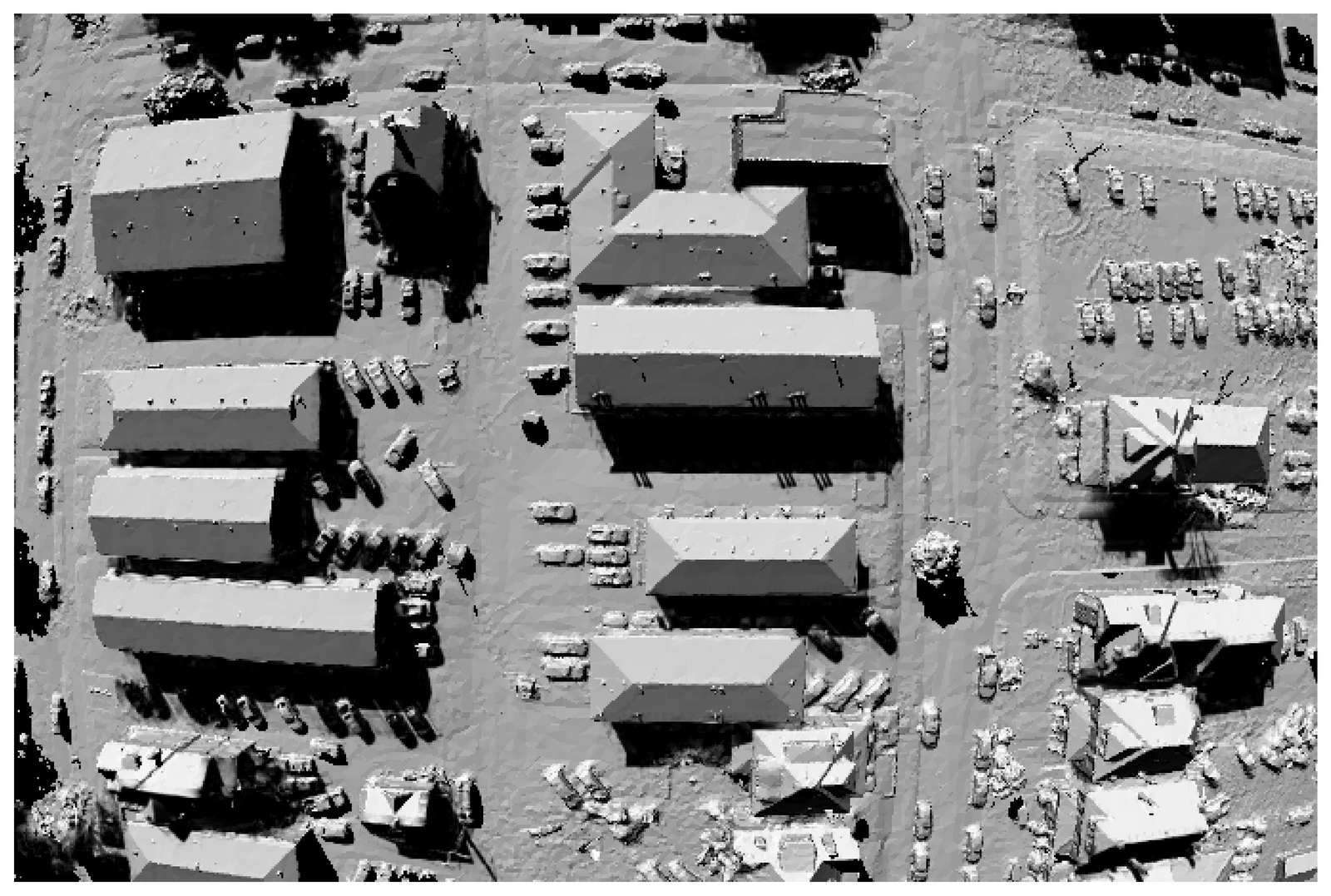} &
        \includegraphics[width=0.164\linewidth]{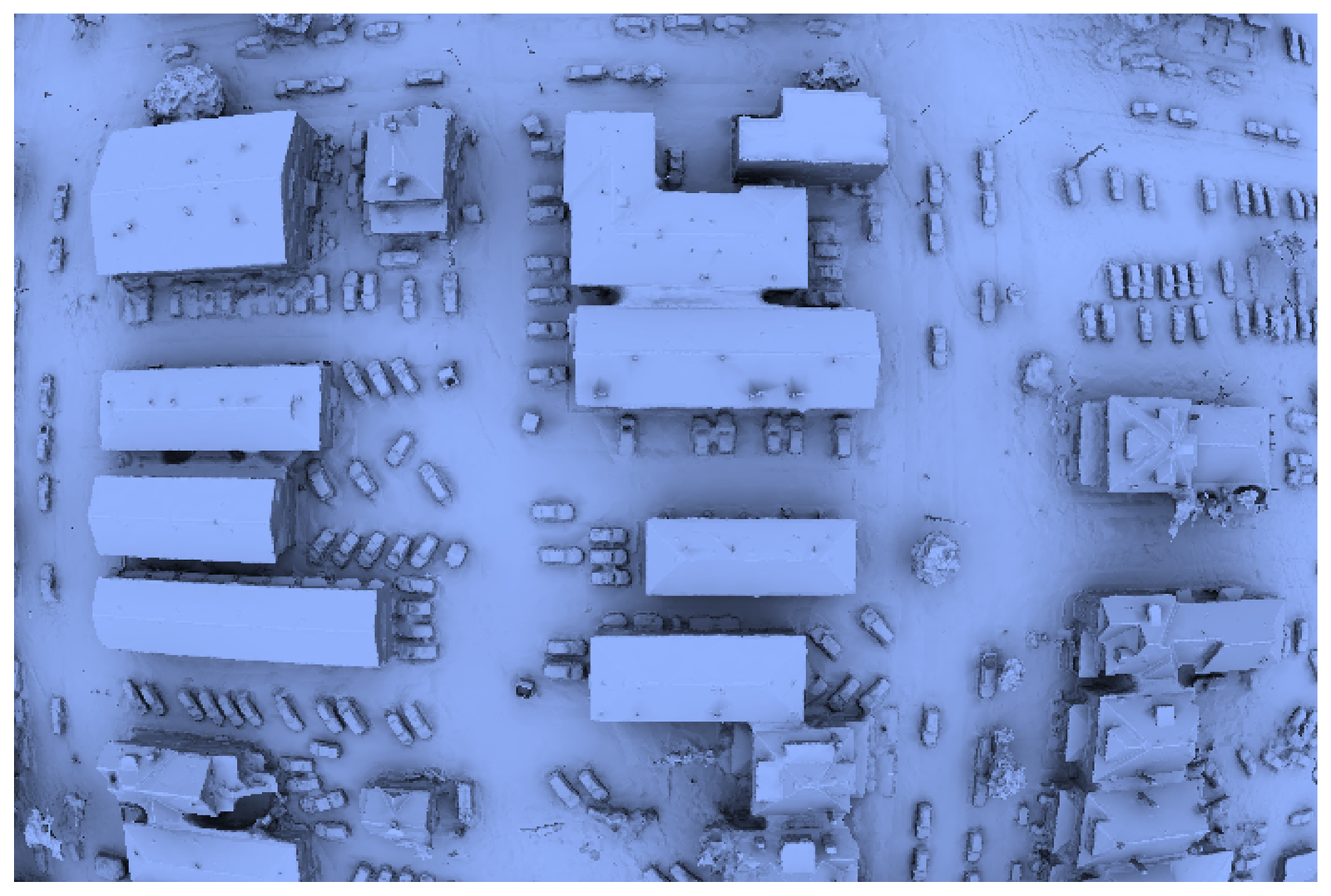} 
        \\
        \includegraphics[width=0.164\linewidth]{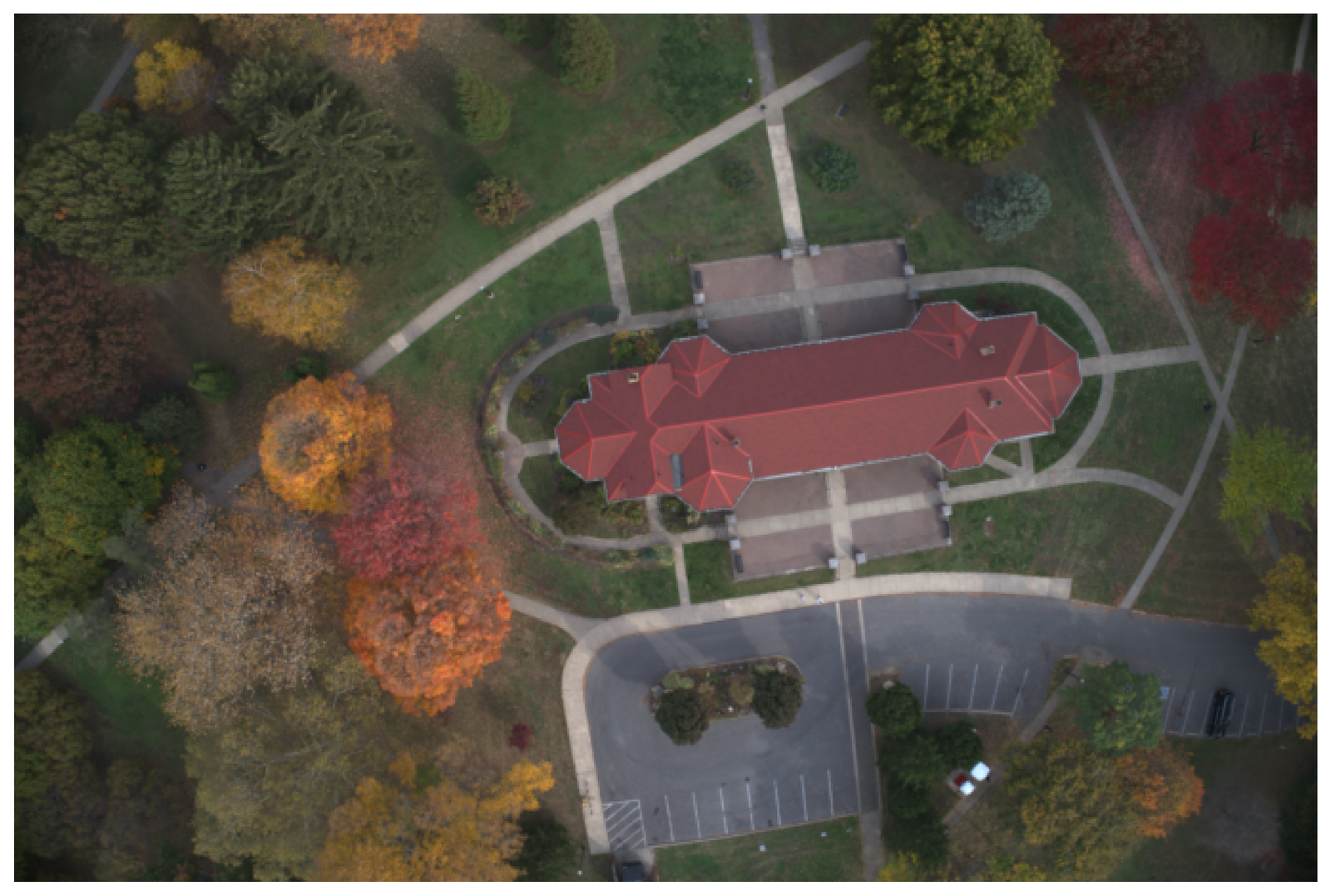} &
        \includegraphics[width=0.164\linewidth]{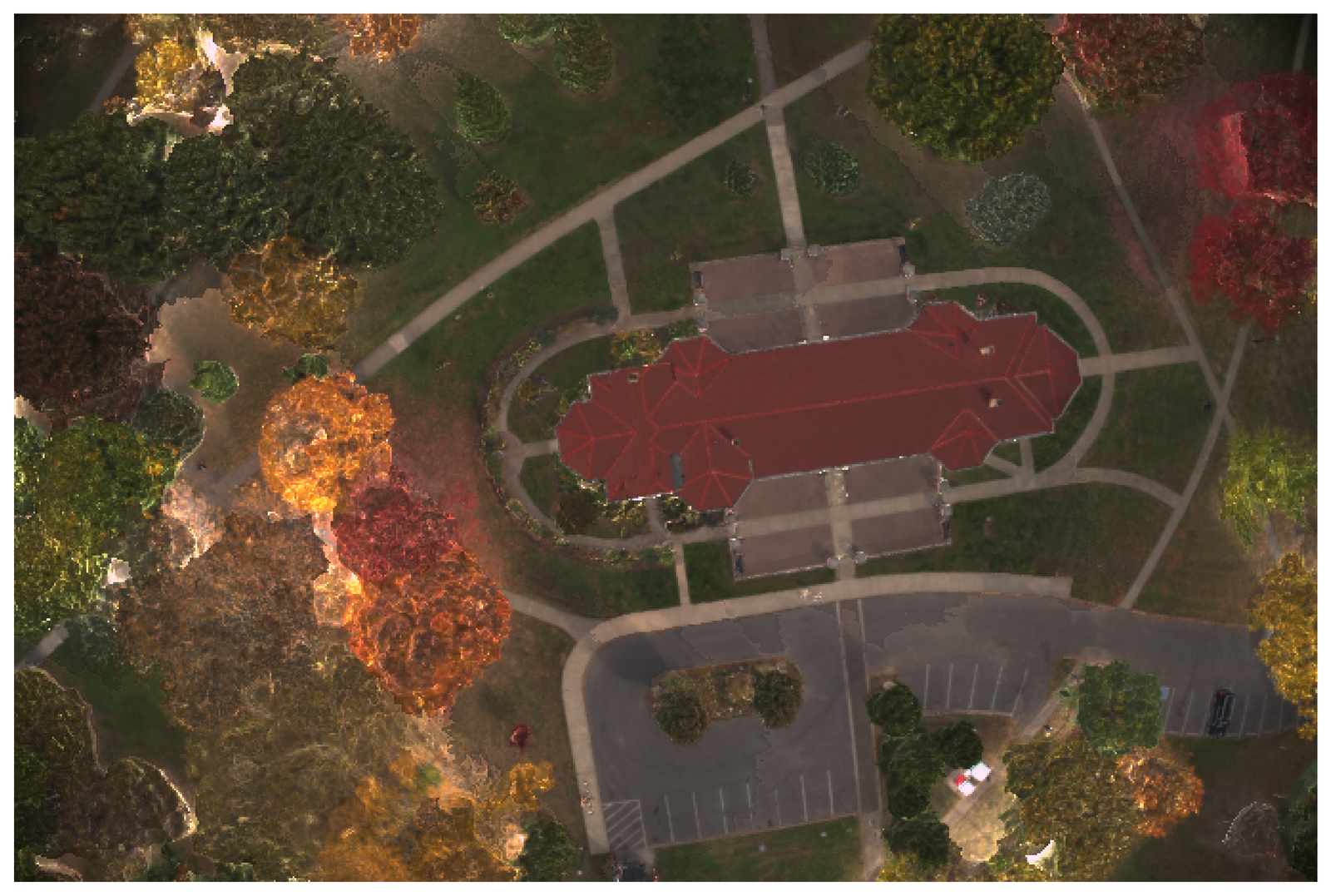} &
        \includegraphics[width=0.164\linewidth]{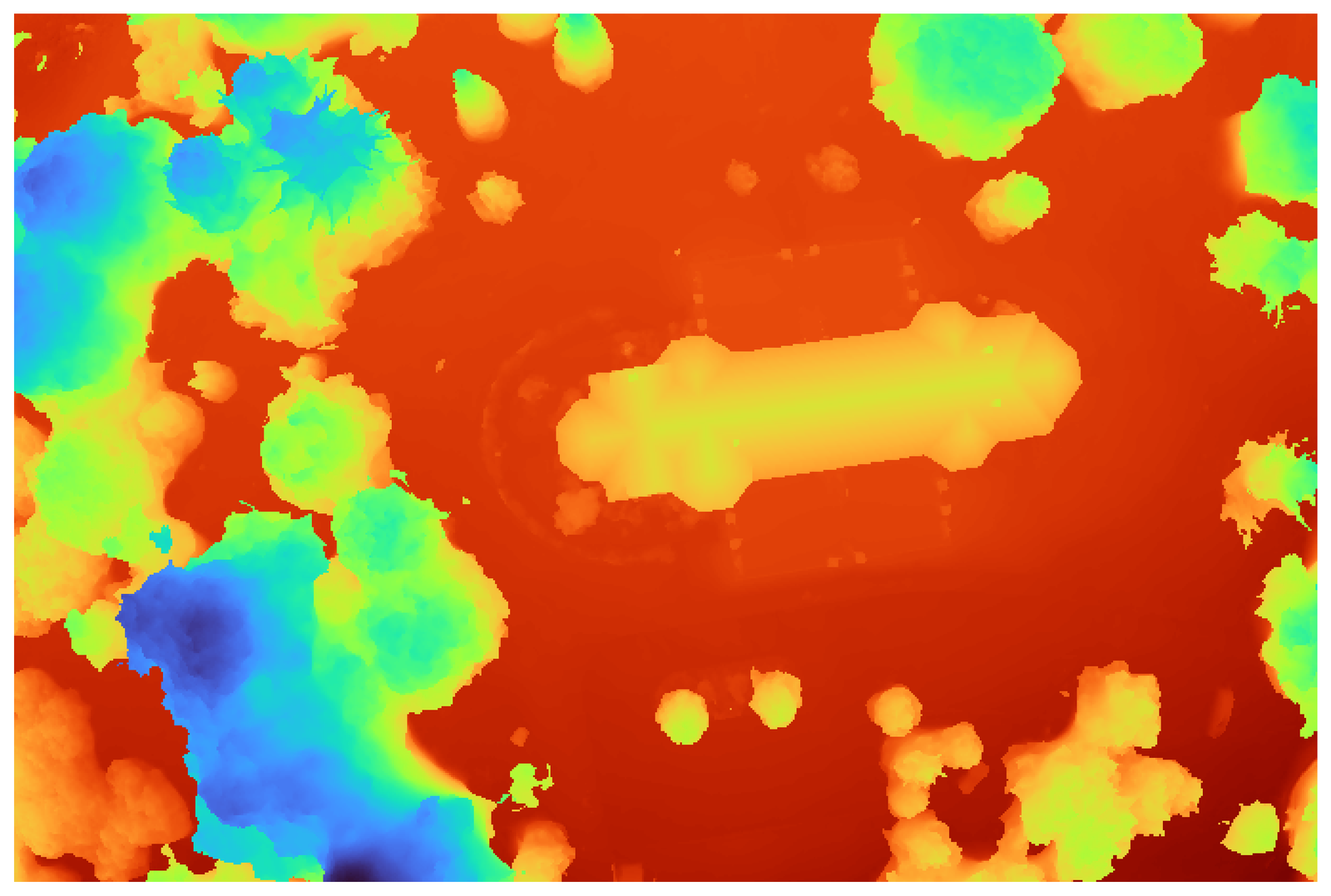} &
        \includegraphics[width=0.164\linewidth]{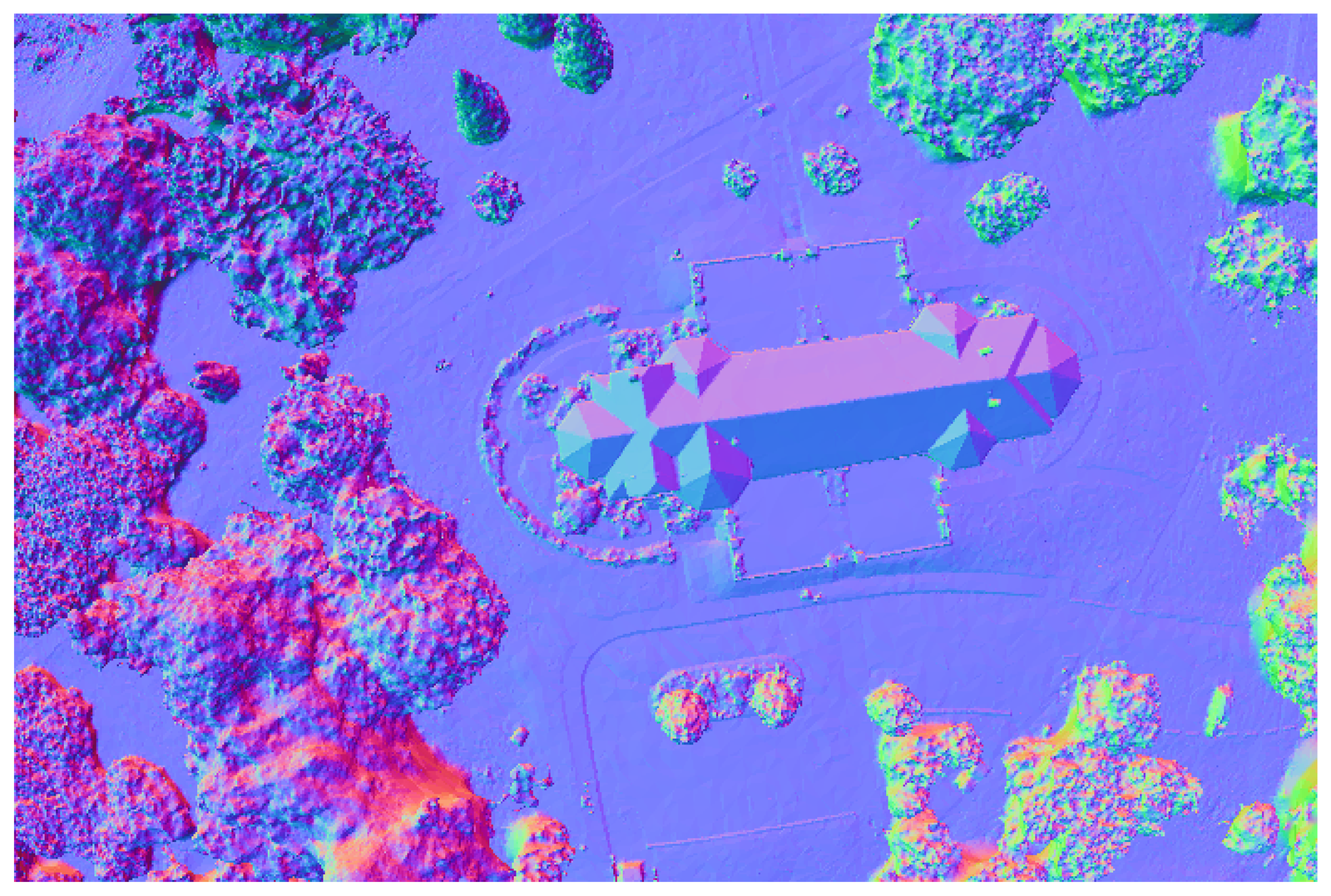} &
        \includegraphics[width=0.164\linewidth]{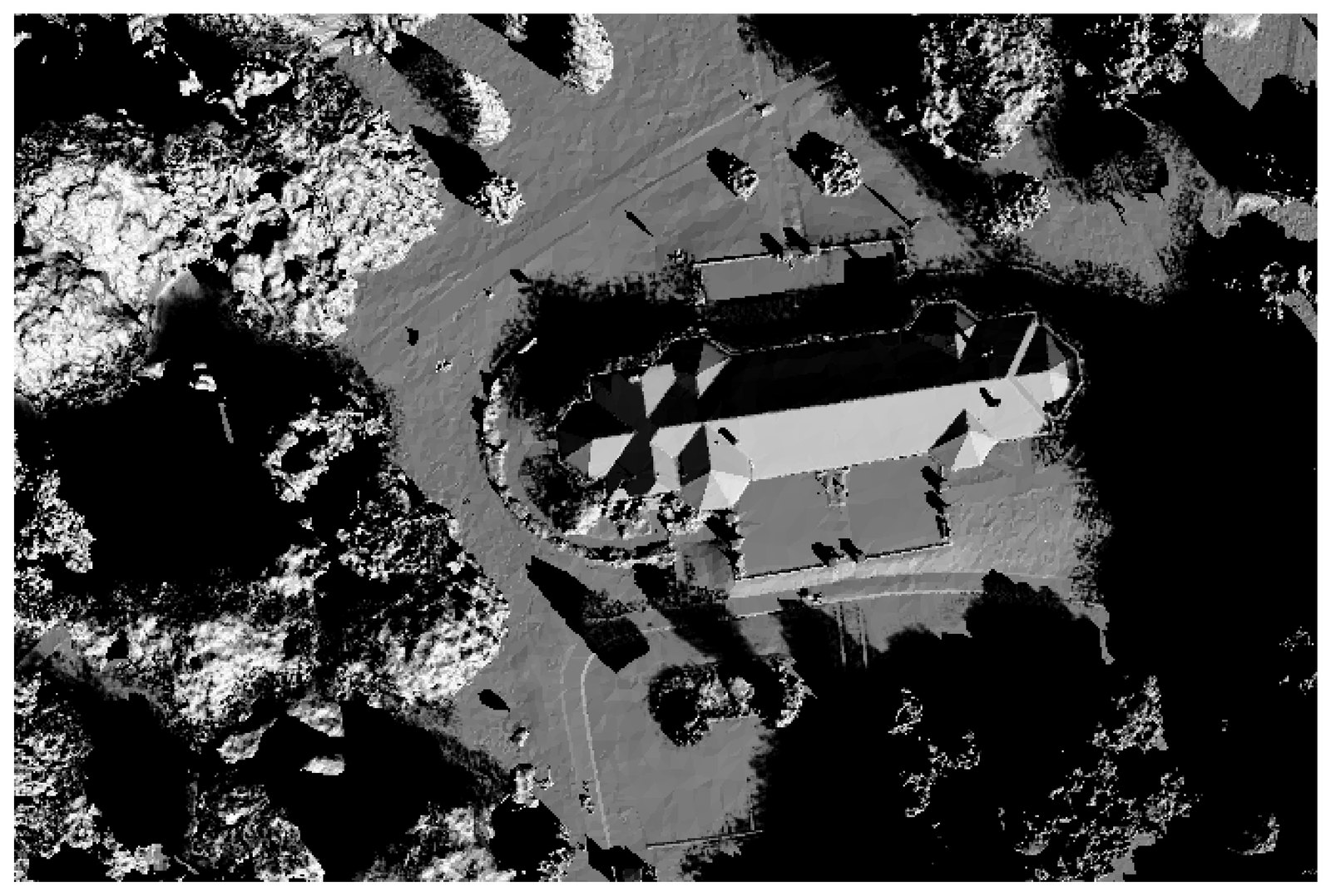} &
        \includegraphics[width=0.164\linewidth]{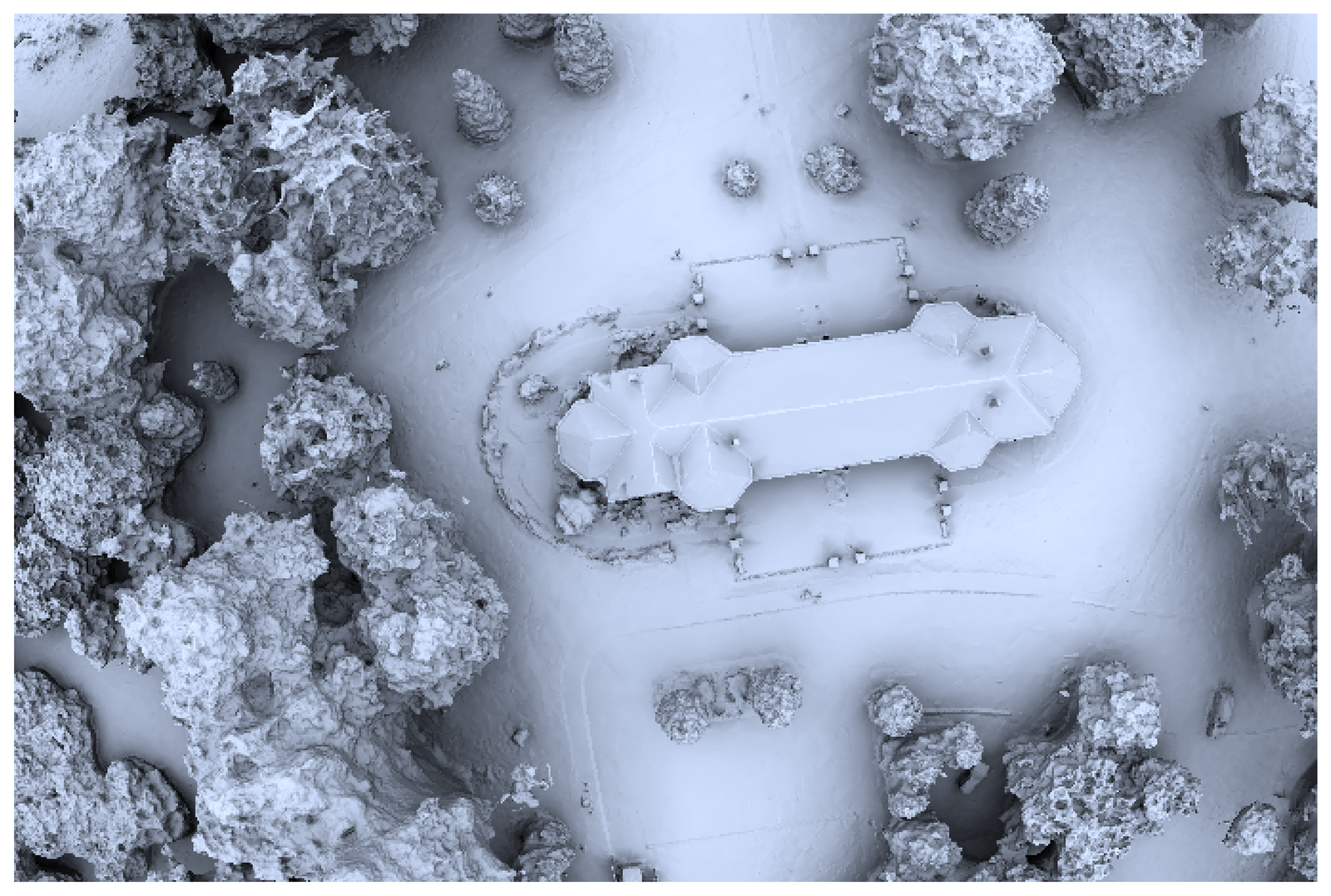} 
        \\
        {RGB} & {Albedo} & {Metric Depth} & {Normal} & {Sun Shading} & {Sky Shading}
    \end{tabular}
    \vspace{-10px}
    \caption{Additional examples from the Olbedo dataset. Rows from top to bottom: arena, office, residential, and park scenes.}
    \label{fig:data_more_example}
\end{figure*}

\section{Image Decomposition Method}
\label{supsec:method}

We adopt an outdoor illumination model proposed by Song et al.~\cite{songGeneralAlbedoRecovery2024} to describe the shading conditions in outdoor images. We start from the rendering equation below. The illumination consists of a directional component from solar radiation and a spherical component from the sky.

\begin{equation}
L_o(\mathbf{\omega}_o) = L_e(\mathbf{\omega}_o) + \int_{\Omega} L_i(\mathbf{\omega}_i) f_r(\mathbf{\omega}_i, \mathbf{\omega}_o) \langle \mathbf{\omega}_i \cdot \mathbf{n} \rangle^+ d \mathbf{\omega}_i,
\label{eq:rendering_equation}
\end{equation}
where $L_o(\mathbf{\omega}_o)$ is the outgoing radiance in direction $\mathbf{\omega}_o$, $L_e(\mathbf{\omega}_o)$ is the emitted radiance, $L_i(\mathbf{\omega}_i)$ is the incident radiance from direction $\mathbf{\omega}_i$, $f_r(\mathbf{\omega}_i, \mathbf{\omega}_o)$ is the bidirectional reflectance distribution function (BRDF), $\mathbf{n}$ is the surface normal, and $(\cdot)^+$ denotes the positive-part function, i.e., $\max(0, \cdot)$.

For diffuse, non-emissive surfaces, the rendering equation simplifies to
\begin{equation}
L_o(\mathbf{\omega}_o) = \int_{\Omega} L_i(\mathbf{\omega}_i) \langle \mathbf{\omega}_i \cdot \mathbf{n} \rangle^+ d \mathbf{\omega}_i,
\end{equation}

\begin{equation}
    \mathrm{where} \quad L_i(\mathbf{\omega}_i) = L_{sun}(\mathbf{\omega}_i) + L_{sky}(\mathbf{\omega}_i).
\end{equation}

The incident radiance $L_i(\mathbf{\omega}_i)$ can be further decomposed into sun and sky components as \cref{eq:sun_shading_def,eq:sky_shading_def}.

\begin{equation} \label{eq:sun_shading_def}
    L_{sun}(\mathbf{\omega}_i) = \psi_{sun} \cdot V_{sun} \ast \delta(\mathbf{\omega}_i - \mathbf{\theta}_{sun}),
\end{equation}
where $\psi_{sun}$ is the solar radiance constant, $V_{sun}$ is the visibility function for the sun (1 when the sun is visible, 0 otherwise), and $\delta$ is the Dirac delta function centered at the sun direction $\mathbf{\theta}_{sun}$.

\begin{equation} \label{eq:sky_shading_def}
    L_{sky}(\mathbf{\omega}_i) = \psi_{sky} \cdot \mathcal{G}(\mathbf{\omega}_i - \mathbf{\theta}_{sun}) \cdot V_{sky}(\mathbf{\omega}_i),
\end{equation}
where $\psi_{sky}$ is the sky radiance constant, and $\mathcal{G}$ is a spherical function modeling the sky light distribution, with its lobe centered at the sun direction $\mathbf{\theta}_{sun}$. To simplify the model in practice, we use a uniform sky distribution, i.e., $\mathcal{G}$ is constant over the visible sky. $V_{sky}$ is the sky visibility function; it is defined on the hemisphere above the horizontal plane, matching the support of the sky light distribution.

\begin{equation}
\phi = \frac{\psi_{sun}}{\psi_{sky}}.
\end{equation}

We can integrate the sun and sky components separately and rearrange them as normalized shading components corresponding to the image formation model described in the main paper:

\begin{equation}    
\begin{aligned}
    S_{sun} &= V_{sun} \cdot \langle \mathbf{n}, \mathbf{\theta}_{sun} \rangle^+, \\
    S_{sky} &= \int_{\Omega} V_{sky}(\mathbf{\omega}_i) \cdot \mathcal{G}(\mathbf{\omega}_i - \mathbf{\theta}_{sun}) \cdot \langle \mathbf{n}, \mathbf{\omega}_i \rangle^+ d \mathbf{\omega}_i .
\end{aligned}
\end{equation}

In \cref{fig:data_more_example}, we present additional data samples spanning multiple scenes and modalities from the \textit{Olbedo} dataset.  


\section{Failure Cases in Pseudo-Ground-Truth Generation} \label{supsec:failure_cases}
\begin{figure*}[!htpb]
    \centering
    \setlength{\tabcolsep}{2pt}
    \renewcommand{\arraystretch}{1.0}
    \begin{tabular}{ccccc}
        \includegraphics[height=0.12\textheight]{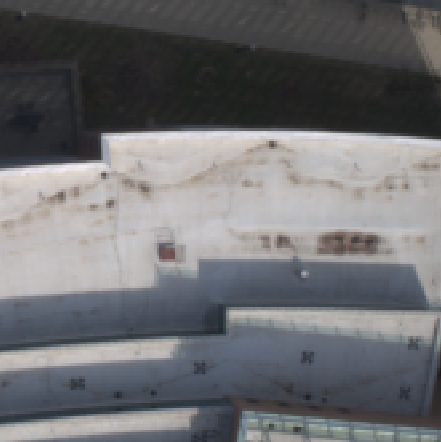} &
        \includegraphics[height=0.12\textheight]{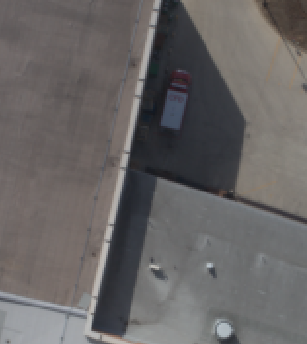} &
        \includegraphics[height=0.12\textheight]{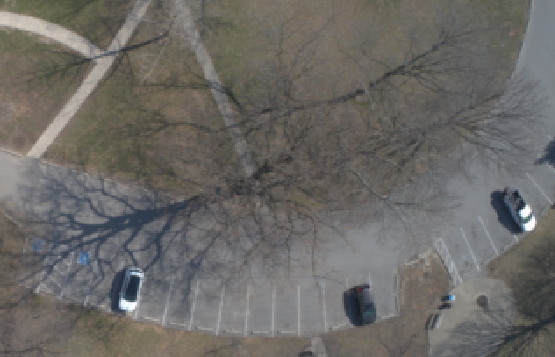} &
        \includegraphics[height=0.12\textheight]{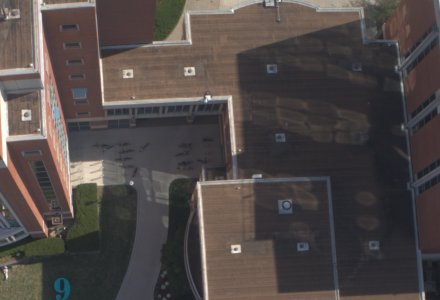} &
        \includegraphics[height=0.12\textheight,trim=0 80 70 0,clip]{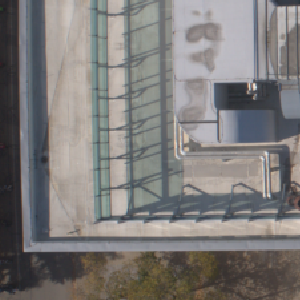} \\
        \includegraphics[height=0.12\textheight]{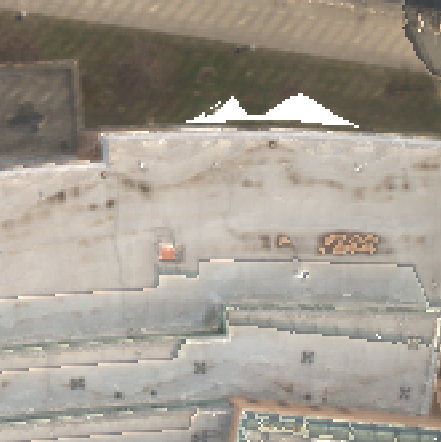} &
        \includegraphics[height=0.12\textheight]{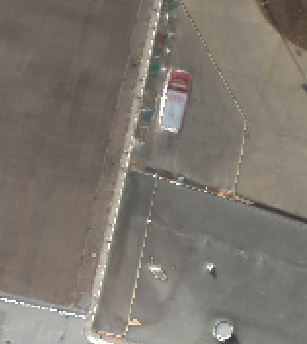} &
        \includegraphics[height=0.12\textheight]{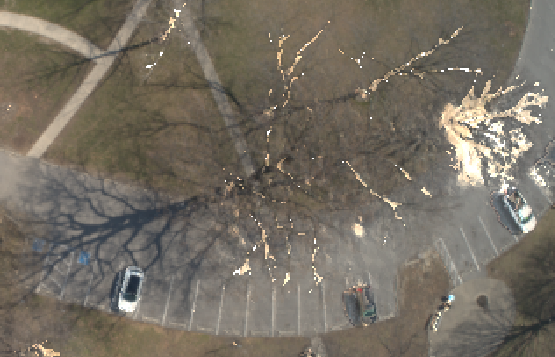} &
        \includegraphics[height=0.12\textheight]{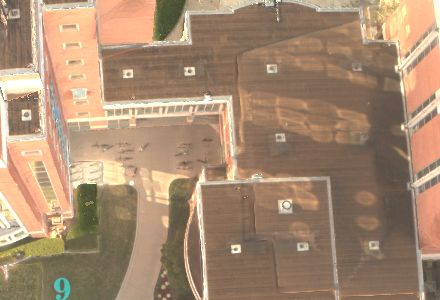} &
        \includegraphics[height=0.12\textheight,trim=0 80 70 0,clip]{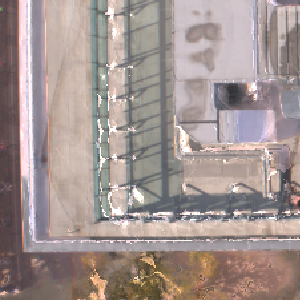} \\
        {\footnotesize (a)} & {\footnotesize (b)} & {\footnotesize (c)} & {\footnotesize (d)} & {\footnotesize (e)}
    \end{tabular}
    \caption{Representative failure cases in pseudo-ground-truth generation. The top row shows the raw inputs and the bottom row shows the corresponding recovered albedo. Columns (a)--(e) illustrate geometry holes, sharp shadow-boundary noise, imperfect tree geometry, glass reflection, and glass transmission, respectively.}
    \label{fig:failure_cases}
\end{figure*}

During data production, we observe several recurring failure modes of the pseudo-ground-truth generation process, shown in \cref{fig:failure_cases}. These include geometry holes, sharp shadow-boundary noise, imperfect tree geometry, glass reflection, and glass transmission. To improve the training data quality, we manually screen the generated results and identify a subset of 2.49k relatively clean images with fewer artifacts such as incorrect shadows, reflections, and transmissions.

\section{Confidence Masks} \label{supsec:confidence_mask}
\begin{figure*}[!htpb]
    \centering
    \includegraphics[width=0.75\linewidth]{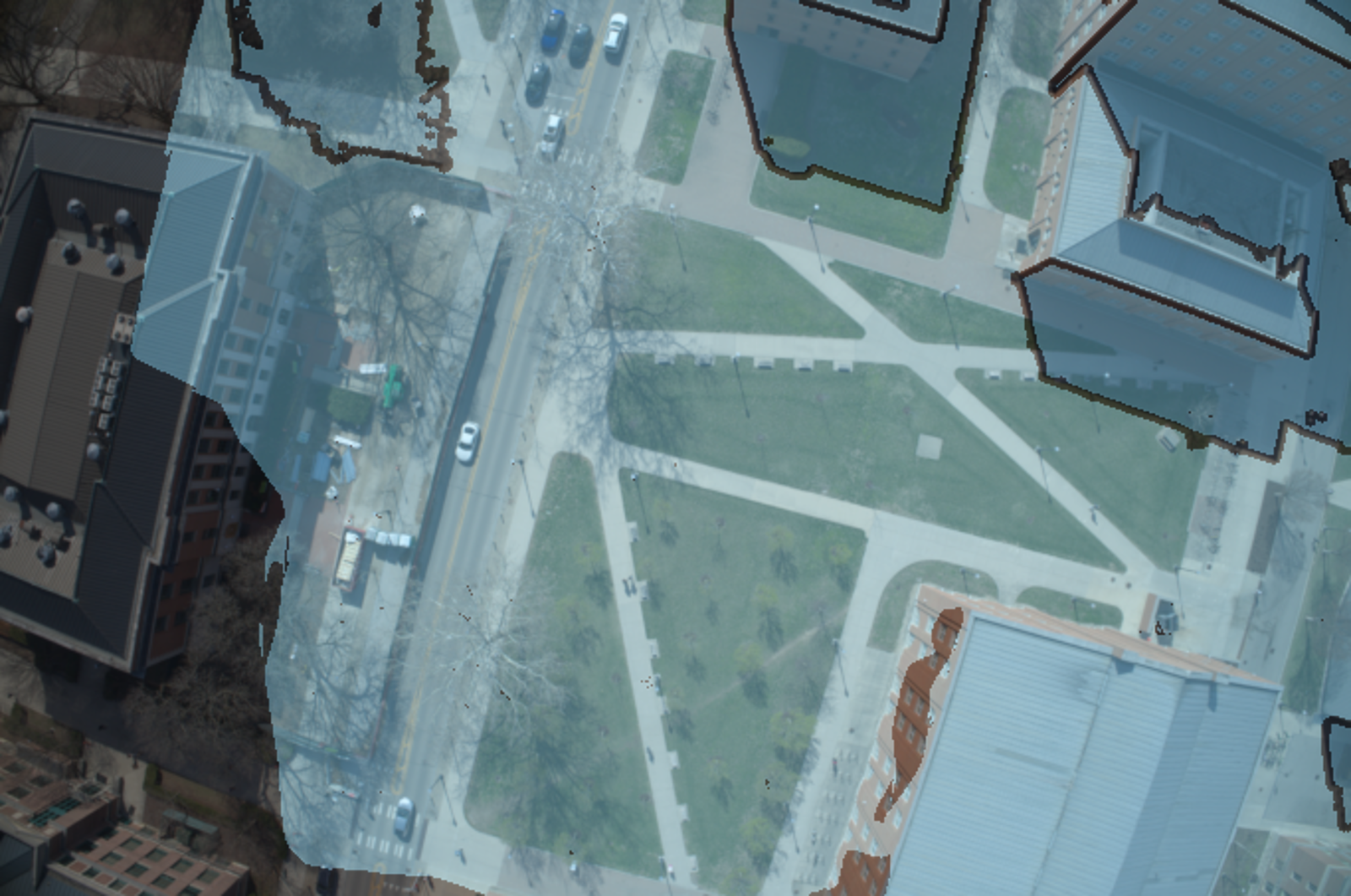}
    \caption{Example of the binary confidence mask used during fine-tuning. The mask is constructed by combining geometry boundaries and shadow-projection boundaries. Blue regions indicate high-confidence supervision, while dimmer regions are excluded from training.}
    \label{fig:confidence_mask}
\end{figure*}

The binary confidence mask (shown in \cref{fig:confidence_mask}) is automatically generated from two cues only: geometric boundaries from the reconstructed scene geometry and shadow-projection boundaries from the estimated cast-shadow map. We dilate and combine these boundaries to suppress supervision near geometry holes, depth discontinuities, and sharp shadow transitions, where the albedo--shading decomposition is most unstable. The resulting mask mainly removes artifacts around holes and shadow boundaries.

During fine-tuning, we apply this confidence mask to restrict supervision to valid regions, as described in Section~\ref{sec:experiments} of the main paper. This helps prevent unreliable pseudo-ground-truth near geometry and shadow boundaries from dominating training.

\section{Multi-view Consistent Retexturing} \label{supsec:retexturing}

\begin{figure*}[!htp]
    \centering
    \begin{subfigure}{\linewidth}
        \centering
        \includegraphics[width=\linewidth]{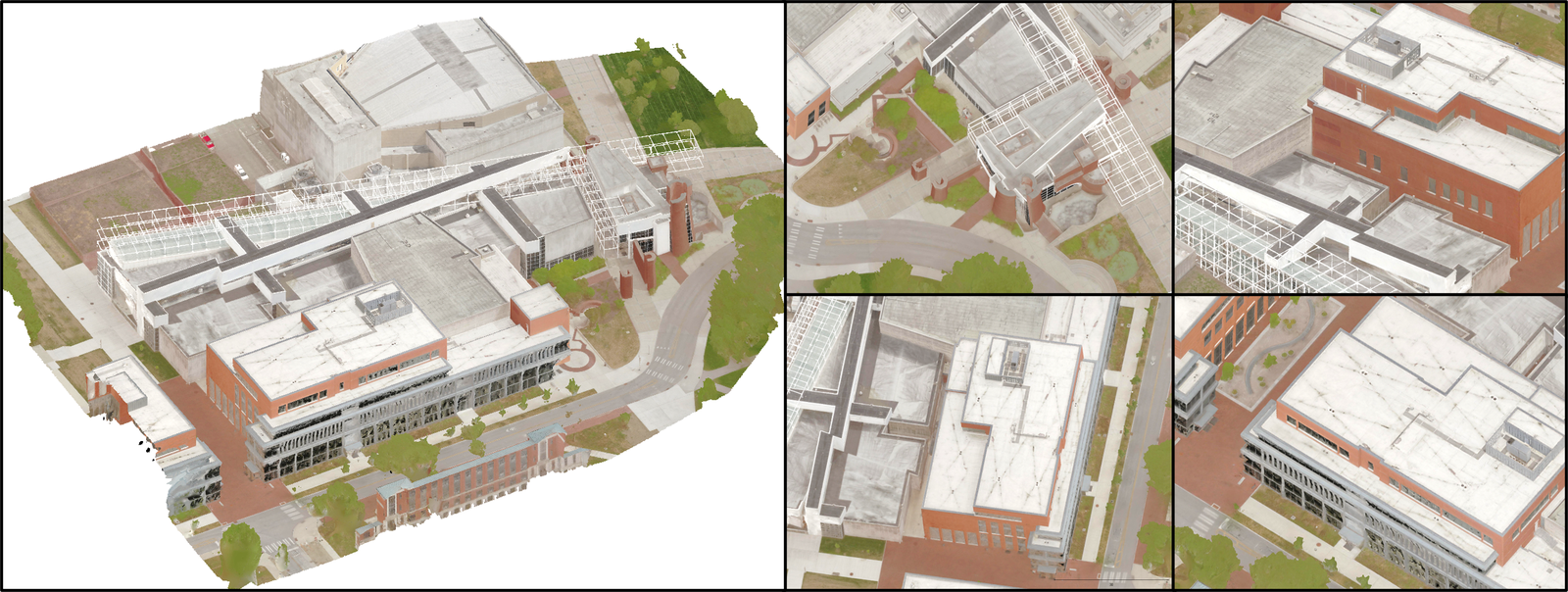}
        \caption{Campus scene.}
        \label{fig:retex_wexner}
    \end{subfigure}
    \begin{subfigure}{\linewidth}
        \centering
        \includegraphics[width=\linewidth]{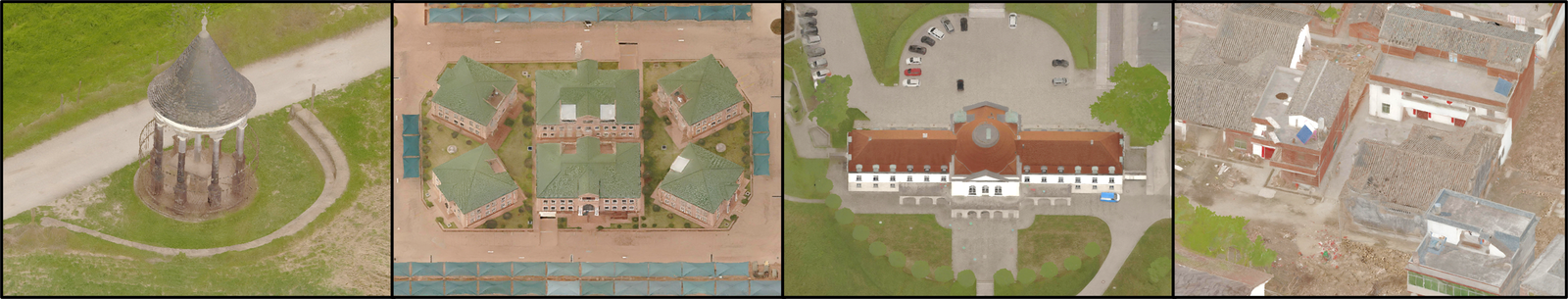}
        \caption{BlendedMVS scenes.}
        \label{fig:retex_blendedmvs}
    \end{subfigure}
    \caption{Retextured models using recovered albedo images from the fine-tuned RGB$\leftrightarrow$X.}
    \label{fig:Re_textured_model}
\end{figure*}

\begin{figure*}[!htp]
    \centering
    \begin{subfigure}{\linewidth}
        \centering
        \includegraphics[width=\linewidth]{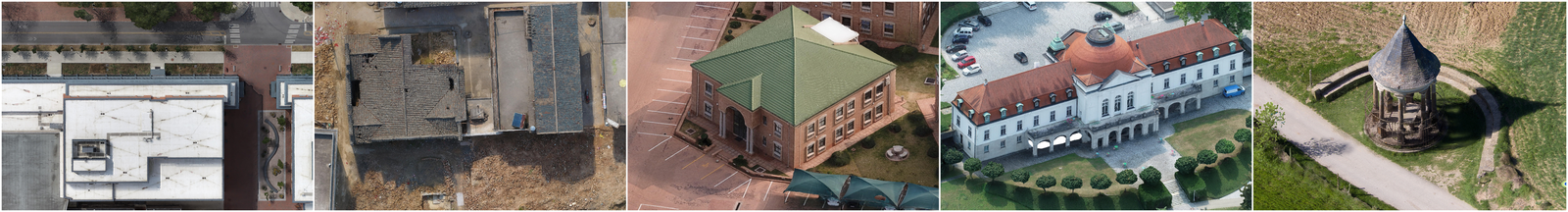}
        \caption{RGB-textured models.}
        \label{fig:supp_rgb_model}
    \end{subfigure}
    \begin{subfigure}{\linewidth}
        \centering
        \includegraphics[width=\linewidth]{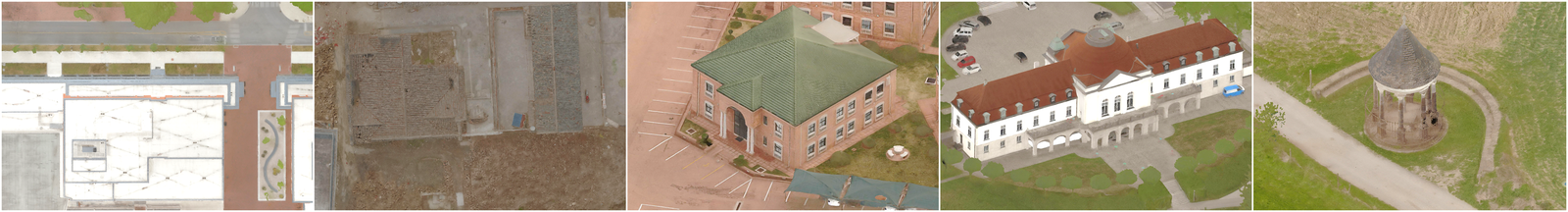}
        \caption{Albedo-retextured models.}
        \label{fig:supp_albedo_model}
    \end{subfigure}
    \caption{Comparison between RGB-textured models and albedo-retextured models. Shading is largely removed while texture details are well preserved.}
    \label{fig:textured_model_comparison}
\end{figure*}
We apply the predicted albedo obtained from our fine-tuned RGB$\leftrightarrow$X model to retexture the reconstructed 3D models. Experiments are conducted on both a self-collected UAV dataset featuring buildings with fine structural details, and the publicly available BlendedMVS dataset~\cite{yaoBlendedMVSLargescaleDataset2020} containing a variety of landscape types. As shown in \cref{fig:Re_textured_model} and \cref{fig:textured_model_comparison}, the retextured models exhibit a consistent appearance across all scenes, with most shading effects effectively removed compared to the original RGB-textured models. Notably, the retextured models preserve fine texture details—for example, building roof patterns and road markings remain clearly visible. These visualizations highlight the advantage of albedo recovery: data acquisition becomes less constrained by lighting conditions, and the albedo-based textures further enable flexible downstream applications such as relighting and material editing.

\section{Relighting Renders} \label{supsec:relighting}

\begin{figure*}[!htp]
    \centering
    \vspace{-10px}
    \setlength{\tabcolsep}{0pt} 
    \renewcommand{\arraystretch}{1.1}

    \begin{tabular}{
        >{\centering\arraybackslash}p{0.16666\linewidth}
        >{\centering\arraybackslash}p{0.16666\linewidth}
        >{\centering\arraybackslash}p{0.16666\linewidth}
        >{\centering\arraybackslash}p{0.16666\linewidth}
        >{\centering\arraybackslash}p{0.16666\linewidth}
        >{\centering\arraybackslash}p{0.16666\linewidth}
    }
        \multicolumn{6}{c}{
            \adjustbox{trim=0 0 0 {0.05\height}, clip}%
            {\includegraphics[width=\linewidth]{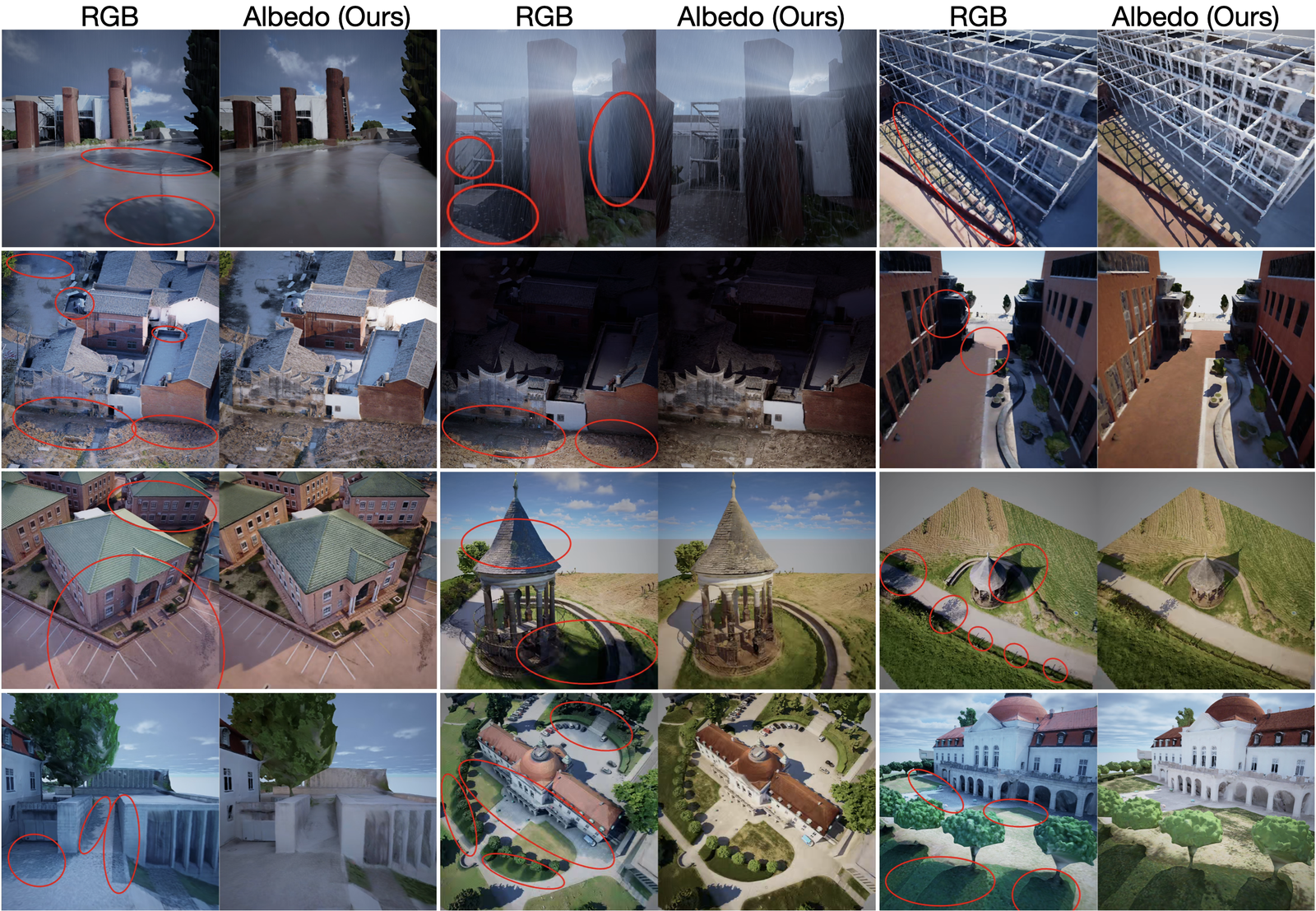}}
        } \\

        RGB & Albedo(Ours) & RGB & Albedo(Ours) & RGB & Albedo(Ours) 
    \end{tabular}
    \vspace{-10px}
    
    \caption{Visual comparisons between the RGB textures (left column of each pair) and our recovered albedo textures (right column of each pair) under novel lighting conditions. The scenes encompass diverse architectural styles, materials, and environmental conditions. Original textures contain baked-in lighting such as cast shadows, shading gradients, and environment-induced color shifts, whereas our albedo textures remove these illumination effects and retain only the intrinsic diffuse reflectance. These results highlight the stability and consistency of our method across varied outdoor environments.}
    \label{fig:relighting_sup}
\end{figure*}

We provide visual comparisons in \cref{fig:relighting_sup} across a diverse set of outdoor scenes. Each scene is rendered using both the original RGB texture and the recovered albedo texture. The examples span a wide range of structural geometries, surface materials, vegetation densities, and illumination conditions, including strong directional sunlight, partially occluded skylight, overcast weather, rainy environments, and nighttime lighting. Across all scenarios, the RGB-textured models exhibit baked-in shading effects. In contrast, the albedo-textured models consistently suppress these shading artifacts and produce spatially coherent, radiometrically stable appearances while preserving high-frequency material details. These results demonstrate that our approach generates albedo textures substantially more suitable for downstream rendering, relighting, and weather-agnostic asset creation.

\section{Additional Segmentation Results} \label{supsec:segmentation}
\begin{figure*}[!htp]
    \centering
    \vspace{-10px}
    \setlength{\tabcolsep}{0pt} 
    \renewcommand{\arraystretch}{1.1} 
    
    \begin{tabular}{
        >{\centering\arraybackslash}p{0.16666\linewidth}
        >{\centering\arraybackslash}p{0.16666\linewidth}
        >{\centering\arraybackslash}p{0.16666\linewidth}
        >{\centering\arraybackslash}p{0.16666\linewidth}
        >{\centering\arraybackslash}p{0.16666\linewidth}
        >{\centering\arraybackslash}p{0.16666\linewidth}
    }
        \multicolumn{6}{c}{
            \adjustbox{trim=0 {0.37\height} 0 0, clip}%
            {\includegraphics[width=\linewidth]{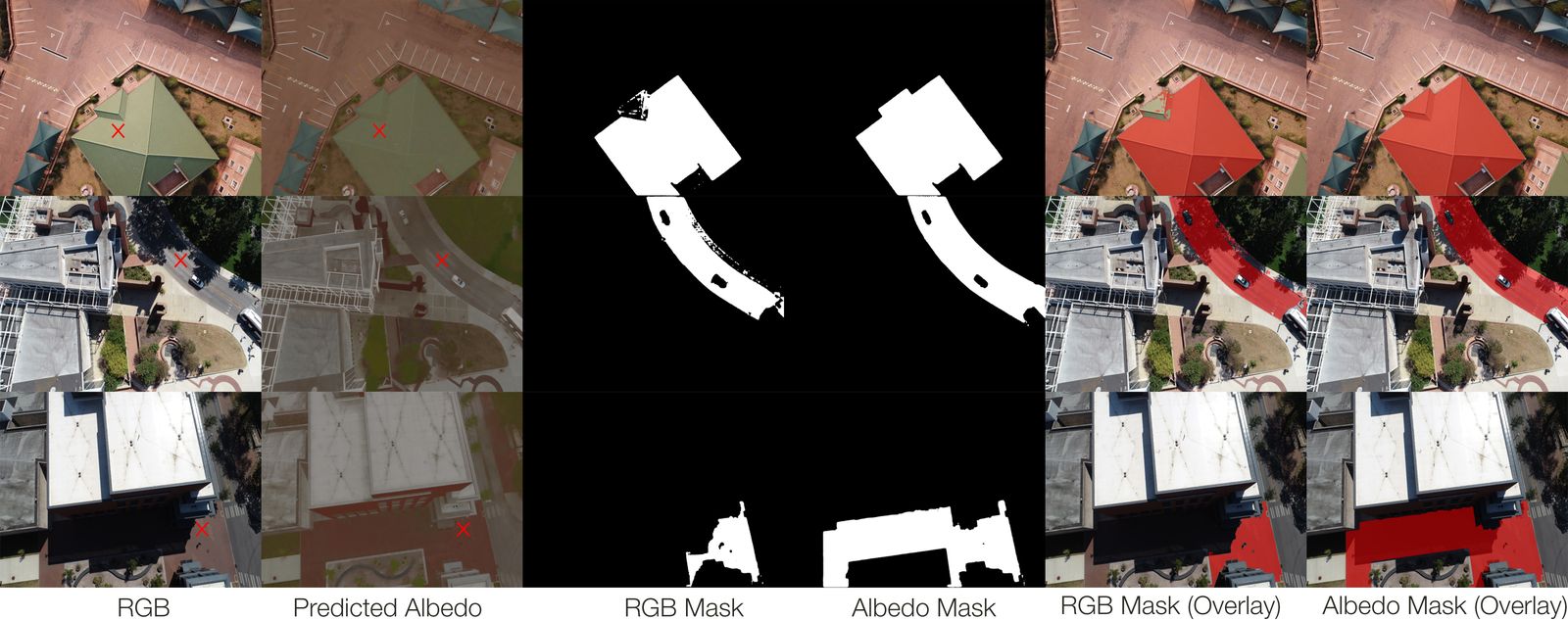}}
        } \\

        Original & Albedo & Original Mask & Albedo Mask & Original Overlay & Albedo Overlay \\
    \end{tabular}
    \vspace{-10px}

    \caption{Additional segmentation results using SAM on RGB and albedo images. Each group of six images shows, from left to right: the RGB image, the corresponding albedo, the SAM mask predicted on RGB, the SAM mask predicted on albedo, and the two masks overlaid on the RGB image. Across diverse scenes, masks obtained from albedo remain more stable in shadowed regions and better preserve object boundaries than those obtained from RGB.}
    \label{fig:SAM_comp_sup}
\end{figure*}

We provide additional qualitative segmentation examples using SAM~\cite{raviSAM2Segment2024} on both RGB and albedo images. As shown in \cref{fig:SAM_comp_sup}, the segmentation performance is consistently better on albedo images across a broad range of outdoor scenes, particularly those containing complex cast shadows.

\end{document}